\documentclass[journal,comsoc,onecolumn]{IEEEtran}
\usepackage{graphicx}
\usepackage{url}
\usepackage{amsmath,amssymb}
\usepackage{booktabs}
\usepackage{tabularx}
\usepackage{array}
\usepackage{makecell}
\usepackage{rotating}
\usepackage{tikz}
\usetikzlibrary{arrows.meta,positioning}
\usepackage{comment} % enables \begin{comment} ... \end{comment}

\renewcommand{\arraystretch}{1.15}

\makeatletter
\newcommand{\subtitle}[1]{\gdef\@subtitle{#1}}
\def\@subtitle{}
\let\oldtitle\title
\renewcommand{\title}[1]{\oldtitle{#1\\\large\@subtitle}}
\makeatother

\ifCLASSINFOpdf
\else
\fi
\begin{document}
\title{Rethinking Communication Metrics: How Should We Measure Meaning?}
\subtitle{A Survey of KPIs for Semantic Communication Systems}

% \title{Rethinking Communication Metrics: How Should We Measure Meaning?\\
% A Survey of KPIs for Semantic Communication Systems}

\author{Niloofar Tavakolian, Hakimeh Purmehdi, and Jungyeon Baek%
\thanks{N. Tavakolian, H. Purmehdi, and J. Baek are with Ericsson Canada, Montreal, QC, Canada (e-mail: niloofar.tavakolian@ericsson.com; hakimeh.purmehdi@ericsson.com; jungyeon.baek@ericsson.com).}%
\thanks{N. Tavakolian is also with Concordia University, Montreal, QC, Canada (e-mail: niloofar.tavakolian@mail.concordia.ca).}}


% NOTE (arXiv): it's best to avoid journal-identifying running headers unless the paper is
% actually published there. Toggle \arxivtrue for the arXiv upload PDF.
\newif\ifarxiv
\arxivtrue % uncomment for arXiv

\ifarxiv
  % No journal running header for arXiv/preprint version
\else
  \markboth{IEEE Communications Surveys \& Tutorials,~Vol.~XX, No.~X, Month~2026}%
  {Tavakolian \MakeLowercase{\textit{et al.}}: Rethinking Communication Metrics: How Should We Measure Meaning?}
\fi



\maketitle

\begin{abstract}
Semantic communication shifts the objective of communication systems from accurate symbol reconstruction toward meaning preservation, task accomplishment, and efficient information exchange. However, its evaluation remains fragmented across telecommunications, natural language processing, computer vision, and machine learning, and no single metric can characterize semantic quality across modalities, tasks, and channel conditions. This article surveys key performance indicators (KPIs) for text- and image-based semantic communication systems from a unified, evaluation-centered perspective. Unlike prior surveys primarily organized around architectures, applications, or transmission strategies, this work focuses on how semantic success should be defined and measured. Existing KPIs are classified according to communication goal, source modality, receiver output, reference availability, evaluation level, and channel or resource constraints. The survey reviews reconstruction-based, task-oriented, reference-free, representation-level, perceptual, and channel-aware metrics, and presents a cross-modality comparison of their roles, strengths, and limitations. It further analyzes how unresolved semantic-KPI challenges affect monitoring, quality assurance, resource optimization, fault diagnosis, and standardization. Key open problems include the absence of universal semantic success criteria and standardized semantic ground truth, semantic drift, limited reference-free evaluation, weak integration of machine-learning metrics with communication constraints, and the lack of relation-level and multimodal KPIs. Finally, future research directions are outlined toward standardized, interpretable, adaptive, task-aware, and communication-aware evaluation frameworks.
\end{abstract}


% \fi
% --- Main paper (9 sections) ---
\section{Introduction}
\label{sec:introduction}

Historically, communication system design has been guided by the Shannon paradigm, which emphasizes the reliable delivery and accurate recovery of transmitted symbols in the presence of channel impairments and noise. Consequently, communication performance has been assessed through well-established Key Performance Indicators (KPIs) such as bit
error rate, symbol error rate, channel capacity, throughput, latency, and spectral efficiency~\cite{Shannon1948, 3GPP_TR25913,
Andrews2014What5G}. More broadly, KPIs provide the fundamental
measurement framework that enables telecommunication systems to be
designed, operated, benchmarked, and optimized, serving as a common
language among network operators, equipment vendors, standardization
bodies, and end users. The role of KPIs has continuously evolved
alongside successive generations of wireless networks~\cite{Banovic2017Telecom}.
Early cellular systems (1G--3G) primarily focused on metrics such as
call blocking probability, drop-call rate, carrier-to-interference
ratio, and coverage, whereas the emergence of data-centric services in 4G~LTE broadened the KPI landscape beyond traditional reliability metrics to encompass packet-level reliability, user throughput, latency, and spectral efficiency \cite{3GPP_TR25913,
Andrews2014What5G, Osseiran2014METIS}. This progression reached a new milestone with 5G New Radio (NR), where the International Telecommunication Union
Radiocommunication Sector (ITU-R) formalized a comprehensive KPI framework through the IMT-2020 requirements, establishing ambitious performance targets for peak data rates, user-experienced throughput, latency, connection density, and spectral efficiency~\cite{ITUR_M2083,
ITUR_M2410, Reddy2021_5G_NR_KPI}. These objectives
were subsequently translated into concrete technical specifications by
the Third Generation Partnership Project (3GPP), illustrating a highly
systematic KPI-driven approach to network engineering. However, while
these conventional KPIs remain fundamental for evaluating transmission fidelity and network performance, they are insufficient for emerging
communication paradigms in which the receiver's goal extends beyond accurate symbol reconstruction to the successful recovery of meaning, knowledge, or intent, and ultimately the completion of downstream tasks~\cite{Gunduz2023BeyondBits, Yang2023SemanticFuture}.
 
To address these limitations, semantic communication has emerged as a transformative communication paradigm that prioritises the delivery of meaning rather than the accurate reproduction of transmitted symbols~\cite{Gunduz2023BeyondBits,Yang2023SemanticFuture}. In this framework, communication effectiveness is evaluated not only by traditional transmission accuracy metrics but also by the extent to which semantic information is preserved and the intended task or objective can be successfully accomplished at the receiver~\cite{Xie2021DeepSC,Bourtsoulatze2019DeepJSCC}. As a result, the role of communication extends beyond information transport, encompassing semantic representation, context understanding, reasoning, and decision-making. This fundamental shift in communication objectives necessitates a corresponding evolution in performance evaluation methodologies and metrics.
 
% Recent advances in deep learning have accelerated this shift from
% concept to implementation~\cite{Gunduz2023BeyondBits,
% Yang2023SemanticFuture}. End-to-end learned communication systems,
% Transformer-based language models, deep vision encoders, and multimodal
% neural architectures have enabled new ways to extract, compress, and
% transmit semantic representations directly. For example, DeepSC demonstrated a
% Transformer-based semantic communication system for text transmission
% that aims to recover sentence meaning rather than minimize bit- or
% symbol-level errors~\cite{Xie2021DeepSC}, while deep JSCC showed that wireless image
% transmission can be learned end-to-end without explicit source and
% channel coding~\cite{Bourtsoulatze2019DeepJSCC}.
% These developments have expanded semantic communication research from
% early theoretical discussion into practical systems for text, images,
% and multimodal tasks~\cite{Gunduz2023BeyondBits, Yang2023SemanticFuture}.
Recent advances in deep learning have accelerated the development of semantic communication by enabling end-to-end, data-driven optimization of communication systems, moving beyond the traditional separation of source coding, channel coding, and reconstruction~\cite{Gunduz2023BeyondBits,Yang2023SemanticFuture}. Representative approaches such as DeepSC \cite{Xie2021DeepSC} and DeepJSCC \cite{Bourtsoulatze2019DeepJSCC} demonstrate that neural architectures can learn communication strategies that preserve information relevant to communication objectives, rather than merely minimizing bit- or symbol-level errors. More broadly, semantic communication shifts the emphasis of communication design and evaluation from fidelity-oriented transmission toward the preservation of meaning and task-relevant information, thereby requiring performance metrics that extend beyond traditional reconstruction-based measures.

The implications of this shift become evident when considering how communication objectives vary across applications and data modalities. In text-based semantic communication, the objective may be to preserve meaning, context, and intent, whereas in image-based semantic communication, preserving semantically relevant visual information or downstream utility may be more important than exact pixel reconstruction. More generally, the notion of successful communication depends on the receiver's objective, which may range from source reconstruction to tasks such as classification, retrieval, question answering, or decision making. Consequently, no single universal KPI can adequately characterise semantic communication performance. Instead, evaluation metrics must be aligned with the communication objective, source modality, receiver output, and communication constraints. This motivates the development of semantic-aware and task-aware KPIs, as well as communication-oriented semantic metrics that quantify reliability, efficiency, and performance under network resource constraints.

Motivated by the need to understand how semantic information can be measured, preserved, and evaluated, this survey provides a comprehensive review of KPI design and evaluation methodologies for semantic communication systems. Focusing on text and image modalities, we examine how existing metrics capture diverse communication objectives, including semantic reconstruction, semantic state recovery, representation-level similarity, perceptual quality, downstream task performance, and telecom-oriented semantic metrics related to reliability, efficiency, and communication constraints. By analyzing these KPIs across different modalities and application settings, we identify common evaluation principles, highlight modality-specific challenges, and discuss the limitations of current approaches. Finally, we outline emerging research directions toward the development of standardized, interpretable, robust, and broadly applicable KPI frameworks for future semantic communication networks.
 
The main contributions of this survey are summarized as follows:
\begin{enumerate}
 
    \item We provide a KPI-centred review of semantic communication for text and image modalities, highlighting the relationship between communication objectives and performance evaluation.
 
    \item We develop a unified taxonomy of semantic communication KPIs, covering reconstruction fidelity, semantic similarity, semantic state recovery, task utility, reference-free evaluation, perceptual quality, distribution-level realism, and telecom-oriented semantic reliability and efficiency metrics.
 
    \item We compare KPI design and evaluation methodologies across text and image modalities, identifying both common principles and modality-specific challenges.
 
    \item We discuss the applicability and limitations of existing KPI categories for different semantic communication objectives, deployment scenarios, and communication constraints.

    \item We identify key open challenges and future research directions, including semantic state representation, reference-free evaluation, benchmark standardization, KPI interpret-ability, and robust semantic-aware performance assessment.
 
    % \item It proposes a systematic KPI selection framework that maps
    % semantic communication scenarios to recommended KPI combinations.
    % While prior surveys have organized metrics by modality and task,
    % they have not provided a structured procedure for selecting KPI
    % combinations based on the full set of evaluation conditions a
    % practitioner faces. The framework takes five input axes (source
    % modality, communication goal, receiver output type, reference
    % availability, and channel constraint) and follows a four-step
    % procedure to produce a scenario-specific evaluation protocol. A
    % scenario mapping table covering ten canonical semantic
    % communication scenarios and a three-level composite evaluation
    % protocol are provided to support reproducible and
    % communication-aware system evaluation.

\end{enumerate}
 
The remainder of this paper is organized as follows. Section II presents the proposed taxonomy of semantic communication KPIs. Sections III and IV review KPIs for text and image semantic communication. Section V compares these KPIs across modalities and communication objectives. Section VI discusses open challenges in semantic KPI design, Section VII outlines future research directions, and Section VIII concludes the paper.
%Section~6 introduces the systematic KPI selection framework and scenario-based evaluation protocol.
\section{Evaluation Taxonomy for Semantic Communication}
\label{sec:evaluation_taxonomy}
Semantic communication systems differ not only in their architectures but also in their definitions of communication success. Unlike conventional communication systems, where performance is typically evaluated using metrics such as bit error rate, throughput, latency, or distortion, semantic communication requires evaluation criteria that depend on the information that must be preserved, recovered, or utilised by the receiver. As a result, communication success may be defined in terms of source reconstruction, semantic preservation, semantic state recovery, or downstream task accomplishment.

To systematically analyse semantic communication KPIs, this survey distinguishes five related concepts: use-case, task, communication goal, receiver output, and KPI. A use-case defines the application scenario, such as intelligent transportation, remote monitoring, multimedia delivery, or human-machine interaction. A task specifies the operation performed by the receiver, such as reconstruction, classification, retrieval, segmentation, or question answering. The communication goal determines what information must be retained for successful communication, while the receiver output specifies the form in which the result is produced. Finally, a KPI quantifies the extent to which the communication goal is achieved under given communication constraints.

Table~\ref{tab:evaluation_axes} summarizes the main evaluation axes considered in this survey. These include the communication goal, source modality, receiver output, reference availability, evaluation level, and communication constraints. Together, these axes determine whether performance should be assessed using reconstruction-oriented, semantic similarity, task-oriented, reference-free, or telecom-oriented semantic KPIs.

\begin{table*}[t]
\centering
\caption{General evaluation axes for selecting semantic communication KPIs}
\label{tab:evaluation_axes}
\small
\renewcommand{\arraystretch}{1.25}
\setlength{\tabcolsep}{4pt}
\begin{tabularx}{\textwidth}{p{0.18\textwidth} p{0.26\textwidth} X}
\toprule
\textbf{Axis} & \textbf{Possible categories} & \textbf{Why it matters} \\
\midrule
Communication goal &
Reconstruction, semantic preservation, semantic state recovery, task accomplishment &
Determines whether success is measured by similarity, task completion, or semantic distortion. \\

Source modality &
Text, image, multimodal &
Determines the form of semantic content and the suitable KPI families. \\

Receiver output &
Reconstructed source, embedding, label, action, structured symbols &
Determines what variables can be compared during evaluation. \\

Reference availability &
Reference based, reference free &
Determines whether the original source or gold reference is needed. \\

Evaluation level &
Sample level, dataset level, system level &
Determines whether metrics are computed per example, over distributions, or across systems. \\

Channel/resource condition &
SNR, bandwidth, rate, latency, energy &
Enables communication-aware semantic evaluation and telecom-oriented KPIs such as semantic reliability, throughput, efficiency, and QoS/QoE measures. \\

\bottomrule
\end{tabularx}
\end{table*}

\begin{figure*}[t]
\centering
\resizebox{0.95\textwidth}{!}{%
\begin{tikzpicture}[
    node distance=1.2cm and 1.0cm,
    box/.style={
        draw,
        rounded corners,
        align=center,
        minimum height=0.95cm,
        font=\footnotesize
    },
    arrow/.style={-{Latex[length=2.0mm]}, thick}
]

% Main pipeline (wider boxes + larger horizontal spacing)
\node[box, text width=2.4cm] (S) {$S$\\Source input};
\node[box, right=of S, text width=2.7cm] (W) {$W$\\Semantic\\representation};
\node[box, right=of W, text width=2.0cm] (C) {Channel};
\node[box, right=of C, text width=3.1cm] (Wh) {$\hat{W}$\\Recovered\\semantic representation};
\node[box, right=of Wh, text width=2.7cm] (Sh) {$\hat{S}$\\Reconstructed output};

\draw[arrow] (S) -- (W);
\draw[arrow] (W) -- (C);
\draw[arrow] (C) -- (Wh);
\draw[arrow] (Wh) -- (Sh);

% Downstream task output
\node[box, below=1.3cm of W, text width=3cm] (Y) {$Y$\\Ground-truth task output};
\node[box, below=1.3cm of Wh, text width=3cm] (Yh) {$\hat{Y}$\\Predicted task output};

\draw[arrow] (W) -- (Y);
\draw[arrow] (Wh) -- (Yh);

\end{tikzpicture}%
}
\caption{Generic semantic communication evaluation pipeline. A source signal $S$ (e.g., text, image, or multi-modal content) is encoded into a semantic representation $W$, transmitted through a communication channel, and recovered as $\hat{W}$. Depending on the communication objective, the receiver may reconstruct an output $\hat{S}$ or generate a task-specific output $\hat{Y}$. Modality-specific notations used in later sections are derived from this generic framework. Different KPIs evaluate performance at different stages of the pipeline, including reconstruction fidelity, semantic preservation, task effectiveness, and telecom-oriented semantic reliability and efficiency under communication constraints.}
\label{fig:semantic_comm_pipeline}
\end{figure*}
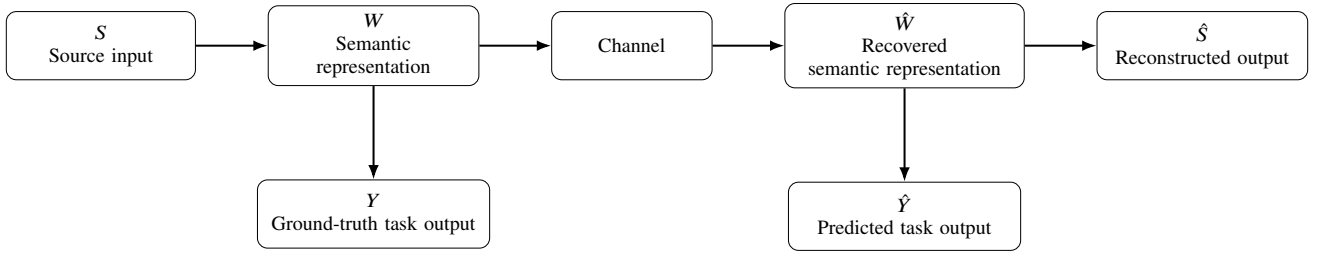

To facilitate the discussion of KPI families throughout this survey, Fig.~\ref{fig:semantic_comm_pipeline} illustrates a generic semantic communication pipeline. A source signal is encoded into a semantic representation, transmitted through a communication channel, and recovered at the receiver. Depending on the communication objective, the receiver may reconstruct the source content, recover a semantic state, or generate a task-specific output.

\begin{comment}
\begin{figure*}[t]
\centering
\resizebox{0.95\textwidth}{!}{%
\begin{tikzpicture}[
    node distance=1.1cm and 0.55cm,
    box/.style={
        draw,
        rounded corners,
        align=center,
        minimum height=0.9cm,
        minimum width=1.55cm,
        font=\footnotesize
    },
    arrow/.style={-{Latex[length=2.5mm]}, thick}
]

\node[box] (S) {$S$\\Original\\sentence};
\node[box, right=of S] (X) {$X$\\Token\\sequence};
\node[box, right=of X] (W) {$W$\\Semantic\\representation};
\node[box, right=of W] (C) {Channel};
\node[box, right=of C] (Wh) {$\hat{W}$\\Recovered\\semantic\\representation};
\node[box, right=of Wh] (Xh) {$\hat{X}$\\Reconstructed\\token sequence};
\node[box, right=of Xh] (Sh) {$\hat{S}$\\Reconstructed\\sentence};

\draw[arrow] (S) -- (X);
\draw[arrow] (X) -- (W);
\draw[arrow] (W) -- (C);
\draw[arrow] (C) -- (Wh);
\draw[arrow] (Wh) -- (Xh);
\draw[arrow] (Xh) -- (Sh);

\node[box, below=1.2cm of W] (Y) {$Y$\\Ground-truth\\task output};
\node[box, below=1.2cm of Wh] (Yh) {$\hat{Y}$\\Predicted\\task output};

\draw[arrow] (W) -- (Y);
\draw[arrow] (Wh) -- (Yh);

\end{tikzpicture}%
}
\caption{Notation and evaluation pipeline for text-based semantic communication. The source sentence $S$ is encoded into a token sequence $X$ and mapped to a semantic representation $W$, which is transmitted through the channel and recovered as $\hat{W}$. The receiver may reconstruct the token sequence $\hat{X}$ and final sentence $\hat{S}$, or produce a task output $\hat{Y}$.}
\label{fig:text_semcom_pipeline}
\end{figure*}
\end{comment}
Based on the evaluation axes in Table~\ref{tab:evaluation_axes} and the pipeline in Fig.~\ref{fig:semantic_comm_pipeline}, KPI selection depends on the communication objective, receiver output, source modality, and operating conditions. Evaluation may focus on reconstruction fidelity, semantic similarity, semantic state recovery, downstream task effectiveness, perceptual quality, or telecom-oriented semantic metrics such as semantic reliability, semantic throughput, and semantic efficiency. The proposed taxonomy therefore provides a unified framework for selecting and comparing KPI families across semantic communication systems.
\section{Text-Based Semantic Communication KPIs}
\label{sec:text_kpis}

Text-based semantic communication is among the earliest and most extensively studied forms of deep learning-enabled semantic communication. In this setting, the generic source $S$ in Fig.~\ref{fig:semantic_comm_pipeline} corresponds to a text message, sentence, or document. For clarity, we denote the tokenized representation of the source text by $X$, its semantic representation by $W$W, and their reconstructed counterparts by $\hat{X}$ and $\hat{W}$, respectively. Depending on the communication objective, the receiver may reconstruct a semantically equivalent sentence $\hat{S}$, recover a task-relevant semantic representation $\hat{W}$, or produce a downstream task output $\hat{Y}$. Consequently, text-based semantic communication KPIs can be broadly categorized into semantic reconstruction metrics, task-oriented metrics, reference-free evaluation metrics, and telecom-oriented semantic state recovery metrics, as summarized in Table~\ref{tab:text_kpi_categories}.

\begin{table*}[t]
\centering
\caption{KPI Categories for Text-Based Semantic Communication}
\label{tab:text_kpi_categories}
\small
\renewcommand{\arraystretch}{1.25}
\setlength{\tabcolsep}{4pt}
\begin{tabularx}{\textwidth}{p{0.22\textwidth} p{0.33\textwidth} X}
\toprule
\textbf{Text KPI category} & \textbf{Core question} & \textbf{Representative KPIs} \\
\midrule
Semantic reconstruction &
Did the reconstructed text preserve the original meaning or wording? &
BLEU, ROUGE, METEOR, chrF, BERT cosine, BERTScore, BLEURT, COMET \\

Goal-oriented task success &
Did the receiver complete the intended task? &
accuracy, F1, exact match, Recall@K, MRR, task success rate \\

Reference-free evaluation &
Can quality be estimated without a gold reference at the receiver? &
COMET-QE, anchor similarity, semantic outage, perplexity \\

Telecom-style semantic state recovery &
Was the semantic state recovered under channel/resource constraints? &
semantic distortion, semantic error probability, rate--distortion curves, outage probability \\
\bottomrule
\end{tabularx}
\end{table*}

The following subsections discuss these categories in more detail and explain how each KPI family reflects a different definition of successful text-based semantic communication.

\subsection{Semantic Reconstruction Metrics}

Semantic reconstruction is the most common evaluation setting in early text-based semantic communication systems. In this setting, the source message $S$ is an input sentence, and the receiver generates a reconstructed sentence $\hat{S}$ after transmission through a noisy channel. The goal is not necessarily to reproduce the exact same word sequence, but to preserve the meaning of the original sentence. Therefore, reconstruction-based KPIs are computed at the sentence level by comparing $S$ and $\hat{S}$, and they can be divided into surface-overlap metrics, embedding-based semantic similarity metrics, and learned evaluation metrics.

Surface-overlap metrics, such as BLEU, ROUGE, METEOR, and chrF, compare the reconstructed sentence with a reference sentence using word, phrase, or character-level overlap. These metrics are simple, widely used, and useful for measuring lexical reconstruction quality. However, they may penalize valid paraphrases because they focus mainly on visible token overlap rather than meaning.

\subsubsection{BLEU}

BLEU\cite{Papineni2002BLEU} measures modified n-gram precision between the reconstructed sentence $\hat{S}$ and the reference sentence $S$, together with a brevity penalty.

For each n-gram order $n$, the modified precision is:
\begin{equation}
p_n =
\frac{
\sum_{g \in \hat{S}} \min \left( \mathrm{count}_{\hat{S}}(g), \mathrm{count}_{S}(g) \right)
}{
\sum_{g \in \hat{S}} \mathrm{count}_{\hat{S}}(g)
}
\end{equation}

where $g$ denotes an n-gram. The brevity penalty is:
\begin{equation}
BP =
\left\{
\begin{array}{l}
1 \quad \mathrm{if}\ c > r,\\
e^{1-r/c} \quad \mathrm{if}\ c \leq r.
\end{array}
\right.
\end{equation}

where $c$ is the length of the reconstructed sentence and $r$ is the length of the reference sentence. BLEU is computed as:
\begin{equation}
BLEU = BP \cdot \exp \left( \sum_{n=1}^{N} w_n \log p_n \right)
\end{equation}

In text-based semantic communication, BLEU compares $S$ and $\hat{S}$ at the sentence level, rather than directly comparing the tokenized sequences $X$ and $\hat{X}$ or the semantic representations $W$ and $\hat{W}$. A higher BLEU score indicates stronger lexical overlap with the reference. However, BLEU may penalize valid paraphrases because it depends on exact n-gram matches. A numerical example is provided in Appendix~\ref{apx_bleu_example}.

\subsubsection{ROUGE-N and ROUGE-L}

ROUGE-N\cite{Lin2004ROUGE} measures reference n-gram recall. It checks how much of the reference/source sentence $S$ is covered by the reconstructed sentence $\hat{S}$.

\begin{equation}
ROUGE\text{-}N =
\frac{
\sum_{g \in S} \min \left( \mathrm{count}_{\hat{S}}(g), \mathrm{count}_{S}(g) \right)
% \sum_{g \in S} \min \left( \mathrm{count}*{\hat{S}}(g), \mathrm{count}*{S}(g) \right)
}{
\sum_{g \in S} \mathrm{count}_{S}(g)
}
\end{equation}

ROUGE-L is based on the longest common subsequence, denoted by $LCS(S,\hat{S})$. Precision and recall are:

\begin{equation}
P_{LCS} =
\frac{LCS(S,\hat{S})}{|\hat{S}|}
\end{equation}

\begin{equation}
R_{LCS} =
\frac{LCS(S,\hat{S})}{|S|}
\end{equation}

ROUGE-L is then computed as:

\begin{equation}
ROUGE\text{-}L =
\frac{(1+\beta^2) P_{LCS} R_{LCS}}
{R_{LCS}+\beta^2 P_{LCS}}
\end{equation}

In text-based semantic communication, ROUGE can be useful when the goal is to evaluate how much of the source content is lexically preserved in the reconstructed sentence. However, because ROUGE is still based on surface overlap and word order, it may underestimate meaning preservation when the receiver produces a valid paraphrase. A numerical example illustrating ROUGE-N and ROUGE-L is provided in Appendix~\ref{apx_rouge_example}.
% A numerical example illustrating ROUGE-N and ROUGE-L is provided in Appendix~A.2.

\subsubsection{METEOR}

METEOR\cite{BanerjeeLavie2005METEOR} aligns words between $S$ and $\hat{S}$ using exact matches, stemming, and sometimes synonym matching. It first computes unigram precision and recall:

\begin{equation}
P =
\frac{m}{|\hat{S}|}
\end{equation}

\begin{equation}
R =
\frac{m}{|S|}
\end{equation}

where $m$ is the number of matched unigrams. METEOR then computes a weighted harmonic mean:

\begin{equation}
F_{mean} =
\frac{PR}{\alpha P + (1-\alpha)R}
\end{equation}

To penalize fragmented alignments, METEOR uses a fragmentation penalty:

\begin{equation}
Penalty =
\gamma \left(\frac{ch}{m}\right)^{\theta}
\end{equation}

where $ch$ is the number of matched chunks. The final score is:

\begin{equation}
METEOR =
(1-Penalty)\cdot F_{mean}
\end{equation}

In text-based semantic communication, METEOR is more tolerant of meaning-preserving paraphrases than BLEU and ROUGE because it can include stemming and synonym-based matching. However, it is still reference-based and depends on lexical matching resources. A numerical example illustrating METEOR is provided in Appendix~\ref{apx_meteor_example}.

\subsubsection{chrF and chrF++}

chrF\cite{Popovic2015chrF} evaluates the similarity between $S$ and $\hat{S}$ using character-level n-gram precision and recall. Unlike BLEU and ROUGE, which operate mainly at the word or word n-gram level, chrF compares shorter character sequences. Therefore, it can be more robust to morphological variation, spelling differences, and tokenization mismatches. For each character n-gram order $n$, precision and recall are computed as:

\begin{equation}
P_n =
\frac{
\left|
\mathrm{char}\text{-}\mathrm{ngrams}_n(\hat{S})
\cap
\mathrm{char}\text{-}\mathrm{ngrams}_n(S)
\right|
}{
\left|
\mathrm{char}\text{-}\mathrm{ngrams}_n(\hat{S})
\right|
}
\end{equation}

\begin{equation}
R_n =
\frac{
\left|
\mathrm{char}\text{-}\mathrm{ngrams}_n(\hat{S})
\cap
\mathrm{char}\text{-}\mathrm{ngrams}_n(S)
\right|
}{
\left|
\mathrm{char}\text{-}\mathrm{ngrams}_n(S)
\right|
}
\end{equation}

After averaging precision and recall across n-gram orders, chrF is computed as:

\begin{equation}
chrF =
\frac{(1+\beta^2)PR}{\beta^2 P + R}
\end{equation}

In text-based semantic communication, chrF can be useful when reconstruction errors involve spelling, morphology, or tokenization differences. However, because it operates on character-level surface patterns, it does not directly measure semantic equivalence. A numerical example illustrating chrF and its limitation for semantic paraphrases is provided in Appendix~\ref{apx_chrf_example}.
% A numerical example illustrating chrF and its limitation for semantic paraphrases is provided in Appendix~A.4.

\subsubsection{BERT-based Sentence Similarity}

BERT-based sentence similarity embeds the original sentence $S$ and reconstructed sentence $\hat{S}$ into a semantic vector space. Let $B(\cdot)$ denote a sentence encoder such as BERT or Sentence-BERT\cite{ReimersGurevych2019SentenceBERT}.

\begin{equation}
Sim(S,\hat{S}) =
\cos \left( B(S), B(\hat{S}) \right)
\end{equation}

Equivalently:

\begin{equation}
Sim(S,\hat{S}) =
\frac{
B(S) \cdot B(\hat{S})
}{
|B(S)| , |B(\hat{S})|
}
\end{equation}

In practice, both sentences are tokenized, encoded by the language model, pooled into sentence-level vectors, and compared using cosine similarity. This metric is more semantic than n-gram overlap because it compares learned sentence representations rather than exact surface forms, but the score depends on the encoder, pooling method, layer choice, and normalization.
A numerical example illustrating BERT-based sentence similarity is provided in Appendix~\ref{apx_bert_similarity_example}.
% A numerical example illustrating BERT-based sentence similarity is provided in Appendix~A.5.

\subsubsection{BERTScore}

BERTScore\cite{Zhang2020BERTScore} compares token-level contextual embeddings. Let $x_i$ be the embedding of the $i$-th token in the reference sentence $X$, and let $\hat{x}_j$ be the embedding of the $j$-th token in the reconstructed sentence $\hat{X}$. The token-pair similarity is:

\begin{equation}
s_{ij} =
\cos \left( x_i, \hat{x}_j \right)
\end{equation}

BERTScore precision is:

\begin{equation}
P_{BERT} =
\frac{1}{|\hat{X}|}
\sum_j \max_i s_{ij}
\end{equation}

BERTScore recall is:

\begin{equation}
R_{BERT} =
\frac{1}{|X|}
\sum_i \max_j s_{ij}
\end{equation}

The BERTScore F1 score is:

\begin{equation}
F1_{BERT} =
\frac{2P_{BERT}R_{BERT}}{P_{BERT}+R_{BERT}}
\end{equation}

In text-based semantic communication, BERTScore is useful because it compares contextual token meanings rather than exact token overlap. This makes it more robust to meaning-preserving paraphrases than BLEU, ROUGE, or chrF. However, BERTScore still requires a reference sentence and depends on the pretrained encoder used to extract contextual embeddings. A conceptual example illustrating BERTScore is provided in Appendix~\ref{apx_bertscore}.
% A conceptual example illustrating BERTScore is provided in Appendix~A.6.

Learned evaluators, such as BLEURTBLEURT \cite{Sellam2020BLEURT} and COMET\cite{Rei2020COMET}, use neural models trained to predict human quality judgments. These metrics can capture more complex aspects of meaning preservation and fluency than simple n-gram overlap. However, they are model-dependent, may be sensitive to domain shift, and require careful reporting of the chosen model version and input format.

\subsubsection{BLEURT}

BLEURT\cite{Sellam2020BLEURT} is a learned metric. Instead of using a fixed overlap formula, it uses a neural model to predict a quality score for a candidate sentence relative to a reference.

Conceptually:

\begin{equation}
BLEURT(X,\hat{X}) =
f_{\theta}(X,\hat{X})
\end{equation}

The input is usually packed as:

\begin{equation}
[\mathrm{CLS}]; X; [\mathrm{SEP}]; \hat{X}; [\mathrm{SEP}]
\end{equation}

The model takes the contextual representation of the $[\mathrm{CLS}]$ token and uses a regression head:

\begin{equation}
score =
w^{\top} h_{\mathrm{CLS}} + b
\end{equation}

BLEURT is trained to correlate with human quality judgments. In semantic communication, it can be used as a learned semantic reconstruction metric, but it is model-dependent and may suffer from domain shift. A conceptual example illustrating BLEURT-based evaluation is provided in Appendix~\ref{apx_bleurt_example}.
% A conceptual example illustrating BLEURT-based evaluation is provided in Appendix~A.7.

Table~\ref{tab:text_semantic_reconstruction_metrics} summarizes the main text semantic reconstruction metrics, their comparison targets, formula ideas, and key limitations.

\begin{table*}[t]
\centering
\caption{Summary of Text Semantic Reconstruction Metrics}
\label{tab:text_semantic_reconstruction_metrics}
\small
\renewcommand{\arraystretch}{1.25}
\resizebox{\textwidth}{!}{%
\begin{tabular}{p{2.2cm} p{2.8cm} p{3.0cm} p{4.0cm} p{3.5cm}}
\toprule
\textbf{Metric} & \textbf{Type} & \textbf{Main comparison} & \textbf{Formula idea} & \textbf{Main limitation} \\
\midrule
BLEU & Surface-overlap & $S$ vs. $\hat{S}$ & Modified n-gram precision and brevity penalty & Penalizes paraphrases \\
ROUGE & Surface-overlap & $S$ vs. $\hat{S}$ & n-gram recall or LCS overlap & Mostly lexical \\
METEOR & Surface-overlap and alignment & $S$ vs. $\hat{S}$ & Unigram F-score with fragmentation penalty & Still reference-bound \\
chrF / chrF++ & Character/word overlap & $S$ vs. $\hat{S}$ & Character n-gram F-score & Not truly semantic \\
BERT cosine & Embedding-based & $B(S)$ vs. $B(\hat{S})$ & Cosine similarity & Depends on encoder and pooling \\
BERTScore & Embedding-based & Contextual token embeddings & Max token cosine matching & Requires reference \\
BLEURT & Learned evaluator & $S$, $\hat{S}$ & Neural regression score & Model/domain dependent \\
COMET & Learned evaluator & Source, candidate, reference & Neural regression score & Variant/input dependent \\
\bottomrule
\end{tabular}%
}
\end{table*}

\subsection{Goal-Oriented Text Semantic Communication KPIs}
\label{sec:goal_oriented_text_kpis}
Unlike semantic reconstruction metrics, goal-oriented text semantic communication evaluates whether the receiver can successfully complete a downstream task using the received or reconstructed information. In this setting, the exact wording of the reconstructed sentence may be less important than whether the receiver correctly infers the intended meaning, decision, or action. Therefore, the KPI is usually defined by the task output rather than by lexical similarity between the source sentence $S$ and the reconstructed sentence $\hat{S}$.

In goal-oriented text semantic communication, the source sentence may be mapped to a task-relevant semantic representation $W$, such as an intent label, slot structure, retrieval query representation, answer, or action command. The receiver then produces a predicted task output $\hat{Y}$, which is compared with the ground-truth task output $Y$. Common goal-oriented KPIs include accuracy, precision, recall, F1 score, exact match, Recall@K, mean reciprocal rank, and task success rate. These metrics are especially important when semantic communication is used to support applications such as intent classification, slot filling, question answering, retrieval, dialogue systems, or command execution.

\subsubsection{Intent Classification Accuracy}

Intent classification evaluates whether the receiver correctly identifies the user's intended action or semantic category. For example, in a task-oriented dialogue system, the source sentence $S$ may be ``Book a flight to Montreal tomorrow,'' and the ground-truth intent $Y$ may be BookFlight. The receiver predicts an intent label $\hat{Y}$, and the prediction is correct if $\hat{Y}=Y$.

Accuracy is computed as:
\begin{equation}
\mathrm{Accuracy} =
\frac{N_{\mathrm{correct}}}{N}
\end{equation}

where $N_{\mathrm{correct}}$ is the number of correctly predicted intent labels and $N$ is the total number of evaluated samples.

In goal-oriented text semantic communication, intent accuracy is useful when the receiver only needs to recover the correct high-level meaning or action type, rather than reconstructing the full sentence. However, intent accuracy alone may be too coarse when the task also requires extracting specific arguments, such as destination, date, time, or object names. An illustrative example of intent classification accuracy is provided in Appendix~B.A.

\subsubsection{Slot Filling Precision, Recall, and F1}

Slot filling evaluates whether the receiver correctly extracts task-relevant entities or arguments from the input sentence. For example, in the sentence ``Book a flight to Montreal tomorrow,'' the relevant slots may be destination = Montreal and date = tomorrow. Unlike intent classification, which predicts one global label, slot filling evaluates whether the receiver preserves the detailed semantic information needed to complete the task.

Precision, recall, and F1 score are commonly used:
\begin{equation}
\mathrm{Precision} =
\frac{TP}{TP+FP}
\end{equation}

\begin{equation}
\mathrm{Recall} =
\frac{TP}{TP+FN}
\end{equation}

\begin{equation}
F_1 =
\frac{2 \cdot \mathrm{Precision} \cdot \mathrm{Recall}}
{\mathrm{Precision}+\mathrm{Recall}}
\end{equation}

where $TP$ denotes correctly predicted slots, $FP$ denotes predicted slots that are incorrect or not present in the ground truth, and $FN$ denotes ground-truth slots that the receiver failed to predict.

In text-based semantic communication, slot F1 is useful because it evaluates whether important semantic details are preserved after transmission. A reconstructed sentence may be grammatically correct but still fail the task if it loses a critical slot value. An illustrative example of slot filling precision, recall, and F1 is provided in Appendix~\ref{apx_slot_filling_example}.
% An illustrative example of slot filling precision, recall, and F1 is provided in Appendix~B.2.

\subsubsection{Frame Accuracy / Exact Match}

Frame accuracy, also called exact match in some task-oriented language understanding settings, evaluates whether the complete semantic frame is predicted correctly. A semantic frame typically includes both the intent and all required slots. Unlike intent accuracy or slot F1, frame accuracy is stricter because a prediction is counted as correct only if the full task meaning is recovered.

For $N$ samples, frame accuracy can be written as:
\begin{equation}
\mathrm{FrameAccuracy} =
\frac{1}{N}
\sum_{i=1}^{N}
\mathbf{1}[\hat{Y}_i=Y_i]
\end{equation}

where $Y_i$ is the complete ground-truth semantic frame for sample $i$, $\hat{Y}_i$ is the predicted semantic frame, and $\mathbf{1}(\cdot)$ is an indicator function that equals 1 when the predicted frame exactly matches the ground truth and 0 otherwise.

In goal-oriented semantic communication, frame accuracy is useful when partial meaning recovery is not sufficient. For example, if the receiver correctly identifies the intent BookFlight but misses the destination or date, the frame is incorrect. An illustrative example of frame accuracy is provided in Appendix~\ref{apx_frame_accuracy_example}.
% An illustrative example of frame accuracy is provided in Appendix~B.3.

\subsubsection{Question Answering Metrics: Exact Match and F1}

In question answering tasks, the receiver is expected to produce an answer rather than reconstruct the original sentence. The task output $Y$ is the ground-truth answer, and $\hat{Y}$ is the receiver-side predicted answer. Two common KPIs are exact match and token-level F1.

Exact match is defined as:
\begin{equation}
\mathrm{EM} =
\frac{1}{N}
\sum_{i=1}^{N}
\mathbf{1}[\hat{Y}_i=Y_i]
\end{equation}

where the prediction is counted as correct only if the predicted answer exactly matches the ground-truth answer after normalization.

Token-level F1 gives partial credit when the predicted answer overlaps with the ground-truth answer. It is computed using precision and recall over answer tokens:
\begin{equation}
\mathrm{Precision} =
\frac{\mathrm{overlapping\ tokens}}
{\mathrm{tokens\ in\ predicted\ answer}}
\end{equation}

\begin{equation}
\mathrm{Recall} =
\frac{\mathrm{overlapping\ tokens}}
{\mathrm{tokens\ in\ ground\ truth\ answer}}
\end{equation}

\begin{equation}
F_1 =
\frac{2 \cdot \mathrm{Precision} \cdot \mathrm{Recall}}
{\mathrm{Precision}+\mathrm{Recall}}
\end{equation}

In semantic communication, question answering metrics are useful when the receiver's goal is to infer the correct answer from transmitted semantic information. The reconstructed sentence does not need to match the original wording as long as the answer remains correct. An illustrative example of question answering exact match and F1 is provided in Appendix~\ref{apx_qa_example}.
% An illustrative example of question answering exact match and F1 is provided in Appendix~B.4.

\subsubsection{Retrieval Metrics: Recall@K, MRR, and NDCG}

Retrieval-based semantic communication evaluates whether the received semantic representation can retrieve the correct item from a candidate set. For example, a transmitted text query may be used to retrieve a document, image, answer, or database entry. In this setting, the receiver output is a ranked list of candidates, and success depends on whether the correct item appears near the top of the ranking.

Recall@K measures whether the correct item appears among the top $K$ retrieved candidates:
\begin{equation}
\mathrm{Recall@K} =
\frac{1}{N}
\sum_{i=1}^{N}
\mathbf{1}[\mathrm{rank}_i \leq K]
\end{equation}

where $\mathrm{rank}_i$ is the rank position of the correct item for query $i$.

Mean Reciprocal Rank, or MRR, measures how early the first correct item appears in the ranking:
\begin{equation}
\mathrm{MRR} =
\frac{1}{N}
\sum_{i=1}^{N}
\frac{1}{\mathrm{rank}_i}
\end{equation}

NDCG is useful when multiple retrieved items can have different relevance levels. It is computed as:
\begin{equation}
\mathrm{NDCG@K} =
\frac{\mathrm{DCG@K}}{\mathrm{IDCG@K}}
\end{equation}

\begin{equation}
\mathrm{DCG@K} =
\sum_{i=1}^{K}
\frac{\mathrm{rel}_i}{\log_2(i+1)}
\end{equation}

where $\mathrm{rel}_i$ is the relevance score of the item at rank $i$. IDCG@K is the ideal DCG value obtained from the best possible ranking.

In text-based semantic communication, retrieval metrics are useful when the goal is not to reconstruct the original text but to preserve enough semantic information to retrieve the correct content. This is especially relevant for semantic search, image-text retrieval, and retrieval-augmented applications. An illustrative example of Recall@K and MRR is provided in Appendix~\ref{apx_retrieval_metrics}.
% An illustrative example of Recall@K and MRR is provided in Appendix~B.5.

\subsubsection{Task Success Rate}

Task success rate measures whether the receiver successfully completes the final goal of the communication process. Unlike intermediate metrics such as intent accuracy or slot F1, task success rate evaluates the end result. For example, in a dialogue or command-execution system, success may mean that the correct booking is made, the correct API call is executed, or the correct action is selected.

Task success rate is computed as:
\begin{equation}
\mathrm{TaskSuccessRate} =
\frac{N_{\mathrm{success}}}{N}
\end{equation}

where $N_{\mathrm{success}}$ is the number of successfully completed tasks and $N$ is the total number of tasks.

This KPI is especially important for goal-oriented semantic communication because it directly reflects whether the communication system achieved its intended purpose. A system may have imperfect sentence reconstruction but still achieve high task success if the essential semantic information is preserved. Conversely, a system may generate fluent text but fail the task if it loses a critical intent or slot. An illustrative example of task success rate is provided in Appendix~\ref{apx_task_success_example}.
\begin{table*}[t]
\centering
\caption{Summary of Goal-Oriented Text Semantic Communication KPIs}
\label{tab:goal_oriented_text_kpis}
\small
\renewcommand{\arraystretch}{1.25}
\setlength{\tabcolsep}{3pt}
\begin{tabularx}{\textwidth}{p{0.16\textwidth} p{0.18\textwidth} p{0.20\textwidth} p{0.20\textwidth} X}
\toprule
\textbf{KPI} & \textbf{Task type} & \textbf{Main comparison} & \textbf{Formula idea} & \textbf{Main limitation} \tabularnewline
\midrule
Intent accuracy & Intent classification & $Y$ vs. $\hat{Y}$ & correct predictions / total samples & Too coarse when slot details matter \tabularnewline
Slot F1 & Slot filling / entity extraction & predicted slots vs. ground-truth slots & harmonic mean of precision and recall & Sensitive to boundary and label errors \tabularnewline
Frame accuracy / exact match & Semantic frame prediction & full predicted frame vs. full ground-truth frame & exact frame match rate & Very strict; no partial credit \tabularnewline
QA exact match / F1 & Question answering & predicted answer vs. ground-truth answer & exact answer match or token overlap & May miss semantic equivalence \tabularnewline
Recall@K / MRR / NDCG & Retrieval & ranked output vs. relevant item(s) & correct item rank or relevance-weighted ranking & Depends on candidate set and relevance labels \tabularnewline
Task success rate & End-to-end goal completion & final task outcome & successful tasks / total tasks & Requires clear definition of success \tabularnewline
\bottomrule
\end{tabularx}
\end{table*}

Table~\ref{tab:goal_oriented_text_kpis} summarizes these goal-oriented KPIs. Overall, they evaluate whether the receiver can accomplish the intended task rather than whether the reconstructed sentence matches the source text. These metrics are therefore more directly aligned with task-oriented semantic communication. However, most of them still require ground-truth task labels during evaluation. In practical deployment, the receiver may not have access to the original sentence or a gold reference, which motivates reference-free semantic communication KPIs.

\subsection{Reference-Free Text Semantic Communication KPIs}
\label{sec:reference_free_text_kpis}
Most semantic reconstruction metrics require the original sentence $S$ or a reference sentence to evaluate the reconstructed output $\hat{S}$. However, in practical semantic communication systems, the receiver may not have access to the original source message or a gold reference during deployment. This creates a reference-free evaluation setting, where the receiver must estimate whether the received message is semantically acceptable using only the received output, auxiliary semantic information, a transmitted semantic anchor, or a learned quality estimator.

Reference-free KPIs are important because they are closer to real communication scenarios. In these settings, success may be measured using quality estimation models, semantic anchors, threshold-based semantic success or outage, or fluency-based proxies such as perplexity. These KPIs do not fully replace reference-based evaluation, but they provide practical tools for monitoring semantic reliability when the original source is unavailable at the receiver.

\begin{table*}[t]
\centering
\caption{Summary of Reference-Free Text Semantic Communication KPIs}
\label{tab:reference_free_text_kpis}
\small
\renewcommand{\arraystretch}{1.25}
\setlength{\tabcolsep}{3pt}
\resizebox{\textwidth}{!}{%
\begin{tabular}{p{3.0cm} p{4.0cm} p{4.0cm} p{3.3cm} p{3.5cm}}
\toprule
\textbf{KPI} & \textbf{Main idea} & \textbf{Required at receiver} & \textbf{Strength} & \textbf{Limitation} \tabularnewline
\midrule
Quality estimation & A learned model predicts output quality without a gold reference & Received text, and sometimes source or side information & Practical for deployment & Model-dependent and may drift \tabularnewline
Anchor similarity & Compare the received output embedding with a transmitted semantic anchor & Decoded output and semantic anchor & Reference-free semantic check & Adds overhead \tabularnewline
Semantic success / outage & Decide success using a semantic similarity threshold & Similarity score and threshold & Telecom-friendly reliability measure & Threshold selection is subjective \tabularnewline
Perplexity & Use language model likelihood as a fluency proxy & Decoded text only & Detects garbled text & Fluency does not guarantee meaning \tabularnewline
\bottomrule
\end{tabular}%
}
\end{table*}

\subsubsection{Quality Estimation Metrics}

Quality estimation (QE) metrics use a learned model to predict the quality of a received or reconstructed sentence without requiring a human reference sentence. In text semantic communication, a QE model can be used to estimate whether the receiver-side output is meaningful, fluent, or useful for the intended communication goal.

A reference-free quality score can be written conceptually as:
\begin{equation}
q =
Q_{\theta}(\hat{S})
\end{equation}

or, when the source sentence or side information is available to the evaluator,
\begin{equation}
q =
Q_{\theta}(S,\hat{S})
\end{equation}

where $Q_{\theta}(\cdot)$ is a learned quality estimation model and $q$ is the predicted quality score.

In practice, the model may encode the received sentence $\hat{S}$ alone, or encode both $S$ and $\hat{S}$, and then use a regression head to output a scalar score:
\begin{equation}
q =
f_{\theta}(h)
\end{equation}

where $h$ is the hidden representation produced by the encoder.

In semantic communication, QE metrics are useful when no gold reference is available at the receiver. However, they are model-dependent and may be unreliable under domain shift, especially if the evaluator was trained on data that differs from the semantic communication task. An illustrative example of quality estimation is provided in Appendix~\ref{apx_quality_estimation}.

\subsubsection{Anchor-Embedding Similarity}

Anchor-embedding similarity provides a reference-free semantic check by transmitting a compact semantic anchor together with the message. The transmitter computes an embedding of the source sentence and sends this anchor, or a compressed version of it, to the receiver. The receiver then computes an embedding of the reconstructed sentence and compares it with the transmitted anchor.

Let the transmitted semantic anchor be
\begin{equation}
e =
B(S)
\end{equation}

where $B(\cdot)$ is a sentence encoder. After decoding, the receiver computes
\begin{equation}
\hat{e} =
B(\hat{S})
\end{equation}

The anchor similarity score is then
\begin{equation}
r =
\cos(e,\hat{e}) =
\frac{e \cdot \hat{e}}{|e||\hat{e}|}
\end{equation}

where $r$ is the anchor-based semantic similarity score.

In this setting, the receiver does not need the original sentence $S$ to evaluate semantic consistency. It only needs the transmitted anchor $e$ and the reconstructed sentence $\hat{S}$. This makes anchor similarity closer to a deployment setting than standard reference-based metrics. However, the anchor introduces additional communication overhead and may raise privacy or security concerns because embeddings can leak information about the source message. An illustrative example of anchor-embedding similarity is provided in Appendix~\ref{apx_anchor_embedding_example}.
% An illustrative example of anchor-embedding similarity is provided in Appendix~C.2.

\subsubsection{Semantic Threshold Success Rate and Semantic Outage}

A semantic similarity score can be converted into a binary success or failure decision by comparing it with a threshold $\tau$. For a given message, let $r_i$ denote the semantic similarity score for sample $i$. Semantic success can be defined as
\begin{equation}
Success_i =
\mathbf{1}[r_i \geq \tau]
\end{equation}

where $\tau$ is the chosen threshold and $\mathbf{1}[\cdot]$ is an indicator function.

For $N$ evaluated samples, the semantic success rate is
\begin{equation}
P_{\mathrm{success}} =
\frac{1}{N}
\sum_{i=1}^{N}
\mathbf{1}[r_i \geq \tau]
\end{equation}

The semantic outage probability is the probability that the semantic similarity falls below the threshold:
\begin{equation}
P_{\mathrm{outage}} =
\frac{1}{N}
\sum_{i=1}^{N}
\mathbf{1}[r_i < \tau]
\end{equation}

Equivalently,
\begin{equation}
P_{\mathrm{outage}} =
1-P_{\mathrm{success}}
\end{equation}

In semantic communication, semantic outage is especially useful because it resembles classical telecom reliability metrics such as packet error rate or outage probability, but it is defined in terms of meaning preservation rather than bit-level correctness. However, the threshold $\tau$ must be chosen carefully, and different embedding models or similarity functions may lead to different outage results. An illustrative example of semantic success and outage is provided in Appendix~\ref{apx_semantic_success_outage}.
% An illustrative example of semantic success and outage is provided in Appendix~C.3.

\subsubsection{Perplexity / Fluency Proxy}

Perplexity is a reference-free language-model-based metric that measures how likely a sentence is under a pretrained language model. It can be used as a weak proxy for fluency or grammatical naturalness of the reconstructed output.

For a reconstructed sentence $\hat{S}=(w_1,w_2,\ldots,w_T)$, a language model assigns a probability to each token conditioned on the previous tokens. The average negative log-likelihood is
\begin{equation}
L_{LM} =
-\frac{1}{T}
\sum_{t=1}^{T}
\log p(w_t \mid w_1,\ldots,w_{t-1})
\end{equation}

The perplexity is then
\begin{equation}
PPL(\hat{S}) =
\exp(L_{LM})
\end{equation}

Equivalently,
\begin{equation}
PPL(\hat{S}) =
\exp \left(
-\frac{1}{T}
\sum_{t=1}^{T}
\log p(w_t \mid w_1,\ldots,w_{t-1})
\right)
\end{equation}

A lower perplexity value indicates that the reconstructed sentence is more probable or more fluent according to the language model. In semantic communication, perplexity can help detect garbled or unnatural received text without requiring a reference sentence. However, fluency does not guarantee semantic correctness. A sentence can be fluent but still have the wrong meaning. Therefore, perplexity should be treated as a supporting proxy rather than a primary semantic KPI. An illustrative example of perplexity is provided in Appendix~\ref{apx_perplexity_example}.
% An illustrative example of perplexity is provided in Appendix~C.4.

Table~\ref{tab:reference_free_text_kpis} summarizes these reference-free KPIs. Overall, they are useful for deployment-oriented semantic communication because they do not require a full gold reference at the receiver. However, each reference-free metric involves a tradeoff. QE models depend on training data, anchors introduce overhead, semantic outage depends on threshold choice, and perplexity measures fluency rather than meaning. This motivates telecom-style semantic state recovery metrics, where the goal is to define semantic distortion or semantic error more directly at the level of the transmitted semantic representation.

\subsection{Telecom-Style Semantic State Recovery KPIs}
\label{sec:telecom_state_text_kpis}
Telecom-style semantic communication formulates the communication objective as the recovery of a semantic state rather than the exact reconstruction of the source sentence. In this setting, the source message $S$ is mapped to a semantic state or representation $W$, which is transmitted through the channel and recovered as $\hat{W}$. The semantic state may be a discrete intent label, a set of knowledge-graph triplets, a continuous embedding, or a posterior distribution over possible meanings. Therefore, evaluation focuses on how accurately $\hat{W}$ preserves $W$ under channel and resource constraints.

Unlike reconstruction metrics such as BLEU, ROUGE, or BERTScore, telecom-style KPIs are usually defined directly on the semantic state. They include semantic distortion, semantic error probability, semantic outage probability, and rate--semantic tradeoff curves. These metrics are useful because they connect semantic communication to classical communication concepts such as reliability, outage, distortion, and resource cost.

\subsubsection{Semantic Distortion}

Semantic distortion measures the difference between the original semantic state $W$ and the recovered semantic state $\hat{W}$. It generalizes classical distortion measures by applying the distance function to semantic representations rather than raw symbols or pixels. In general, semantic distortion can be written as
\begin{equation}
D_{\mathrm{sem}} =
\mathbb{E}[d(W,\hat{W})]
\end{equation}

where $d(W,\hat{W})$ is a semantic distortion function. For an empirical test set of $N$ samples, the expected distortion can be estimated as
\begin{equation}
\hat{D}*{\mathrm{sem}} =
\frac{1}{N}
\sum*{i=1}^{N}
d(W_i,\hat{W}_i)
\end{equation}

The choice of $d(\cdot)$ depends on how the semantic state is represented. If $W$ is a discrete label or intent, the distortion may be a binary mismatch. If $W$ is an embedding, the distortion may be Euclidean distance, cosine distance, or mean squared error. If $W$ is a set of semantic triplets, the distortion may count missing entities, wrong relations, or incorrect triples, possibly with importance weights. If $W$ is a probability distribution over meanings, the distortion may be based on divergence measures such as KL divergence or Wasserstein distance.

In text-based semantic communication, semantic distortion is useful when the system is designed to transmit meaning representations rather than reconstruct full sentences. However, the main challenge is that the distortion function must be explicitly defined and justified. An illustrative example of semantic distortion is provided in Appendix~\ref{apx_semantic_distortion}.

% An illustrative example of semantic distortion is provided in Appendix~D.1.

\subsubsection{Semantic Error Probability}

Semantic error probability measures the probability that the recovered semantic state does not match the original semantic state. For a discrete semantic state, such as an intent label, class label, or semantic symbol, it can be written as
\begin{equation}
P_e^{\mathrm{sem}} =
\Pr(\hat{W} \neq W)
\end{equation}

For an empirical test set, it can be estimated as
\begin{equation}
\hat{P}*e^{\mathrm{sem}} =
\frac{1}{N}
\sum*{i=1}^{N}
\mathbf{1}(\hat{W}_i \neq W_i)
\end{equation}

where the indicator function is an indicator function that equals 1 when the recovered semantic state is incorrect and 0 otherwise.

This KPI is closely related to classical symbol error probability, but the symbol being evaluated is semantic rather than purely syntactic or bit-level. For example, if the semantic state is the intent BookFlight, then a semantic error occurs when the receiver predicts a different intent such as CancelBooking. Semantic error probability is clean and interpretable for discrete semantic states, but it becomes less straightforward for continuous embeddings or structured semantic graphs, where partial semantic recovery may still be useful. An illustrative example of semantic error probability is provided in Appendix~\ref{apx_semantic_error_prob_example}.
% An illustrative example of semantic error probability is provided in Appendix~D.2.

\subsubsection{Semantic Outage Probability}

Semantic outage probability measures how often the recovered semantic state is unacceptable under a chosen semantic quality threshold. This is useful when semantic recovery is not all-or-nothing. For example, a reconstructed semantic embedding may be close enough to preserve meaning even if it is not identical to the original embedding.

If semantic quality is measured by a similarity score $Sim(W,\hat{W})$, then semantic outage can be defined as
\begin{equation}
P_{\mathrm{out}}^{\mathrm{sem}}(\tau) =
\Pr(Sim(W,\hat{W}) < \tau)
\end{equation}

where tau is the minimum acceptable semantic similarity threshold. For an empirical test set, this can be estimated as
\begin{equation}
\hat{P}*{\mathrm{out}}^{\mathrm{sem}}(\tau) =
\frac{1}{N}
\sum*{i=1}^{N}
\mathbf{1}(Sim(W_i,\hat{W}_i) < \tau)
\end{equation}

If semantic quality is measured by a distortion function, outage can instead be defined as
\begin{equation}
P_{\mathrm{out}}^{\mathrm{sem}}(\delta) =
\Pr(d(W,\hat{W}) > \delta)
\end{equation}

where delta is the maximum acceptable semantic distortion.

Semantic outage is especially useful for wireless semantic communication because it can be plotted against SNR, bandwidth, rate, or fading conditions. It provides a telecom-friendly way to express how often the communication system fails to preserve meaning at an acceptable level. However, the threshold must be chosen carefully, and results may vary depending on the semantic similarity or distortion measure used. An illustrative example of semantic outage probability is provided in Appendix~\ref{apx_semantic_outage_prob_example}.
% An illustrative example of semantic outage probability is provided in Appendix~D.3.

\subsubsection{Rate--Semantic Tradeoff}

A central motivation of semantic communication is to transmit less information while preserving task-relevant meaning. Therefore, telecom-style evaluation often considers the tradeoff between semantic quality and communication cost. A general rate--semantic tradeoff can be studied by plotting semantic quality or semantic distortion against a resource measure such as transmitted symbols, bits, bandwidth, latency, or energy.

One simple semantic efficiency score can be written as
\begin{equation}
\eta_{\mathrm{sem}} =
\frac{Q_{\mathrm{sem}}}{C_{\mathrm{res}}}
\end{equation}

where $Q_{\mathrm{sem}}$ is a semantic quality score, such as similarity, task utility, or success rate, and $C_{\mathrm{res}}$ is a communication resource cost, such as number of transmitted symbols, bandwidth, latency, or energy.

Alternatively, one can formulate an optimization problem:
\begin{equation}
\min C_{\mathrm{res}}
\quad
\mathrm{subject\ to}
\quad
Q_{\mathrm{sem}} \geq \tau
\end{equation}

or equivalently,
\begin{equation}
\min D_{\mathrm{sem}}
\quad
\mathrm{subject\ to}
\quad
C_{\mathrm{res}} \leq C_{\max}
\end{equation}

where tau is a required semantic quality threshold and $C_{\max}$ is a maximum resource budget.

In text-based semantic communication, this tradeoff is important because a system may transmit fewer semantic symbols, compressed embeddings, or selected semantic features while still preserving enough meaning for the receiver. This KPI family is useful for comparing systems under different SNRs, bandwidth constraints, or semantic compression rates. However, the resource cost and semantic quality score must be clearly defined to make comparisons meaningful. An illustrative example of rate--semantic tradeoff is provided in Appendix~\ref{apx_rate_tradeoff}.
% An illustrative example of rate--semantic tradeoff is provided in Appendix~D.4.

\begin{table*}[t]
\centering
\caption{Summary of Telecom-Style Text Semantic State Recovery KPIs}
\label{tab:telecom_style_text_kpis}
\small
\renewcommand{\arraystretch}{1.25}
\setlength{\tabcolsep}{3pt}
\resizebox{\textwidth}{!}{%
\begin{tabular}{p{3.0cm} p{5.0cm} p{4.75cm} p{4.75cm}}
\toprule
\textbf{KPI} & \textbf{Formula idea} & \textbf{Best suited for} & \textbf{Main limitation} \tabularnewline
\midrule
Semantic distortion & Expected semantic distance between $W$ and $\hat{W}$ & Embeddings, triplets, belief states & Requires defining semantic distance \tabularnewline
Semantic error probability & Probability of semantic mismatch & Discrete labels or symbols & Too strict for partial semantic recovery \tabularnewline
Semantic outage probability & Probability of unacceptable semantic quality & Wireless reliability analysis & Threshold selection is subjective \tabularnewline
Rate--semantic tradeoff & Semantic quality per resource cost, or constrained optimization & Resource-aware SemCom & Requires standardized cost and quality definitions \tabularnewline
\bottomrule
\end{tabular}%
}
\end{table*}

Table~\ref{tab:telecom_style_text_kpis} summarizes these telecom-style KPIs. Overall, they shift the evaluation focus from sentence-level reconstruction to semantic reliability under channel and resource constraints. They are especially important when semantic communication is treated as a communication-theoretic problem, where the main goal is to recover a task-relevant semantic state $W$ efficiently and reliably. After completing the text-based KPI categories, the next section extends the same evaluation-centered logic to image-based semantic communication KPIs.

\section{Image-Based Semantic Communication KPIs}
\label{sec:image_kpis}

Image-based semantic communication evaluates how effectively visual information is preserved, recovered, or utilized after transmission through a communication channel. Unlike text, where semantic information is often represented through words and sentences, images contain visual semantics associated with objects, spatial relationships, scene context, texture, and perceptual content. Consequently, no single KPI can adequately characterize image semantic communication performance. The appropriate evaluation metric depends on whether the communication objective is image reconstruction, perceptual quality preservation, semantic representation recovery, or downstream visual task accomplishment.

In the generic framework of Fig.~\ref{fig:semantic_comm_pipeline}, the source $S$ corresponds to an image, denoted here by $I$, and the reconstructed output $\hat{S}$ corresponds to the reconstructed image $\hat{I}$. Let $Z = F(I)$ denote a visual semantic representation extracted from $I$ using an encoder $F(\cdot)$, and let $\hat{Z} = F(\hat{I})$ denote the corresponding representation at the receiver. For task-oriented image semantic communication, let $Y$ denote the ground-truth visual task output and $\hat{Y}$ denote the receiver-side predicted output. Under this notation, pixel-fidelity and perceptual metrics compare $I$ and $\hat{I}$, representation-level semantic metrics compare $Z$ and $\hat{Z}$, and task-oriented metrics compare $Y$ and $\hat{Y}$.

Table~\ref{tab:image_kpi_categories} summarizes the main image KPI families considered in this section. The organization follows the same evaluation logic used for text-based semantic communication: low-level reconstruction, perceptual similarity, semantic representation similarity, downstream task utility, distribution-level realism, and channel-aware semantic reliability.

\begin{table*}[t]
\centering
\caption{KPI categories for image-based semantic communication}
\label{tab:image_kpi_categories}
\begin{tabular}{p{0.22\textwidth} p{0.45\textwidth} p{0.27\textwidth}}
\toprule
\textbf{Image KPI category} & \textbf{Core question} & \textbf{Representative KPIs} \\
\midrule
Pixel fidelity & Is the received image numerically close to the source image? & MSE, PSNR \\
Structural/perceptual quality & Does the received image preserve visual structure and perceptual quality? & SSIM, MS-SSIM, LPIPS \\
Visual semantic similarity & Are high-level visual features preserved? & Feature cosine similarity, CLIP similarity \\
Downstream visual task utility & Can the receiver complete the intended visual task? & Accuracy, mAP, IoU, Dice, Recall@K \\
Distribution-level realism & Are reconstructed or generated images realistic at the dataset level? & FID, KID, Inception Score \\
Channel-aware semantic reliability & Is semantic visual quality maintained under channel/resource constraints? & Semantic outage, semantic distortion, task outage vs. SNR/rate \\
\bottomrule
\end{tabular}
\end{table*}

\subsection{Pixel-Fidelity Metrics}

Pixel-fidelity metrics evaluate how close the reconstructed image $\hat{I}$ is to the original image $I$ at the numerical pixel level. These metrics are widely used in traditional image compression, image restoration, wireless image transmission, and deep JSCC systems because they are simple, objective, and easy to compute \cite{Wallace1992JPEG} \cite{Bourtsoulatze2019DeepJSCC}. However, pixel-fidelity metrics do not directly measure semantic preservation. A reconstruction can have high pixel fidelity while losing task-relevant details, and another reconstruction can have lower pixel fidelity while preserving the main semantic content.

\subsubsection{Mean Squared Error (MSE)}

Mean squared error measures the average squared pixel difference between the original image $I$ and the reconstructed image $\hat{I}$. For an image of height $H$, width $W$, and $C$ channels, MSE is defined as:
\begin{equation}
\mathrm{MSE}(I,\hat{I})
=
\frac{1}{HWC}
\sum_{i=1}^{H}
\sum_{j=1}^{W}
\sum_{c=1}^{C}
\left(I_{ijc} - \hat{I}_{ijc}\right)^2 .
\label{eq:mse_image}
\end{equation}

A lower MSE indicates that the reconstructed image is numerically closer to the original image. In image-based semantic communication, MSE is useful when the system objective is close visual reconstruction. However, it treats all pixels equally and does not know which regions are semantically important. For example, a small error on a traffic light, lesion boundary, or object edge may be much more important than a larger error in a flat background region, but MSE does not distinguish between these cases. A numerical example is provided in Appendix~\ref{apx_sample_level_results}.
% A numerical example is provided in Appendix~E.1.

\subsubsection{Peak Signal-to-Noise Ratio (PSNR)}

Peak signal-to-noise ratio is derived from MSE and expresses reconstruction quality on a logarithmic scale. For an image with maximum possible pixel value $\mathrm{MAX}_I$, PSNR is defined as:
\begin{equation}
\mathrm{PSNR}(I,\hat{I})
=
10 \log_{10}
\left(
\frac{\mathrm{MAX}_I^2}{\mathrm{MSE}(I,\hat{I})}
\right).
\label{eq:psnr_image}
\end{equation}

A higher PSNR indicates better pixel-level reconstruction quality. PSNR is commonly reported in wireless image transmission and deep JSCC studies because it enables simple comparison across channel conditions such as SNR, bandwidth ratio, and compression rate \cite{Bourtsoulatze2019DeepJSCC} \cite{Dai2022Nonlinear}. However, PSNR remains a pixel-level metric. It often correlates poorly with human perception and semantic task success, especially when reconstructions are blurry, over-smoothed, or perceptually plausible but not pixel-aligned. Table~\ref{tab:pixel_fidelity_metrics} summarizes these pixel-fidelity metrics. A numerical example is provided in Appendix~\ref{apx_sample_level_results}.
% A numerical example is provided in Appendix~E.2.

\begin{table*}[t]
\centering
\caption{Summary of pixel-fidelity metrics for image-based semantic communication}
\label{tab:pixel_fidelity_metrics}
\begin{tabular}{p{0.14\textwidth} p{0.24\textwidth} p{0.12\textwidth} p{0.16\textwidth} p{0.28\textwidth}}
\toprule
\textbf{Metric} & \textbf{Type} & \textbf{Range} & \textbf{Higher is better?} & \textbf{Main limitation} \\
\midrule
MSE & Pixel-level error & $[0,\infty)$ & No & Treats all pixels equally; not semantic \\
PSNR & Logarithmic pixel fidelity & $[0,\infty)$ dB & Yes & Weak correlation with perception and task meaning \\
\bottomrule
\end{tabular}
\end{table*}

\subsection{Structural and Perceptual Quality Metrics}

Pixel-fidelity metrics such as MSE and PSNR measure numerical closeness but do not fully account for how humans perceive visual quality. Two images may have similar PSNR values while appearing very different to a human observer, because human vision is sensitive to structural patterns, edges, contrast, and texture \cite{Wang2004SSIM}. Structural and perceptual metrics address this limitation by evaluating image quality in ways that are more aligned with visual perception.

In image-based semantic communication, structural and perceptual metrics are important because they can detect degradation that PSNR may miss, such as blurring of object boundaries, loss of texture, or distortion of spatial structure \cite{Bourtsoulatze2019DeepJSCC}. However, they still evaluate visual quality rather than full semantic content. A high SSIM or low LPIPS score does not guarantee that task-relevant objects, attributes, or spatial relations are preserved.

\subsubsection{Structural Similarity Index Measure (SSIM)}

The structural similarity index measure evaluates perceptual similarity between the original image $I$ and the reconstructed image $\hat{I}$ by comparing local luminance, contrast, and structure \cite{Wang2004SSIM}. For two image patches $p$ and $\hat{p}$ extracted from $I$ and $\hat{I}$, respectively, SSIM can be written as:
\begin{equation}
\mathrm{SSIM}(p,\hat{p})
=
l(p,\hat{p}) \cdot c(p,\hat{p}) \cdot \mathrm{str}(p,\hat{p}),
\label{eq:ssim_components}
\end{equation}
where $l(p,\hat{p})$, $c(p,\hat{p})$, and $\mathrm{str}(p,\hat{p})$ denote the luminance comparison, contrast comparison, and structure comparison terms, respectively. These components are defined as:
\begin{equation}
l(p,\hat{p})
=
\frac{2\mu_p \mu_{\hat{p}} + C_1}
{\mu_p^2 + \mu_{\hat{p}}^2 + C_1},
\label{eq:ssim_luminance}
\end{equation}

\begin{equation}
c(p,\hat{p})
=
\frac{2\sigma_p \sigma_{\hat{p}} + C_2}
{\sigma_p^2 + \sigma_{\hat{p}}^2 + C_2},
\label{eq:ssim_contrast}
\end{equation}

\begin{equation}
\mathrm{str}(p,\hat{p})
=
\frac{\sigma_{p\hat{p}} + C_3}
{\sigma_p \sigma_{\hat{p}} + C_3}.
\label{eq:ssim_structure}
\end{equation}

Here, $\mu_p$ and $\mu_{\hat{p}}$ are the mean values of patches $p$ and $\hat{p}$; $\sigma_p$ and $\sigma_{\hat{p}}$ are their standard deviations; $\sigma_{p\hat{p}}$ is their cross-covariance; and $C_1$, $C_2$, and $C_3$ are small stabilizing constants. The structure comparison term $\mathrm{str}(p,\hat{p})$ measures how similar the local structural patterns of the two patches are after accounting for luminance and contrast. The overall SSIM score is usually computed by averaging local patch-level SSIM values across the image. SSIM is typically reported in the range $[0,1]$, where 1 indicates perfect structural similarity, although the theoretical range can extend to $[-1,1]$.

In semantic communication, SSIM is useful because it is more sensitive to structural degradation than PSNR. For example, blurred object boundaries, smeared edges, and local contrast loss can reduce SSIM even when pixel-wise error is moderate. However, SSIM is still not a semantic metric. It does not explicitly know whether the degraded structure belongs to an important object, a background region, or a task-critical visual cue. An illustrative example is provided in Appendix~\ref{apx_sample_level_results}.
% An illustrative example is provided in Appendix~E.3.

\subsubsection{Multi-Scale Structural Similarity Index Measure (MS-SSIM)}

Multi-scale structural similarity extends SSIM by evaluating structural similarity at multiple image resolutions \cite{Wang2003MSSSIM}. Visual information exists at different spatial scales: global scene layout appears at coarse scales, while local edges, textures, and small objects appear at fine scales. MS-SSIM accounts for this by progressively downsampling the image and computing structural comparisons at several scales.

\begin{equation}
\mathrm{MS\text{-}SSIM}(I,\hat{I})
=
l_M(I,\hat{I})^{\alpha_M}
\cdot
\prod_{j=1}^{M}
\left[
c_j(I,\hat{I})^{\beta_j}
\cdot
s_j(I,\hat{I})^{\gamma_j}
\right].
\label{eq:ms_ssim}
\end{equation}

In this expression, $l_M$ is the luminance comparison at the coarsest or selected scale, while $c_j$ and $s_j$ are contrast and structure comparisons at scale $j$. The parameters $\alpha_M$, $\beta_j$, and $\gamma_j$ control the contribution of each component and scale. MS-SSIM usually ranges from 0 to 1, where 1 indicates perfect multi-scale structural similarity.

In image-based semantic communication, MS-SSIM is often preferred over single-scale SSIM because it captures both coarse scene structure and fine detail \cite{Bourtsoulatze2019DeepJSCC} \cite{Kurka2020DeepJSCC}. At coarse scales, it can capture whether the overall layout is preserved; at fine scales, it can capture edges, textures, and small objects. Nevertheless, MS-SSIM remains a structural/perceptual metric and does not directly evaluate task-level semantic correctness. An illustrative example is provided in Appendix~\ref{apx_sample_level_results}.
% An illustrative example is provided in Appendix~E.4.

\subsubsection{Learned Perceptual Image Patch Similarity (LPIPS)}

Learned perceptual image patch similarity measures perceptual distance using deep neural network features rather than hand-crafted image statistics \cite{Zhang2018LPIPS}. A pretrained network, such as AlexNet, VGG, or SqueezeNet, extracts feature maps from both $I$ and $\hat{I}$. LPIPS then computes a weighted distance between these feature activations across multiple layers.

\begin{equation}
\mathrm{LPIPS}(I,\hat{I})
=
\sum_l
\frac{1}{H_l W_l}
\sum_{h,v}
\left\|
w_l \odot
\left(
F^l(I)_{hv} - F^l(\hat{I})_{hv}
\right)
\right\|_2^2 .
\label{eq:lpips}
\end{equation}

Here, $F^l(\cdot)$ denotes the feature map at layer $l$, $H_l$ and $W_l$ are the spatial dimensions of that feature map, $w_l$ is a learned channel-wise weight vector, and $\odot$ denotes element-wise multiplication. A lower LPIPS score indicates higher perceptual similarity. LPIPS is particularly useful for generative or learned image transmission systems, where images may look perceptually realistic even when they are not pixel-aligned with the original \cite{Kurka2020DeepJSCC}.

The main limitation of LPIPS is that it is model-dependent. Scores can vary depending on the chosen backbone and calibration. In addition, LPIPS measures perceptual similarity rather than task-specific semantic utility. A reconstruction may receive a low LPIPS score but still fail a downstream task if a small task-critical visual detail is distorted. Table~\ref{tab:structural_perceptual_metrics} summarizes these structural and perceptual metrics. An illustrative example is provided in Appendix~\ref{apx_sample_level_results}.
% An illustrative example is provided in Appendix~E.5.

\begin{table*}[t]
\centering
\caption{Summary of structural and perceptual quality metrics for image-based semantic communication}
\label{tab:structural_perceptual_metrics}
\setlength{\tabcolsep}{3pt}% tighten to avoid overfull table
\begin{tabular}{p{0.13\textwidth} p{0.22\textwidth} p{0.26\textwidth} p{0.13\textwidth} p{0.18\textwidth}}
\toprule
\textbf{Metric} & \textbf{Type} & \textbf{Range} & \textbf{Higher is better?} & \textbf{Main limitation} \\
\midrule
SSIM & Structural similarity & Typically $[0,1]$, theoretically $[-1,1]$ & Yes & Single-scale; not explicitly semantic \\
MS-SSIM & Multi-scale structural similarity & $[0,1]$ & Yes & Still not fully semantic \\
LPIPS & Learned perceptual distance & Non-negative, implementation-dependent & No & Model-dependent; may not reflect task meaning \\
\bottomrule
\end{tabular}
\end{table*}

\subsection{Visual Semantic Similarity Metrics}

Visual semantic similarity metrics evaluate whether the high-level visual content of the original image is preserved after transmission. Instead of directly comparing pixels or local structural statistics, these metrics compare images in a learned feature space. This is closer to the goal of semantic communication because semantic content is often represented by objects, scene context, attributes, and relationships rather than exact pixel values.

Let $F(\cdot)$ be a visual encoder, such as a CNN, Vision Transformer, or multimodal encoder. The original and reconstructed images can be mapped into feature vectors $Z = F(I)$ and $\hat{Z} = F(\hat{I})$. Semantic similarity can then be evaluated by comparing $Z$ and $\hat{Z}$. The choice of encoder is important because different encoders preserve different types of visual meaning. Therefore, semantic communication papers should report the encoder architecture, checkpoint, layer, pooling method, and normalization strategy \cite{Kornblith2019Similarity}.

\subsubsection{Feature Cosine Similarity}

Feature cosine similarity compares the direction of learned visual representations extracted from the original and reconstructed images. It can be written as:
\begin{equation}
\mathrm{Sim}_{\mathrm{feat}}(I,\hat{I})
=
\cos\left(F(I),F(\hat{I})\right)
=
\frac{F(I) \cdot F(\hat{I})}
{\|F(I)\| \|F(\hat{I})\|}.
\label{eq:feature_cosine_similarity}
\end{equation}

This KPI is useful when the goal is to preserve high-level visual meaning rather than exact pixel values. For example, a slightly blurry image of a cyclist may have low PSNR but high feature similarity if a pretrained visual encoder still recognizes the same object and scene. Conversely, an image with good pixel fidelity in background regions but a missing object may receive a lower semantic feature similarity. However, this metric depends strongly on the encoder and may ignore fine spatial errors that are important for some applications. 
% An illustrative example is provided in Appendix~E.6.

\subsubsection{CLIP-Based Image Similarity}

CLIP-based similarity uses a vision-language model to evaluate visual semantic preservation in a multimodal feature space \cite{Radford2021CLIP}. The CLIP image encoder $E_I(\cdot)$ maps images into an embedding space aligned with text embeddings. Image-image semantic similarity can be computed as:
\begin{equation}
\mathrm{Sim}_{\mathrm{CLIP,img}}(I,\hat{I})
=
\cos\left(E_I(I),E_I(\hat{I})\right).
\label{eq:clip_image_similarity}
\end{equation}

When a textual description, caption, or semantic goal $T$ is available, CLIP can also evaluate image-text alignment:
\begin{equation}
\mathrm{Sim}_{\mathrm{CLIP,txt}}(T,\hat{I})
=
\cos\left(E_T(T),E_I(\hat{I})\right).
\label{eq:clip_text_similarity}
\end{equation}

This is especially useful for semantic communication systems where the receiver does not need to reproduce the exact image but must preserve the meaning described by a caption, prompt, or task instruction. For example, if the transmitted semantic goal is ``a cyclist waiting at an intersection,'' CLIP image-text similarity can evaluate whether the reconstructed image remains aligned with that description. However, CLIP similarity may miss fine-grained spatial details, small objects, medical abnormalities, or domain-specific semantics not well represented in CLIP training data. Table~\ref{tab:visual_semantic_similarity_metrics} summarizes these visual semantic similarity metrics. An illustrative example is provided in Appendix~\ref{apx_sample_level_results}.
% An illustrative example is provided in Appendix~E.7.

\begin{table*}[t]
\centering
\caption{Summary of visual semantic similarity metrics for image-based semantic communication}
\label{tab:visual_semantic_similarity_metrics}
\begin{tabular}{p{0.22\textwidth} p{0.24\textwidth} p{0.24\textwidth} p{0.24\textwidth}}
\toprule
\textbf{Metric} & \textbf{Main comparison} & \textbf{Strength} & \textbf{Main limitation} \\
\midrule
Feature cosine similarity & $F(I)$ vs. $F(\hat{I})$ & Captures learned high-level visual similarity & Encoder-dependent; may miss task-critical details \\
CLIP image-image similarity & $E_I(I)$ vs. $E_I(\hat{I})$ & Evaluates semantic similarity in a vision-language space & May miss fine spatial/domain-specific details \\
CLIP image-text similarity & $E_T(T)$ vs. $E_I(\hat{I})$ & Useful when semantic goal is textual or caption-based & Depends on caption/prompt and CLIP domain coverage \\
\bottomrule
\end{tabular}
\end{table*}

\subsection{Downstream Visual Task Utility Metrics}

In goal-oriented image semantic communication, the receiver may not need to reconstruct the image at all. Instead, it may need to classify the scene, detect objects, segment regions, retrieve relevant images, or support a decision. In this setting, the most meaningful KPI is often the downstream task performance after transmission. These metrics compare the ground-truth task output $Y$ with the receiver-side prediction $\hat{Y}$, rather than directly comparing $I$ and $\hat{I}$.

Task utility metrics are important because they reveal whether the transmitted visual information is sufficient for the intended application. A received image can have low PSNR but still support correct classification, or it can look visually acceptable while failing object detection or segmentation. Therefore, for task-oriented semantic communication, task metrics should be reported together with reconstruction or perceptual metrics \cite{Shao2022TaskOriented}.

\subsubsection{Classification Accuracy}

Classification accuracy measures whether the receiver predicts the correct class label. For $N$ test images, it is defined as:
\begin{equation}
\mathrm{Accuracy}
=
\frac{N_{\mathrm{correct}}}{N}.
\label{eq:classification_accuracy}
\end{equation}

Classification accuracy is useful when the communication objective is to preserve enough semantic information for image recognition. For example, if the original image contains a bicycle and the receiver classifies the reconstructed image as bicycle, the semantic task may be successful even if the image is not perfectly reconstructed. However, accuracy is coarse and does not capture localization, spatial relations, or partial semantic errors. 
% An illustrative example is provided in Appendix~E.8.

\subsubsection{Object Detection Mean Average Precision (mAP)}

Object detection evaluates whether the receiver can correctly localize and classify objects. Detection performance is commonly measured using mean average precision, which averages precision-recall performance across object classes and intersection-over-union thresholds \cite{Lin2014COCO}. A predicted bounding box is usually considered correct if its IoU with the ground-truth box exceeds a threshold and the predicted class is correct.

\begin{equation}
\mathrm{IoU}_{\mathrm{box}}
=
\frac{\mathrm{Area}(B \cap \hat{B})}
{\mathrm{Area}(B \cup \hat{B})}.
\label{eq:box_iou}
\end{equation}

\begin{equation}
\mathrm{mAP}
=
\frac{1}{C}
\sum_c \mathrm{AP}_c .
\label{eq:map}
\end{equation}

In semantic communication, mAP is useful when preserving object-level visual meaning is more important than reconstructing every pixel. For instance, an autonomous driving or surveillance system may care more about detecting pedestrians, vehicles, or traffic signs than about image texture fidelity. However, mAP depends on the detector, dataset, class definitions, and IoU threshold. 
% An illustrative example is provided in Appendix~E.9.

\subsubsection{Segmentation Metrics: IoU and Dice}

Segmentation metrics evaluate whether the receiver preserves pixel-level semantic regions such as objects, organs, lesions, roads, or buildings. Let $Y$ denote the ground-truth segmentation mask and $\hat{Y}$ denote the predicted mask. Intersection over union and Dice score are defined as:
\begin{equation}
\mathrm{IoU}
=
\frac{|Y \cap \hat{Y}|}
{|Y \cup \hat{Y}|},
\label{eq:segmentation_iou}
\end{equation}

\begin{equation}
\mathrm{Dice}
=
\frac{2|Y \cap \hat{Y}|}
{|Y| + |\hat{Y}|}.
\label{eq:dice}
\end{equation}

These metrics are important for semantic communication applications in medical imaging, remote sensing, autonomous driving, and industrial inspection. In medical image communication, for example, a reconstructed CT image may look visually acceptable, but if a lesion or organ boundary is lost, the semantic task fails. Dice and IoU directly evaluate whether the task-relevant spatial regions are preserved. However, these metrics require ground-truth segmentation masks and can be sensitive to small boundary errors, especially for small objects. 
% An illustrative example is provided in Appendix~E.10.

\subsubsection{Image Retrieval Metrics: Recall@K and MRR}

Image retrieval metrics evaluate whether the received visual representation can retrieve the correct image, caption, class, or database entry from a candidate set. Recall@K measures whether the correct item appears among the top $K$ retrieved candidates:

\begin{equation}
\mathrm{Recall@K}
=
\frac{1}{N}
\sum_i
\mathbf{1}
\left[
\mathrm{rank}_i \leq K
\right].
\label{eq:recall_at_k}
\end{equation}

Mean reciprocal rank measures how early the first correct item appears:
\begin{equation}
\mathrm{MRR}
=
\frac{1}{N}
\sum_i
\frac{1}{\mathrm{rank}_i}.
\label{eq:mrr}
\end{equation}

Retrieval metrics are useful for image-text semantic communication, visual search, and multimodal retrieval. They are especially relevant when the receiver output is a ranked list rather than a reconstructed image or class label. However, these metrics depend on the candidate set, relevance labels, and embedding model. Table~\ref{tab:downstream_visual_task_metrics} summarizes these downstream task utility metrics.
% An illustrative example is provided in Appendix~E.11.

\begin{table*}[t]
\centering
\caption{Summary of downstream visual task utility metrics}
\label{tab:downstream_visual_task_metrics}
\begin{tabular}{p{0.18\textwidth} p{0.18\textwidth} p{0.28\textwidth} p{0.28\textwidth}}
\toprule
\textbf{KPI} & \textbf{Task type} & \textbf{Main comparison} & \textbf{Main limitation} \\
\midrule
Accuracy & Classification & $Y$ vs. $\hat{Y}$ class labels & Too coarse for localization or structure \\
mAP & Object detection & Predicted boxes/classes vs. ground truth & Depends on detector and IoU thresholds \\
IoU / Dice & Segmentation & Predicted mask vs. ground-truth mask & Requires masks; sensitive for small regions \\
Recall@K / MRR & Retrieval & Ranked output vs. relevant item(s) & Depends on candidate set and relevance labels \\
\bottomrule
\end{tabular}
\end{table*}

\subsection{Distribution-Level Realism Metrics}

Distribution-level realism metrics evaluate whether a set of reconstructed or generated images follows the same distribution as real images. These metrics are especially relevant for generative semantic communication, diffusion-based semantic communication, and image synthesis systems, where the receiver may generate a visually plausible image rather than reconstruct the exact source image. Unlike PSNR, SSIM, or LPIPS, distribution-level metrics are computed over image sets rather than individual samples.

\subsubsection{Fréchet Inception Distance (FID)}

Fréchet Inception Distance compares the distributions of real and generated images in a deep feature space, usually using activations from an Inception network \cite{Heusel2017FID}. If the real-image feature distribution has mean $\mu_r$ and covariance $\Sigma_r$, and the generated-image feature distribution has mean $\mu_g$ and covariance $\Sigma_g$, FID is defined as:
\begin{equation}
\mathrm{FID}
=
\left\|
\mu_r - \mu_g
\right\|_2^2
+
\mathrm{Tr}
\left(
\Sigma_r + \Sigma_g
-
2
\left(
\Sigma_r \Sigma_g
\right)^{1/2}
\right).
\label{eq:fid}
\end{equation}

A lower FID indicates that the generated or reconstructed image distribution is closer to the real-image distribution. In semantic communication, FID is useful when evaluating whether the receiver produces realistic images under channel constraints. However, FID is dataset-level, not sample-level. It cannot tell whether a particular received image preserves the semantic content of its corresponding source image. It is also sensitive to dataset size and feature extractor choice.

\subsubsection{Kernel Inception Distance (KID)}

Kernel Inception Distance also compares real and generated image distributions in a deep feature space, but it uses maximum mean discrepancy with a polynomial kernel \cite{Binkowski2018KID}. It can be written conceptually as:
\begin{equation}
\mathrm{KID}
=
\mathrm{MMD}^2
\left(
\{\phi(I_{\mathrm{real}})\},
\{\phi(\hat{I}_{\mathrm{gen}})\}
\right),
\label{eq:kid}
\end{equation}
where $\phi(\cdot)$ is a feature extractor, often based on Inception activations. KID has an unbiased estimator, which can make it useful when evaluating smaller datasets. In semantic communication, KID can complement FID for distribution-level assessment. However, like FID, it does not directly evaluate whether each reconstructed image preserves the semantic content of its corresponding source.

\subsubsection{Inception Score (IS)}

Inception Score evaluates whether generated images are both classifiable and diverse \cite{Salimans2016InceptionScore}. It uses a pretrained classifier to compute the conditional label distribution $p(y|x)$ and the marginal label distribution $p(y)$. It is commonly written as:
\begin{equation}
\mathrm{IS}
=
\exp
\left(
\mathbb{E}_x
\left[
\mathrm{KL}
\left(
p(y|x) \| p(y)
\right)
\right]
\right).
\label{eq:inception_score}
\end{equation}

A higher Inception Score suggests that individual images produce confident class predictions while the full generated set covers diverse classes. However, IS has important limitations: it does not compare generated images with real images directly, depends on the pretrained classifier, and may not align with semantic preservation of a source image. Therefore, it should be used cautiously in semantic communication and usually together with FID, KID, perceptual metrics, and task metrics. Table~\ref{tab:distribution_level_realism_metrics} summarizes these distribution-level realism metrics.

\begin{table*}[t]
\centering
\caption{Summary of distribution-level realism metrics}
\label{tab:distribution_level_realism_metrics}
\begin{tabular}{p{0.18\textwidth} p{0.30\textwidth} p{0.18\textwidth} p{0.28\textwidth}}
\toprule
\textbf{Metric} & \textbf{Evaluation level} & \textbf{Higher is better?} & \textbf{Main limitation} \\
\midrule
FID & Dataset-level distribution similarity & No & Not sample-specific; sensitive to features/dataset size \\
KID & Dataset-level distribution similarity & No & Still not sample-specific semantic preservation \\
Inception Score & Dataset-level realism/diversity & Yes & Does not compare to real images directly \\
\bottomrule
\end{tabular}
\end{table*}

\subsection{Channel-Aware Semantic Reliability Metrics}

A central goal of semantic communication is to preserve task-relevant meaning under channel and resource constraints. Therefore, image semantic communication should not only report image quality, but also how semantic quality changes with SNR, bandwidth ratio, transmitted symbols, latency, or energy. Channel-aware semantic reliability metrics connect image semantic communication to classical communication concepts such as distortion, outage, and rate-reliability tradeoff.

\subsubsection{Visual Semantic Distortion}

Visual semantic distortion measures the difference between the original visual semantic representation $Z$ and the recovered representation $\hat{Z}$. A general form is:
\begin{equation}
D_{\mathrm{sem}}
=
\mathbb{E}
\left[
d(Z,\hat{Z})
\right].
\label{eq:semantic_distortion_expected}
\end{equation}

For an empirical test set, it can be estimated as:
\begin{equation}
\hat{D}_{\mathrm{sem}}
=
\frac{1}{N}
\sum_i
d(Z_i,\hat{Z}_i).
\label{eq:semantic_distortion_empirical}
\end{equation}

The distance function $d(\cdot)$ depends on the semantic representation. If $Z$ is an embedding, $d$ may be cosine distance or Euclidean distance. If $Z$ is a set of objects or relations, $d$ may count missing objects, wrong attributes, or incorrect spatial relations. If $Z$ is a segmentation mask or detection output, $d$ may be based on task-specific errors such as $1 - \mathrm{Dice}$ or $1 - \mathrm{mAP}$. This metric is flexible, but the main challenge is defining and justifying the semantic distance function. 
% An illustrative example is provided in Appendix~E.15.

\subsubsection{Visual Semantic Outage Probability}

Semantic outage probability measures how often the visual semantic quality falls below an acceptable threshold. If semantic quality is measured by a similarity score $\mathrm{Sim}(Z,\hat{Z})$, then outage can be defined as:
\begin{equation}
P_{\mathrm{out}}^{\mathrm{sem}}(\tau)
=
\Pr
\left(
\mathrm{Sim}(Z,\hat{Z}) < \tau
\right).
\label{eq:semantic_outage_similarity}
\end{equation}

For an empirical test set:
\begin{equation}
\hat{P}_{\mathrm{out}}^{\mathrm{sem}}(\tau)
=
\frac{1}{N}
\sum_i
\mathbf{1}
\left[
\mathrm{Sim}(Z_i,\hat{Z}_i) < \tau
\right].
\label{eq:semantic_outage_empirical}
\end{equation}

Alternatively, if semantic quality is measured by a distortion function, outage can be defined as:
\begin{equation}
P_{\mathrm{out}}^{\mathrm{sem}}(\delta)
=
\Pr
\left(
d(Z,\hat{Z}) > \delta
\right).
\label{eq:semantic_outage_distortion}
\end{equation}
This KPI is useful because it provides a telecom-style reliability measure for semantic image transmission. It can be plotted against SNR, compression ratio, bandwidth, latency, or number of transmitted semantic symbols. However, the threshold $\tau$ or $\delta$ must be chosen carefully and reported clearly. An illustrative example is provided in Appendix~\ref{apx_image_success_outage}.
% This KPI is useful because it provides a telecom-style reliability measure for semantic image transmission. It can be plotted against SNR, compression ratio, bandwidth, latency, or number of transmitted semantic symbols. However, the threshold $\tau$ or $\delta$ must be chosen carefully and reported clearly. An illustrative example is provided in Appendix~\ref{apx_semantic_success_outage}.
% An illustrative example is provided in Appendix~E.16.

\subsubsection{Task Outage Under Channel Constraints}

For task-oriented image semantic communication, outage can be defined directly using a downstream task KPI. For example, if the receiver performs segmentation and the minimum acceptable Dice score is $\tau_{\mathrm{Dice}}$, task outage can be defined as:
\begin{equation}
P_{\mathrm{out}}^{\mathrm{task}}(\tau_{\mathrm{Dice}})
=
\Pr
\left(
\mathrm{Dice}(Y,\hat{Y}) < \tau_{\mathrm{Dice}}
\right).
\label{eq:task_outage_dice}
\end{equation}

Similarly, for object detection, outage may be defined when mAP falls below a required threshold, and for classification, outage may be defined when the predicted class is incorrect. This type of KPI is highly relevant for practical systems because it measures failure in terms of the final application goal rather than image appearance. For example, a medical image communication system may declare outage when the liver lesion Dice score falls below a clinically acceptable threshold, even if the reconstructed CT image appears visually smooth. 
% An illustrative example is provided in Appendix~E.17.

\subsubsection{Rate-Semantic and SNR-Semantic Tradeoff Curves}

Semantic communication systems should report semantic quality together with communication cost. A simple semantic efficiency score can be written as:
\begin{equation}
\eta_{\mathrm{sem}}
=
\frac{Q_{\mathrm{sem}}}{C_{\mathrm{res}}},
\label{eq:semantic_efficiency}
\end{equation}
where $Q_{\mathrm{sem}}$ is a semantic quality score, such as feature similarity, task accuracy, Dice, or semantic success rate, and $C_{\mathrm{res}}$ is a resource cost, such as number of transmitted symbols, bandwidth, latency, or energy. Alternatively, one can formulate a constrained optimization problem:
\begin{equation}
\min C_{\mathrm{res}}
\quad
\mathrm{subject\ to}
\quad
Q_{\mathrm{sem}} \geq \tau,
\label{eq:semantic_cost_minimization}
\end{equation}

\begin{equation}
\min D_{\mathrm{sem}}
\quad
\mathrm{subject\ to}
\quad
C_{\mathrm{res}} \leq C_{\max}.
\label{eq:semantic_distortion_minimization}
\end{equation}

In image-based semantic communication, these tradeoff curves are important because a system may transmit fewer pixels, features, tokens, or semantic descriptors while preserving enough visual meaning for the receiver. Reporting semantic quality versus SNR, bandwidth ratio, compression rate, or latency makes the KPI communication-aware rather than only image-quality-aware. Table~\ref{tab:channel_aware_semantic_reliability_metrics} summarizes these channel-aware reliability metrics. An illustrative example is provided in Appendix~\ref{apx_image_success_outage}.
% An illustrative example is provided in Appendix~E.18.

\begin{table*}[t]
\centering
\caption{Summary of channel-aware semantic reliability metrics for image semantic communication}
\label{tab:channel_aware_semantic_reliability_metrics}
\begin{tabular}{p{0.22\textwidth} p{0.22\textwidth} p{0.28\textwidth} p{0.22\textwidth}}
\toprule
\textbf{KPI} & \textbf{Formula idea} & \textbf{Best suited for} & \textbf{Main limitation} \\
\midrule
Visual semantic distortion & Expected distance between $Z$ and $\hat{Z}$ & Feature, object, relation, or mask representations & Requires defining semantic distance \\
Semantic outage probability & Probability semantic quality falls below threshold & Wireless reliability analysis & Threshold selection is subjective \\
Task outage & Probability task KPI falls below threshold & Goal-oriented image SemCom & Application-specific threshold \\
Rate/SNR-semantic curves & Semantic quality vs. resource/channel condition & Resource-aware comparison & Requires standardized quality and cost definitions \\
\bottomrule
\end{tabular}
\end{table*}

Overall, channel-aware semantic reliability metrics are essential for evaluating image semantic communication as a communication system rather than only as an image reconstruction system. They make it possible to compare whether semantic content remains useful under channel noise, compression, rate limitations, or latency constraints.

\section{Cross-modality Comparison of Text and Image Semantic Communication KPIs}
\label{sec:cross_modality}

The previous sections reviewed semantic communication KPIs separately for text and image data. Although these modalities differ in their data structure, representation, and downstream applications, their evaluation logic follows a common pattern. In both text-based and image-based semantic communication, KPIs can be organized according to what the receiver is expected to preserve or accomplish: low-level reconstruction, perceptual or human-aligned quality, representation-level semantic similarity, downstream task utility, reference-free monitoring, or semantic reliability under channel and resource constraints.

Table~\ref{tab:cross_modality_text_image_kpis} summarizes this cross-modality relationship. The purpose of this comparison is not to claim that text and image KPIs are directly interchangeable, but to show that they often play analogous roles in semantic communication evaluation.

\begin{table*}[t]
\centering
\caption{Cross-modality comparison of text and image semantic communication KPIs}
\label{tab:cross_modality_text_image_kpis}
\begin{tabular}{p{0.18\textwidth} p{0.23\textwidth} p{0.23\textwidth} p{0.30\textwidth}}
\toprule
\textbf{Evaluation perspective} & \textbf{Text-based KPIs} & \textbf{Image-based KPIs} & \textbf{Shared evaluation idea} \\
\midrule
Low-level reconstruction & BLEU, ROUGE, chrF & MSE, PSNR & Measures surface-level or pixel-level similarity between the source and receiver output \\

Structural/perceptual quality & METEOR, BLEURT, COMET & SSIM, MS-SSIM, LPIPS & Evaluates quality in a way that is more aligned with human judgment than exact symbol or pixel matching \\

Representation-level semantic similarity & BERT cosine similarity, BERTScore, sentence embeddings & Feature cosine similarity, CLIP similarity, deep visual embeddings & Compares source and received content in a learned semantic feature space \\

Downstream task utility & Intent accuracy, slot F1, frame accuracy, exact match, Recall@K, MRR & Classification accuracy, mAP, IoU, Dice, Recall@K & Measures whether the receiver can complete the intended task after transmission \\

Reference-free evaluation & COMET-QE, anchor similarity, semantic success/outage, perplexity & No-reference image quality assessment, CLIP-based consistency, semantic anchors, task confidence & Estimates receiver-side quality when the original source or gold reference is unavailable \\

Telecom-style semantic reliability & Semantic distortion, semantic error probability, semantic outage, rate-semantic tradeoff & Semantic distortion, task outage, semantic outage versus SNR/rate/bandwidth & Measures whether meaning is preserved reliably under channel and resource constraints \\
\bottomrule
\end{tabular}
\end{table*}

These six perspectives form a layered evaluation structure common to both modalities. At the lowest level, lexical-overlap metrics (BLEU, ROUGE, chrF) and pixel-fidelity metrics (MSE, PSNR) measure surface similarity but do not necessarily indicate preserved meaning. The next level, METEOR/BLEURT/COMET for text and SSIM/MS-SSIM/LPIPS for images, moves closer to human judgment through word alignment, learned quality prediction, or perceptual structure, though without guaranteeing task-relevant correctness. Representation-level metrics, BERT-based similarity and BERTScore for text, feature cosine similarity and CLIP similarity for images, compare source and received content in a learned semantic feature space, offering a stronger connection to meaning but remaining model-dependent and sensitive to domain alignment.

Task-oriented KPIs are often the most meaningful for a specific receiver goal: intent classification, slot filling, question answering, and retrieval for text; classification, detection, segmentation, and visual question answering for images. Here the best KPI measures whether the receiver can perform its task rather than reconstruction fidelity alone, a reconstructed image can have moderate PSNR yet still support accurate object detection, or high PSNR yet lose a task-critical object. Reference-free evaluation, quality estimation, semantic anchors, perplexity, or outage thresholds for text; no-reference image quality, CLIP-based consistency, or task confidence for images, matters because practical receivers often lack access to the original source, though these methods remain underdeveloped and sensitive to domain shift. Finally, telecom-style reliability metrics, semantic distortion, semantic outage, and rate-semantic tradeoff curves, connect both modalities to communication constraints such as SNR, bandwidth, compression rate, latency, and energy, reflecting that semantic communication is a communication-efficiency problem as much as a representation-learning one.

Despite these shared evaluation levels, text and image modalities differ in what ``meaning'' looks like. In text, meaning is usually linguistic, contextual, and intent-based. It may depend on word choice, sentence structure, discourse context, ambiguity, and pragmatic intent. A semantically correct reconstructed sentence may be a paraphrase rather than a word-by-word copy. Therefore, text KPIs must handle synonymy, paraphrasing, context, and task-specific meaning.

In images, meaning is spatial, perceptual, object-based, and task-dependent. It may depend on object identity, object location, boundaries, texture, scene context, saliency, and relations among objects. A semantically correct reconstructed image does not always need to reproduce every pixel, but it should preserve the visual information that matters for the receiver’s goal. For example, in autonomous driving, pedestrians, traffic lights, vehicles, lanes, and spatial relations are more important than background texture. In medical imaging, organs, lesions, boundaries, and abnormal regions are more important than global visual realism alone.

Therefore, the main lesson from the cross-modality comparison is that semantic communication KPIs should be selected according to the receiver objective rather than the modality alone. Reconstruction metrics are useful when the goal is source recovery, representation metrics are useful when the goal is semantic similarity, task metrics are useful when the goal is decision support, reference-free metrics are useful for deployment monitoring, and telecom-style metrics are useful when semantic quality must be studied under channel and resource constraints. A unified evaluation framework for semantic communication should therefore be modality-aware, task-aware, reference-aware, and communication-aware.

\section{Challenges and Open Gaps in Semantic Communication KPI Design}
\label{sec:gap_analysis}
\subsection{Functional Roles of KPIs in Conventional Networks}

\begin{figure*}[t]
\centering
\IfFileExists{Figures/kpis.jpg}{%
  \includegraphics[width=0.95\textwidth]{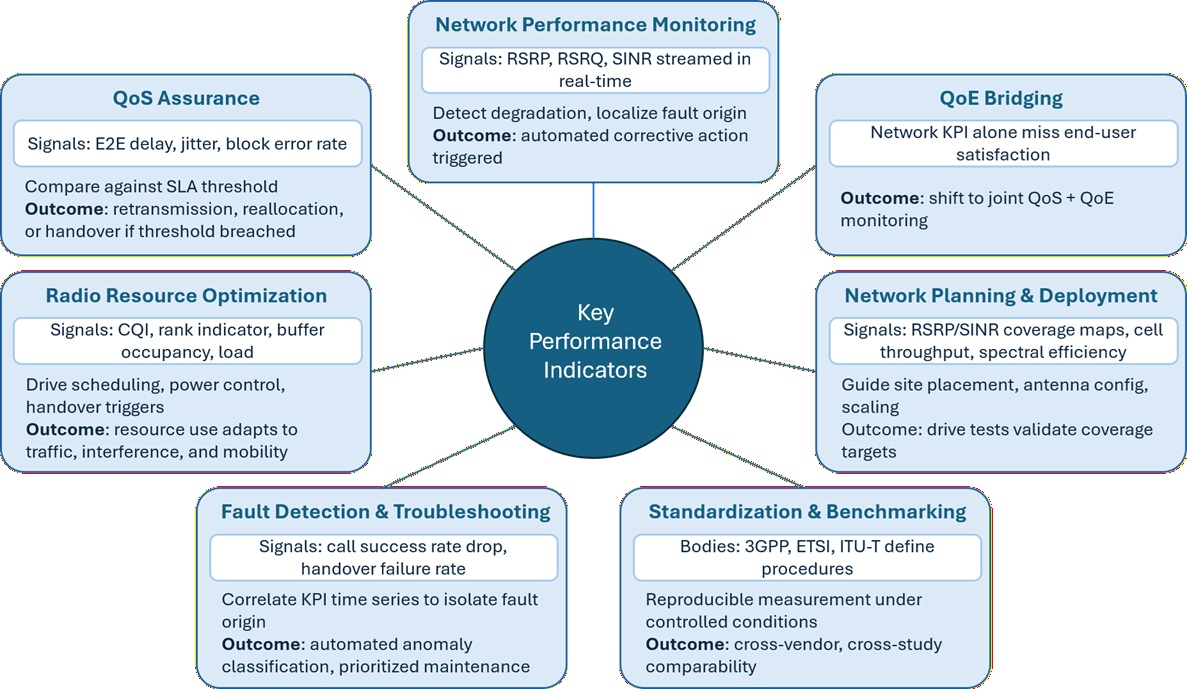}%
}{%
  \fbox{\parbox{0.95\textwidth}{\centering\vspace{1em}
  \textit{Placeholder: ensure the image exists at} \texttt{Figures/kpis.jpg}.\vspace{1em}}}%
che}
% \caption{Functional roles of KPIs in conventional cellular networks.}
\caption{The seven functional roles of KPIs in conventional cellular networks network performance monitoring, QoS assurance, QoE bridging, radio resource optimization, network planning and deployment, fault detection and troubleshooting, and standardization/benchmarking, each of which semantic communication KPIs must eventually serve if semantic systems are to be operationally deployable.}
\label{fig:kpi-functional-roles}
\end{figure*}

KPIs are fundamental metrics used to evaluate, monitor, and optimize the performance of telecommunication systems. In traditional networks, KPIs are not merely descriptive statistics computed after the fact; they are operational instruments embedded in the management, control, and standardization of the entire communication system. Their operational importance can be organized across seven principal functional domains, summarized in Fig.~\ref{fig:kpi-functional-roles} and described individually in this section.

\textbf{Network Performance Monitoring:} KPIs provide the basis for continuous network health assessment. Real-time KPI streams, including Reference Signal Received Power (RSRP), Reference Signal Received Quality (RSRQ), and Signal-to-Interference-plus-Noise Ratio (SINR), enable operators to assess system efficiency and reliability, detect and localize performance degradations and failures, and trigger automated corrective actions. As emphasized in \cite{Delgado2013TETRA}, continuous KPI-based QoS monitoring is a core operational requirement, particularly in large-scale heterogeneous networks where manual fault detection is impractical.

\textbf{Quality of Service (QoS) Assurance:} KPIs provide the technical basis for Service Level Agreements (SLAs), ensuring that applications meet defined latency, throughput, packet loss, and reliability requirements. QoS assurance relies on continuously monitoring metrics such as end-to-end delay, jitter, and block error rate against predefined thresholds. When KPI values fall below acceptable levels, adaptive mechanisms such as retransmission, resource reallocation, or handover can be triggered to restore service quality. Without well-defined KPIs, SLA compliance cannot be reliably verified or critical service degradation distinguished from acceptable performance variations.

\textbf{Quality of Experience (QoE) Bridging:} A critical evolution in telecommunications KPI thinking has been the recognition that network-layer metrics alone do not capture end-user satisfaction. In 5G and beyond, performance monitoring is increasingly transitioning from a purely system-centric to a user-centric architecture, integrating both QoS and QoE dimensions \cite{Liotou2015QoE5G}. 

\textbf{Radio Resource Optimization:} KPIs guide the dynamic allocation of radio resources, including spectrum, power, time slots, and spatial beams. Metrics such as Channel Quality Indicator (CQI), Rank Indicator (RI), buffer occupancy, and load distribution inform scheduling, power control, handover, and congestion management. Continuous KPI monitoring enables network systems to adapt resource allocation to changing traffic, interference, and mobility conditions, improving resource efficiency while maintaining service quality. 

\textbf{Network Planning and Deployment:} KPIs support the design, dimensioning, and deployment of network infrastructure. Coverage metrics such as RSRP and SINR inform base station placement, antenna configuration, and frequency planning, while capacity metrics such as cell throughput, user density, and spectral efficiency guide infrastructure scaling and technology selection. During rollout, KPI-based drive tests and simulations validate deployment targets and identify coverage gaps or interference hotspots requiring corrective action. 

\textbf{Fault Detection and Troubleshooting:} KPIs serve as diagnostic indicators for identifying and localizing network faults. Anomalous patterns, such as drops in call success rate, increased handover failures, or reduced cell throughput, can indicate hardware failures, configuration errors, interference, or protocol issues. Correlating multiple KPI time series supports root cause analysis, while automated anomaly detection can classify faults and prioritize maintenance actions. 

\textbf{Standardization and Benchmarking:} KPIs provide a common basis for objectively comparing network performance across vendors, operators, deployment scenarios, and technology generations. Organizations such as 3GPP, ETSI, and ITU-T define KPI measurement procedures and reporting frameworks that support reproducible comparisons. Standardized KPI benchmarking enables regulators to assess service quality, operators to evaluate infrastructure solutions, and researchers to compare new technologies against established baselines.

These seven functional roles represent the operational requirements that KPIs must collectively satisfy in a mature communication system. Table \ref{tab:canonical-kpis-conventional} summarizes the principal KPIs used in conventional cellular networks, organized by functional layer. Common telecom KPIs include throughput, latency, packet loss rate, spectral efficiency, SNR, SINR, BLER, reliability, and energy efficiency. Together, these metrics characterize performance across the physical, link, and network layers, primarily assessing the successful transmission of bits and symbols over the communication channel.

\begin{table*}[t]
\centering
\caption{Canonical KPIs in Conventional Cellular Communication Systems}
\label{tab:canonical-kpis-conventional}
\renewcommand{\arraystretch}{1.15}
\begin{tabular}{|p{0.18\textwidth}|p{0.52\textwidth}|p{0.25\textwidth}|}
\hline
\textbf{KPI} & \textbf{Definition} & \textbf{Relevant Standard / Reference} \\
\hline
Throughput & Useful data delivered per unit time (bps) & ITU-R M.2150 \\
\hline
Spectral Efficiency & Throughput per unit bandwidth (bps/Hz) & 3GPP TR 36.942 \\
\hline
Peak Data Rate & Maximum achievable single-user rate & ITU-R M.2150: DL 20 Gbps \\
\hline
End-to-End Latency & One-way user-plane delay & ITU-R M.2150: $< 1$ ms (URLLC) \\
\hline
Packet Loss Rate & Fraction of transmitted packets not received & RFC 7679 \\
\hline
SINR & Received signal power vs. interference plus noise & 3GPP TS 38.215 \\
\hline
BLER & Block Error Rate at physical layer & 3GPP TS 38.214 \\
\hline
Reliability & Probability of successful delivery within latency bound & ITU-R M.2150: 99.999\% \\
\hline
Energy Efficiency & Useful bits delivered per unit energy (bits/J) & ITU-R M.2150 \\
\hline
Connection Density & Supported devices per km$^2$ & ITU-R M.2150: $10^6$/km$^2$ \\
\hline
Coverage (RSRP/RSRQ) & Received signal level and quality & 3GPP TS 38.215 \\
\hline
Handover Success Rate & Fraction of successful mobility events & 3GPP TS 32.450 \\
\hline
\end{tabular}
\end{table*}

A defining characteristic of this KPI set is its fundamental orientation toward bit-level fidelity: each metric evaluates the efficiency or reliability of transmitting raw information without considering its semantic content or downstream utility \cite{Gunduz2023BeyondBits, Yang2023SemanticFuture}. As Shannon noted, “the semantic aspects of communication are irrelevant to the engineering problem” \cite{Shannon1948}. This abstraction was intentional and highly effective for the voice-centric and content-delivery applications that shaped successive generations of mobile networks. However, it becomes a fundamental limitation for emerging intelligent and task-oriented communication paradigms. Several trends highlight the insufficiency of purely bit-centric KPIs:

\textbf{Task-oriented machine communication:} Applications in autonomous systems, industrial IoT, and cooperative robotics increasingly require decision accuracy and action effectiveness rather than bit-perfect reconstruction. For example, a robot receiving a compressed scene description gains little from high SINR if semantically critical objects are corrupted while redundant features are preserved \cite{Gunduz2023BeyondBits}.

\textbf{The QoS-to-QoE gap:} For human-centric applications, network KPIs do not always translate directly into perceived quality. This mapping is nonlinear and context-dependent \cite{ITUR_M2410, Delgado2013TETRA}. High average throughput, for example, may coexist with poor QoE due to rebuffering bursts or latency spikes that aggregate metrics fail to capture \cite{Banovic2017Telecom}.

\textbf{Escalating data volumes and spectrum scarcity:} The proliferation of high-resolution video, sensor streams, and multimodal data makes optimizing raw bit transmission increasingly bandwidth-inefficient. Future networks therefore need to move beyond simply increasing capacity toward transmitting fewer but more meaningful representations \cite{Gunduz2023BeyondBits}.

\textbf{The post-Shannon paradigm:} Weaver described three levels of communication: the technical problem of accurate symbol transmission (Level A), the semantic problem of conveying meaning (Level B), and the effectiveness problem of achieving the desired outcome (Level C) \cite{Shannon1948, Weaver1949}. Traditional KPIs primarily address Level A, whereas intelligent, goal-oriented 6G systems require KPI frameworks that also capture Levels B and C \cite{Gunduz2023BeyondBits, Yang2023SemanticFuture}.

This gap between bit-centric KPIs and the requirements of semantic, task-oriented communication motivates the comprehensive re-examination of performance metrics undertaken in this survey. Although prior work has organized semantic communication metrics by modality and task, it has not established a systematic procedure for selecting KPI combinations based on the diverse evaluation conditions faced by practitioners \cite{Zhang2026NativeAI}. Existing metrics span multiple research traditions, including natural language processing, computer vision, image quality assessment, machine learning, and wireless communication, leading to substantial variation in the types of performance reported even for similar evaluation objectives.

This fragmentation also reflects the functional roles traditionally served by KPIs in communication networks: performance monitoring, QoS assurance, QoE assessment, radio resource optimization, network planning, fault detection, and standardization. Semantic communication inherits these operational requirements but introduces additional evaluation dimensions related to meaning and task effectiveness. As discussed in the following subsections, existing semantic KPIs do not yet fully support these seven functions, creating open challenges that must be addressed for semantic communication systems to be reliably evaluated, deployed, and standardized.

\subsection{Open Challenges and Gaps in Semantic Communication KPI Design}

The functional roles described in the previous section presuppose a measurement substrate that is well defined, stable, and reproducible: a KPI is useful only if it can be computed unambiguously, compared across systems, and linked to a clear operational decision. Semantic communication disrupts this foundation. The objective shifts from bit-level fidelity to meaning preservation, which is inherently more context- and application-dependent. This section reviews the principal gaps arising from this shift and identifies which of the seven traditional KPI functions each gap undermines.

\subsubsection{Absence of a Universal Semantic Success Criterion}

In classical communication, the evaluation target is unambiguous: a transmitted bit sequence either matches the received sequence or it does not, while KPIs such as BLER, SINR, and throughput quantify deviations from this objective \cite{Shannon1948, 3GPP_TR25913}. Semantic communication instead asks whether meaning was preserved, the intended task was accomplished, and the receiver's interpretation matched the sender's intent \cite{Gunduz2023BeyondBits, Yang2023SemanticFuture}. Because success varies across applications, no single semantic KPI generalizes across domains. Autonomous driving may require decision correctness, video conferencing perceptual quality, industrial IoT action reliability, multi-agent AI task completion, and medical systems diagnostic accuracy. Semantic KPIs are therefore inherently application-, context-, and receiver-dependent, unlike conventional KPIs.

% This gap directly undermines the standardization and benchmarking function. Without an agreed semantic success criterion, performance claims across different systems and research groups cannot be meaningfully compared, and no common evaluation criterion can exist across vendors, mirroring the absence of a standardized semantic ground truth discussed in Section VII-B. A KPI selection framework that makes the evaluation goal explicit before selecting metrics, as introduced later in this survey, mitigates the practical consequences of this gap but does not resolve the deeper, arguably philosophical, problem of defining a universal reference for meaning.
This gap directly undermines standardization and benchmarking. Without an agreed semantic success criterion, performance claims across systems, vendors, and research groups cannot be meaningfully compared. Addressing this challenge requires either application-specific success criteria agreed in advance or, more fundamentally, a common reference for semantic meaning, which remains an open research problem.

\subsubsection{Difficulty of Quantifying Semantics Numerically}

A related challenge is converting abstract concepts such as meaning, intent, contextual relevance, and semantic distortion into reproducible numerical scores. Unlike bit error rate or throughput, these concepts lack direct and universally accepted measurement procedures. Various semantic KPIs have therefore been proposed, including task accuracy, intent preservation, semantic similarity, knowledge alignment, decision confidence, contextual relevance, semantic entropy, and utility. For example, BERTScore uses contextual embeddings to estimate semantic similarity in text and can better reflect human judgments than lexical metrics such as BLEU or ROUGE for paraphrase and open-ended generation \cite{Zhang2020BERTScore}. In image-based tasks, CLIP-score measures semantic similarity through joint vision-language embeddings, while PSNR, SSIM, LPIPS, and FID capture complementary aspects of reconstruction and perceptual quality \cite{Bourtsoulatze2019DeepJSCC, Radford2021CLIP}.

However, unlike BLER or SINR, these metrics lack standardized measurement methodologies, may provide conflicting assessments for the same output, and can change as their underlying embedding models are updated or fine-tuned. This fragmentation reflects the absence of a standardized framework for measuring and combining semantic KPIs, directly challenging the standardization and benchmarking function that conventional KPIs provide.

\begin{figure*}[t]
\centering
\IfFileExists{Figures/SemDrift.jpg}{%
  \includegraphics[width=0.8\textwidth]{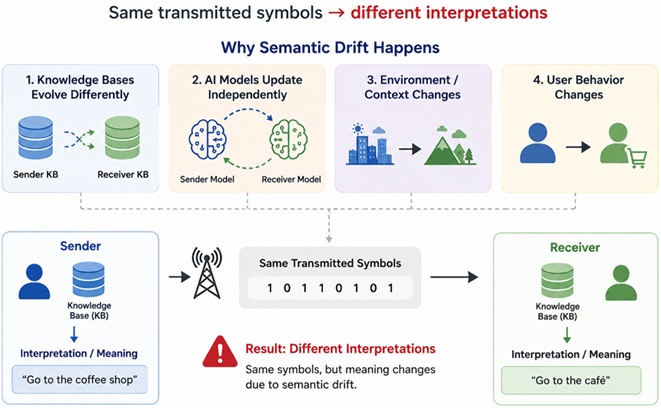}%
}{%
  \fbox{\parbox{0.8\textwidth}{\centering\vspace{1em}
  \textit{Placeholder: ensure the image exists at} \texttt{Figures/semDrift.jpg}.\vspace{1em}}}%
}
% \caption{Semantic drift}
\caption{Illustration of semantic drift: a semantic encoder trained on one deployment context (e.g., urban traffic imagery) produces a compressed representation that is decoded correctly at the receiver only as long as sender and receiver knowledge bases remain aligned. When the operating environment or knowledge base shifts after deployment (e.g., a snowy rural scene), the same representation no longer maps to the same meaning, causing the semantic KPI to degrade even though the channel and encoding pipeline are unchanged.}
\label{fig:sem-drift}
\end{figure*}

\subsubsection{Semantic Drift}

Traditional KPIs are relatively stable over a network's operational lifetime: for example, the relationship between SINR and BLER is governed by physical-layer characteristics and typically changes slowly. Semantic KPIs have no such stationarity. Semantic drift can occur when sender and receiver KBs evolve independently, AI encoder/decoder models are updated asynchronously, or the operating environment and user behavior change. For example, a semantic encoder trained on urban traffic imagery may perform differently when deployed in a snowy rural environment because the contextual assumptions underlying the shared representation no longer hold. This creates KPI instability with no direct analogue in conventional systems, as illustrated in Fig.~\ref{fig:sem-drift}.

% See Fig. \ref{fig:sem-drift}

This gap directly undermines network performance monitoring. Conventional monitoring assumes approximately stable KPI distributions, allowing deviations to indicate faults. Semantic KPIs that evolve with knowledge bases, models, or deployment context make it difficult to distinguish genuine semantic failures from expected effects of non-stationary drift.

\begin{figure*}[t]
\centering
\IfFileExists{Figures/kpi_complexity.jpg}{%
  \includegraphics[width=1\textwidth]{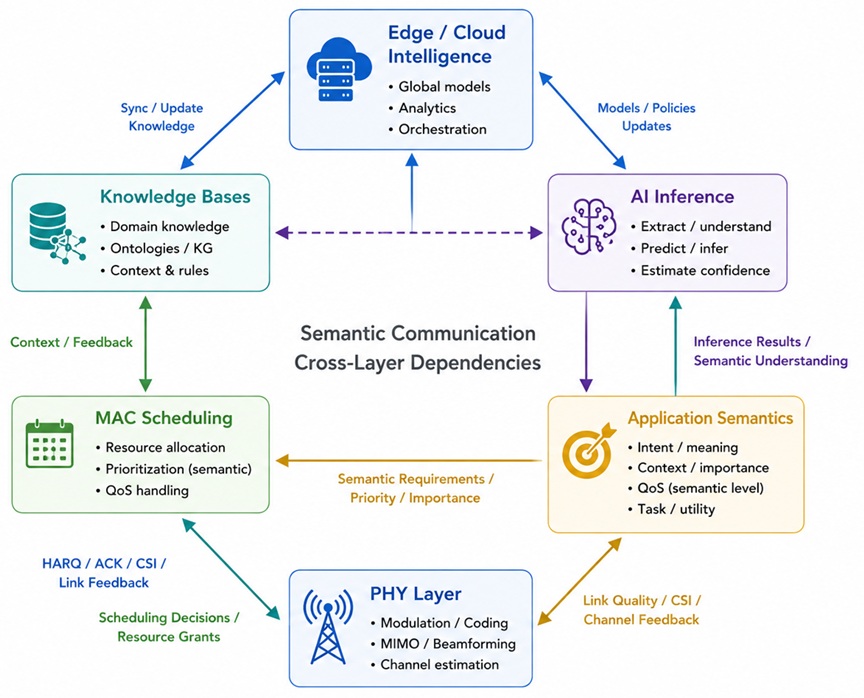}%
}{%
  \fbox{\parbox{1\textwidth}{\centering\vspace{1em}
  \textit{Placeholder: ensure the image exists at} \texttt{Figures/kpi_complexity.jpg}.\vspace{1em}}}%
}
% \caption{Cross-layer KPI complexity.}
\caption{Cross-layer entanglement of KPIs in semantic communication. Unlike conventional systems, where physical-, MAC-, and application-layer KPIs are related through well-characterized mappings, semantic communication tightly couples the physical layer, MAC scheduling, AI inference, knowledge bases, and edge/cloud intelligence, so that a lower physical-layer throughput or higher bit error rate can still yield higher end-to-end semantic utility}
\label{fig:cross-layer-kpi-complexity}
\end{figure*}

\subsubsection{Cross-Layer KPI Entanglement}

% \ See Fig. \ref{fig:cross-layer-kpi-complexity}

In conventional systems, physical-, MAC-, and application-layer KPIs are connected through relatively well-characterized relationships and can often be analyzed independently. Semantic communication introduces tighter dependencies among the physical layer, MAC scheduling, application semantics, AI inference, knowledge bases, and edge/cloud intelligence, as illustrated in Fig.~\ref{fig:cross-layer-kpi-complexity}. Consequently, lower throughput or higher bit error rates can sometimes yield higher semantic utility if semantically important information is prioritized for transmission. Thus, minimizing bit-level error is no longer necessarily a valid proxy for maximizing end-to-end utility.

This gap simultaneously undermines radio resource optimization and network performance monitoring. Resource allocation strategies that maximize throughput or minimize BLER may fail to maximize semantic utility, while physical-layer monitoring cannot detect semantic failures when conventional KPIs remain within normal ranges.

\subsubsection{Real-Time Computability of Semantic KPIs}

Semantic KPI evaluation may require AI inference, embedding computation, multimodal processing, large language model reasoning, or knowledge graph analysis, making it substantially more computationally expensive than measuring SINR or packet loss. These operations introduce additional latency, energy consumption, and processing requirements, raising the challenge of evaluating semantic KPIs in real time and at scale, particularly in 5G/6G RANs, resource-constrained edge devices, IoT deployments, and URLLC systems. The evaluation overhead itself may exceed the latency budget of the application being monitored.

This gap directly challenges network performance monitoring: a KPI is useful for real-time control only if its computation is faster than the process it monitors. For example, a semantic KPI requiring multi-second inference is unsuitable for a URLLC application with a millisecond-scale latency budget.

\subsubsection{Multi-User Semantic Fairness}

Conventional schedulers allocate radio resources according to established fairness criteria such as proportional fairness and max-min fairness. Semantic communication may instead require allocating resources according to meaning importance, task utility, or inference priority, for which no equivalent fairness criterion currently exists. It remains unclear whether semantic fairness should mean equal semantic accuracy, equal task success probability, or priority based on semantic criticality, which may itself be application-dependent and time-varying.

This gap undermines radio resource optimization because scheduler design requires a well-defined fairness objective. Establishing appropriate semantic fairness criteria therefore represents a new dimension of KPI and scheduler research beyond conventional radio resource managemen.

\begin{figure*}[t]
\centering
\IfFileExists{Figures/explainability.jpg}{%
  \includegraphics[width=0.8\textwidth]{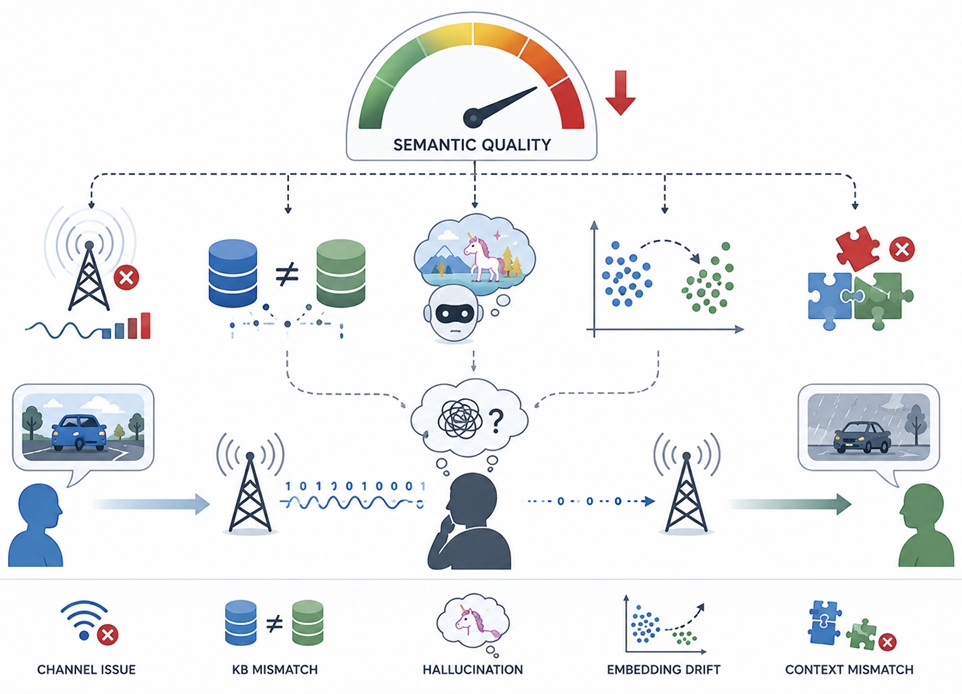}%
}
{%
  \fbox{\parbox{0.8\textwidth}{\centering\vspace{1em}
  \textit{Placeholder: ensure the image exists at} \texttt{Figures/explainability.jpg}.\vspace{1em}}}%
}
% {%
%   \fbox{\parbox{0.8\textwidth}{\centering\vspace{1em}
%   \textit{Placeholder: ensure the image exists at} \texttt{Figures/explainability.jpg}.\vspace{1em}}}%
% }
% \caption{Explainability and trust.}
\caption{Explainability gap in semantic KPI failures: when semantic quality degrades, the underlying cause (channel impairment, knowledge-base mismatch, model hallucination, embedding drift, or train/deployment context mismatch) is obscured by the black-box nature of the neural encoders and decoders, making root-cause diagnosis substantially harder than for a conventional link-layer fault.}
\label{fig:explainability-trust}
\end{figure*}

\subsubsection{Explainability and Trust in Semantic KPI Failures}

Network operators require KPIs with interpretable failure modes to support root-cause diagnosis and maintain trust in automated decisions, as illustrated in Fig.~\ref{fig:explainability-trust}. Semantic communication systems often rely on deep neural networks, latent embeddings, and other black-box models whose failure modes are difficult to observe. When semantic quality degrades, it may be unclear whether the cause is channel impairment, sender--receiver KB mismatch, model hallucination, embedding drift, or a mismatch between training and deployment contexts. Diagnosing such failures is therefore considerably more difficult than troubleshooting conventional link-layer faults, where potential causes are fewer and better instrumented.

This gap undermines the fault detection and troubleshooting function, which depends on linking anomalous KPI patterns to identifiable root causes, an assumption that becomes difficult to maintain when model-internal behavior is opaque.

\subsubsection{Lack of Standardized Semantic Ground Truth}

A related but distinct challenge is the absence of a standardized reference for measuring semantic fidelity. In conventional communication, transmitted and received bit sequences can be directly compared. Semantic communication instead compares meaning, which often lacks a unique canonical representation. In text, different sentences may express the same meaning despite having little lexical overlap. Similarly, an image may have low pixel-level similarity while preserving its key objects and scene meaning, or achieve high PSNR while losing a semantically critical detail.

This gap undermines standardization and benchmarking. Without a standardized semantic reference, performance claims across systems, vendors, and research groups cannot be reliably compared. A systematic KPI-selection framework that defines the evaluation objective before selecting metrics can mitigate this challenge procedurally, but cannot resolve the deeper problem of establishing a universal semantic ground truth.

\subsubsection{KPI Dependence on the Communication Goal}

The suitability of a semantic KPI depends strongly on the communication objective: a metric is meaningful only when aligned with what the receiver is expected to reconstruct, infer, or accomplish. Consequently, no single KPI can adequately evaluate all semantic communication systems. For faithful image reconstruction, metrics such as MSE and PSNR may be appropriate, whereas task-oriented systems require metrics such as classification accuracy, F1 score, mAP, Dice coefficient, or Recall@K \cite{Bourtsoulatze2019DeepJSCC}. A key gap is the absence of standardized guidelines for systematically mapping communication goals to appropriate KPI families.

This gap undermines QoS assurance and network planning. Without goal-aligned KPIs, acceptable semantic service quality cannot be clearly defined, nor can infrastructure be systematically planned around application-specific requirements.

\subsubsection{Weak Connection Between Machine-Learning Metrics and Communication Constraints}

Many semantic KPIs are adopted from machine learning, NLP, and computer vision, where they evaluate output quality or task performance independently of communication constraints. They often omit factors such as SNR, bandwidth, transmission rate, latency, energy consumption, transmitted symbols, and computational complexity. A system may therefore achieve high BERTScore, CLIP similarity, or task accuracy while requiring impractical communication or computational resources.

Addressing this gap requires evaluation protocols that explicitly relate semantic quality to resource cost, for example through semantic-quality-versus-bits tradeoff curves or efficiency metrics such as semantic quality per unit of channel resource. This gap directly undermines radio resource optimization and network performance monitoring, since adaptive scheduling, power control, and congestion management require performance measures that vary predictably with allocated resources.

\subsubsection{Underdevelopment of Reference-Free Evaluation}

Most semantic communication metrics assume access to the original source or a gold-standard reference during evaluation. In deployment, however, the receiver often has no such reference. Reference-free approaches include learned quality estimators, semantic-anchor similarity, and fluency measures such as perplexity, but these remain limited. For example, a reconstructed image may appear realistic and receive a high no-reference quality score while containing incorrect objects, missing task-critical regions, or distorted spatial relationships.

Developing reference-free KPIs that are both computationally practical and semantically reliable therefore remains an open challenge. This gap undermines fault detection and network performance monitoring, since deployed systems must detect semantic failures at the receiver without access to the original source, analogous to conventional network monitoring based on receiver-side KPI anomalies.

These gaps, together with those discussed in the remainder of this section, are consolidated in Table~\ref{tab:open-gaps-vs-functions}, which maps each gap to the traditional KPI function(s) it undermines.
\begin{table*}[t]
\centering
\caption{Open Gaps in Semantic Communication KPI Design and the Traditional KPI Functions They Undermine}
\label{tab:open-gaps-vs-functions}
\renewcommand{\arraystretch}{1.15}
\begin{tabular}{|p{0.50\textwidth}|p{0.50\textwidth}|}
\hline
\textbf{Gap} & \textbf{Traditional Function(s) Undermined} \\
\hline
A. No universal semantic success criterion & Standardization and benchmarking \\
\hline
B. Difficulty quantifying semantics numerically & Standardization and benchmarking \\
\hline
C. Semantic drift & Network performance monitoring \\
\hline
D. Cross-layer KPI entanglement & Radio resource optimization; network performance monitoring \\
\hline
E. Real-time computability & Network performance monitoring \\
\hline
F. Multi-user semantic fairness & Radio resource optimization \\
\hline
G. Explainability and trust & Fault detection and troubleshooting \\
\hline
H. Lack of standardized semantic ground truth & Standardization and benchmarking \\
\hline
I. KPI dependence on communication goal & QoS assurance; network planning \\
\hline
J. Weak link to communication constraints & Radio resource optimization; network performance monitoring \\
\hline
K. Underdeveloped reference-free evaluation & Fault detection; network performance monitoring \\
\hline
L. Task- and intent-aware KPI selection & QoS assurance; fault detection \\
\hline
M. Relation-level and multimodal KPIs & Standardization and benchmarking \\
\hline
\end{tabular}
\end{table*}

\subsubsection{Task-Aware and Intent-Aware KPI Selection}

Semantic communication aims to preserve information relevant to the receiver's goal rather than all source information. Consequently, different intents require different semantic content to be prioritized. For example, a visual weather-understanding task may prioritize clouds, precipitation, and road conditions, while medical diagnosis may prioritize organs, lesions, and anatomical abnormalities over perceptually faithful reconstruction of surrounding tissue. However, many existing semantic communication KPIs still measure global similarity, effectively treating all content as equally important regardless of the receiver's objective.

This gap undermines QoS assurance and fault detection. Without distinguishing task-critical from non-critical information, a system cannot reliably define task-relevant service-quality thresholds or detect failures in which critical content is lost while unimportant content is preserved.

\subsubsection{Need for Relation-Level and Multimodal KPIs}

Many image-based semantic communication metrics evaluate pixels, global features, object labels, or downstream task outputs, But visual meaning also depends on relationships among objects. Relation-level representations, such as subject-predicate-object triplets and scene graphs, capture spatial, functional, and human-object relationships that object-centric metrics may overlook \cite{Krishna2017Visual, Lu2016Visual}. Multimodal semantic communication, including text-image, video-language, and sensor-fusion settings, adds further complexity because semantic correctness depends on cross-modal alignment rather than fidelity within a single modality. Existing single-modality KPIs are therefore insufficient, and no standardized multimodal semantic KPI has yet been established.

This gap undermines standardization and benchmarking across modalities and limits the development of common evaluation criteria for multimodal semantic communication systems.

% \textbf{Summary: } Table II consolidates the fourteen gaps discussed above according to the traditional KPI function each one undermines. A recurring pattern emerges: the standardization and benchmarking function is the most frequently and most severely disrupted, appearing in connection with the absence of a universal success criterion, the absence of standardized ground truth, and the absence of multimodal KPIs. The network performance monitoring function is disrupted in a qualitatively different way, not by the absence of a metric, but by the breakdown of the stationarity and real-time computability assumptions on which conventional monitoring depends. These structural correspondences motivate the systematic KPI selection framework developed later in this survey, which is explicitly organized around restoring, where possible, the link between semantic evaluation and the seven functional requirements established in Section II.

\textbf{Summary: } Table~\ref{tab:open-gaps-vs-functions} consolidates the thirteen gaps according to the traditional KPI functions they undermine. A recurring pattern is that standardization and benchmarking are particularly affected by the absence of universal success criteria, standardized ground truth, and multimodal KPIs. Network performance monitoring is disrupted differently, through the breakdown of conventional assumptions regarding KPI stationarity and real-time computability. These gaps motivate a systematic KPI-selection framework that reconnects semantic evaluation with the seven functional requirements established above.

\section{Future Directions}
\label{sec:future_directions}

\subsection{Standardized Semantic KPI Benchmarks}

A major open problem in semantic communication is the lack of standardized evaluation benchmarks. Studies currently use different datasets, channel models, tasks, metrics, and reporting protocols, making fair comparison difficult even for systems addressing the same modality and communication goal.

For text-based semantic communication, benchmark protocols should specify the task setting—such as sentence reconstruction, intent classification, question answering, retrieval, or semantic state recovery—and define the reference-based and reference-free KPIs to be reported alongside channel and resource conditions. For image-based systems, benchmarks should distinguish reconstruction-, perception-, and task-oriented evaluation and report performance across multiple SNR levels, bandwidth ratios, or compression rates.

Common channel conditions and resource budgets are essential to distinguish genuine semantic gains from improvements achieved simply through greater resource usage. Standardized benchmarks should therefore jointly report semantic quality, task performance, and communication cost under reproducible conditions. The seven KPI functions identified in Section \ref{sec:gap_analysis} (network performance monitoring, QoS assurance, QoE bridging, radio resource optimization, network planning, fault detection, and standardization) provide a natural checklist for ensuring that such benchmarks support both technical evaluation and practical network deployment.

\subsection{Learned and Adaptive KPI Selection}

% The KPI selection framework proposed in Section~6 follows a structured but static procedure: given a fixed set of input axes, it maps a scenario to a recommended KPI combination. This is sufficient for system design and evaluation planning, but it does not adapt to changing channel conditions, task distributions, or receiver states during operation.
Most KPI-selection approaches follow a structured but static procedure: given fixed axes such as modality, communication goal, reference availability, and channel condition, they map a scenario to a recommended KPI set. While suitable for system design and evaluation planning, this approach cannot adapt to changing channel conditions, task distributions, or receiver states during operation.

An open research direction is to make KPI selection learned or adaptive. The selected metric, or the weighting of a composite KPI set, could be adjusted according to observed system behavior rather than fixed at design time. For example, under high SNR and low compression, reconstruction metrics such as BERTScore or SSIM may be most informative; under severe channel degradation, outage and distortion metrics may become more relevant; and under task distribution shifts, task-specific KPIs may require greater weighting. A static protocol cannot adapt equally well to these conditions.

This problem connects to research in automated machine learning and meta-evaluation. For semantic communication, a key question is whether a receiver-side model can learn to select or weight KPIs based on channel state, task confidence, or semantic-anchor similarity without manually redefining the evaluation protocol as conditions change. Such an approach would transform KPI selection from a design-time decision into a runtime capability, supporting autonomous and self-optimizing semantic communication systems.

\subsection{A New Composite KPI for Context-Aware and Multimodal Semantic Communication}

% The gap analysis in Section~8 and the framework analysis in Section~6.5 identify two evaluation gaps that remain unresolved by any existing metric or combination of existing metrics: multimodal context consistency, and structured reference-free semantic quality at the receiver. These gaps are not only theoretical. They appear concretely in scenarios M1, T3, and I5 of the scenario mapping table, where the framework can only recommend global alignment scores or threshold-based outage because no existing metric decomposes semantic quality into interpretable, context-aware components.
The gap analysis in Section \ref{sec:gap_analysis} identifies two evaluation challenges that remain unresolved by existing metrics: multimodal context consistency and structured, reference-free semantic quality at the receiver. These gaps become particularly evident in multimodal and reference-free scenarios, where current metrics largely provide global alignment scores or threshold-based outage measures, without decomposing semantic quality into interpretable, context-aware components.

Addressing these challenges requires a KPI designed specifically for text-image and context-aware semantic communication. Rather than relying on global embeddings or pixel-level comparisons, such a KPI would evaluate semantic preservation at the entity, relation, and context levels. It could be decomposed into interpretable components covering semantic similarity, context consistency, entity and relation preservation, task relevance, multimodal alignment, and communication cost. The KPI should support both reference-based and reference-free evaluation to enable practical deployment, while jointly capturing semantic quality and transmission efficiency.

\section{Conclusion}
\label{sec:conclusion}

Semantic communication shifts the objective from exact symbol recovery toward meaning preservation, task completion, and efficient semantic information exchange. Consequently, evaluation cannot rely on a single universal KPI. The appropriate metric depends on the receiver's objective, reference availability, channel and resource constraints, and transmitted modality.

This survey reviewed semantic communication KPIs for text and image data from an evaluation-centered perspective. For text, metrics range from reconstruction measures such as BLEU, ROUGE, METEOR, chrF, BERTScore, BLEURT, and COMET to goal-oriented measures including intent accuracy, slot F1, exact match, retrieval performance, and task success rate. Reference-free and telecom-oriented metrics, including quality estimation, semantic anchors, semantic distortion, semantic error probability, semantic outage, and rate-semantic tradeoffs, provide additional tools when exact references are unavailable. For images, KPIs range from pixel-level metrics such as MSE and PSNR, through structural and perceptual measures such as SSIM, MS-SSIM, and LPIPS, to representation-level metrics such as feature and CLIP similarity, task-oriented measures such as accuracy, mAP, IoU, Dice, and Recall@K, and channel-aware measures such as semantic and task outage.

The cross-modality analysis in Section \ref{sec:cross_modality} reveals a common layered evaluation logic spanning low-level reconstruction, perceptual quality, representation-level semantic similarity, downstream task utility, reference-free monitoring, and communication-aware reliability. However, semantic correctness remains modality-dependent: text is primarily linguistic, contextual, and intent-based, whereas images are spatial, perceptual, object-based, and task-dependent. A unified framework must therefore remain modality-aware in how these evaluation levels are applied.

% The central contribution of this survey is the systematic KPI selection framework proposed in Section~6. The framework takes five input axes: source modality, communication goal, receiver output type, reference availability, and channel constraint. It follows a four-step procedure to produce a scenario-specific evaluation protocol. The scenario mapping table covers ten canonical semantic communication scenarios spanning text reconstruction, task-oriented text, reference-free deployment, semantic-state recovery, image reconstruction at different quality priorities, generative image systems, and multimodal text-image retrieval. The three-level composite evaluation protocol requires reporting semantic quality, communication efficiency, and reliability under degradation together, so that no evaluation can claim strong semantic performance while ignoring channel constraints or resource cost.

The gap analysis in Section \ref{sec:gap_analysis} links semantic KPI challenges to the seven functional roles of conventional KPIs: network performance monitoring, QoS assurance, QoE bridging, radio resource optimization, network planning, fault detection, and standardization. It shows that existing semantic KPIs do not yet fully support these functions. Two gaps remain particularly unresolved: the lack of a structured, reference-free semantic quality measure with interpretable components, and the lack of a multimodal KPI that jointly evaluates context consistency across text and image modalities rather than relying solely on global alignment.

% These gaps motivate the second research direction of this project: a composite KPI for context-aware and multimodal semantic communication that evaluates preservation at the entity, relation, and context levels; supports both reference-based and reference-free evaluation; and incorporates communication cost to jointly assess semantic quality and transmission efficiency.

Future semantic communication evaluation should therefore move toward standardized, modality-aware, task-aware, reference-aware, and communication-aware KPI frameworks. Existing metrics remain valuable, but no individual metric is sufficient to determine whether a semantic communication system achieves its objective reliably, efficiently, and interpretably.

% --- References ---
% \clearpage
\bibliographystyle{IEEEtran}
\bibliography{references}

% --- Appendices (5) ---
% \clearpage
\appendices

\section{Illustrative Examples of Text Semantic Reconstruction KPIs}
\label{apx_text_reconstruction}
\subsection{BLEU Example}
\label{apx_bleu_example}
To illustrate this limitation, consider the source sentence $S = ``\text{The weather is cold today}''$ and the reconstructed sentence $\hat{S} = ``\text{It is freezing today}.''$ The two sentences are semantically close, since ``cold'' and ``freezing'' express similar weather conditions. However, their surface-level overlap is limited. At the unigram level, only ``is'' and ``today'' are exact matches, giving $p_1 = 2/4 = 0.5$. At the bigram level, the reference bigrams \{``the weather'', ``weather is'', ``is cold'', ``cold today''\} and reconstructed bigrams \{``it is'', ``is freezing'', ``freezing today''\} have no exact overlap, giving $p_2 = 0$. Therefore, sentence-level BLEU becomes very low, and without smoothing it may become zero when higher-order n-gram precisions are included. This example illustrates that BLEU mainly captures lexical similarity and may penalize meaning-preserving paraphrases, which is a key limitation in semantic communication.

\subsection{ROUGE-N and ROUGE-L Example}
\label{apx_rouge_example}

As an illustrative example, consider again $S = ``\text{The weather is cold today}''$ and $\hat{S} = ``\text{It is freezing today}.''$ At the unigram level, the two sentences share the words ``is'' and ``today.'' Since the reference sentence contains five unigrams, the ROUGE-1 recall score is $\mathrm{ROUGE\text{-}1} = 2/5 = 0.4$. At the bigram level, the reference bigrams \{``the weather'', ``weather is'', ``is cold'', ``cold today''\} and reconstructed bigrams \{``it is'', ``is freezing'', ``freezing today''\} have no exact overlap, resulting in $\mathrm{ROUGE\text{-}2} = 0$. For ROUGE-L, the longest common subsequence is \{``is'', ``today''\}, with length 2. Therefore, $P_{\mathrm{LCS}} = 2/4 = 0.5$ and $R_{\mathrm{LCS}} = 2/5 = 0.4$. If $\beta = 1$, ROUGE-L reduces to the harmonic mean of $P_{\mathrm{LCS}}$ and $R_{\mathrm{LCS}}$, giving approximately 0.44. Although ROUGE-L assigns a partial score because the common words appear in the same order, ROUGE still does not explicitly capture that ``cold'' and ``freezing'' are semantically related. This example shows that ROUGE can measure how much reference content is lexically recovered, but it may underestimate meaning preservation when semantic equivalence is expressed using different words.

\subsection{METEOR Example}
\label{apx_meteor_example}

As an illustrative example, consider again $S = ``\text{The weather is cold today}''$ and $\hat{S} = ``\text{It is freezing today}.''$ With exact unigram matching only, the matched words are ``is'' and ``today,'' so $m = 2$. Therefore, the unigram precision and recall are $P = 2/4 = 0.5$ and $R = 2/5 = 0.4$, respectively. However, unlike BLEU and ROUGE, METEOR can also incorporate stemming and synonym-based matching. If ``cold'' and ``freezing'' are treated as semantically related terms, then the number of matched unigrams increases to $m = 3$, giving $P = 3/4 = 0.75$ and $R = 3/5 = 0.6$. As a result, METEOR can assign a higher score to this reconstruction than purely lexical-overlap metrics. This example illustrates why METEOR is more suitable for meaning-preserving paraphrases than BLEU or ROUGE, although it still depends on the availability of a reference sentence and lexical matching resources.

\subsection{chrF / chrF++ Example}
\label{apx_chrf_example}

As an illustrative example, consider again $S = ``\text{The weather is cold today}''$ and $\hat{S} = ``\text{It is freezing today}.''$ At the word level, the two sentences share only ``is'' and ``today,'' so word-overlap metrics assign relatively low scores. chrF, however, compares character n-grams rather than complete words. For example, after lowercasing and removing spaces, the two sentences share several character unigrams, such as t, h, e, i, s, o, d, a, and y, which leads to non-zero character-level precision and recall. However, this overlap reflects surface-form similarity rather than semantic equivalence. Thus, chrF can be useful for spelling, morphology, and tokenization differences, but it does not explicitly recognize that ``cold'' and ``freezing'' are semantically related.

chrF++ extends chrF by including both character n-grams and word n-grams. In semantic communication, chrF is useful when reconstruction errors involve spelling, morphology, or tokenization differences; however, because it operates on surface character patterns, it does not directly measure semantic equivalence.

\subsection{BERT-Based Sentence Similarity Example}
\label{apx_bert_similarity_example}

To illustrate the computation, consider the source sentence $S = ``\text{The weather is cold today}''$ and the reconstructed sentence $\hat{S}_1 = ``\text{It is freezing today}.''$ Both sentences are first encoded by a sentence encoder $B(\cdot)$, producing two sentence-level embedding vectors $B(S)$ and $B(\hat{S}_1)$. The similarity score is then obtained by computing the cosine similarity between these vectors. In practice, these embeddings are high-dimensional vectors generated by a pretrained model such as BERT or Sentence-BERT. For illustration, suppose the encoder maps the two sentences into simplified three-dimensional vectors:
\begin{equation}
B(S) = [0.80, 0.50, 0.30],
\end{equation}
\begin{equation}
B(\hat{S}_1) = [0.75, 0.55, 0.35].
\end{equation}

The dot product is:
\begin{equation}
B(S) \cdot B(\hat{S}_1)
=
(0.80)(0.75) + (0.50)(0.55) + (0.30)(0.35)
=
0.98.
\end{equation}

The vector norms are:
\begin{equation}
\|B(S)\|
=
\sqrt{0.80^2 + 0.50^2 + 0.30^2}
\approx
0.99,
\end{equation}
\begin{equation}
\|B(\hat{S}_1)\|
=
\sqrt{0.75^2 + 0.55^2 + 0.35^2}
\approx
0.994.
\end{equation}

Therefore, the cosine similarity is:
\begin{equation}
\mathrm{Sim}(S,\hat{S}_1)
=
\frac{0.98}{0.99 \times 0.994}
\approx
0.995.
\end{equation}

This high value indicates that the two sentence embeddings point in a similar direction in the semantic vector space. Therefore, even though the sentences have limited exact word overlap, the embedding-based metric can assign a high similarity score because the reconstructed sentence preserves the meaning of the source.

For comparison, an unrelated reconstruction such as $\hat{S}_2 = ``\text{A dog is sleeping on the sofa}''$ would be expected to produce an embedding vector with a lower cosine similarity to $B(S)$. For example, if the encoder produced a simplified vector $B(\hat{S}_2) = [0.10, 0.20, 0.90]$, then:
\begin{equation}
B(S) \cdot B(\hat{S}_2)
=
(0.80)(0.10) + (0.50)(0.20) + (0.30)(0.90)
=
0.45.
\end{equation}

\begin{equation}
\|B(\hat{S}_2)\|
=
\sqrt{0.10^2 + 0.20^2 + 0.90^2}
\approx
0.927.
\end{equation}

\begin{equation}
\mathrm{Sim}(S,\hat{S}_2)
=
\frac{0.45}{0.99 \times 0.927}
\approx
0.49.
\end{equation}

Thus, the meaning-preserving reconstruction obtains a much higher similarity score than the unrelated reconstruction. This example illustrates why BERT-based sentence similarity is useful for semantic communication: it evaluates semantic closeness in a learned representation space rather than relying only on surface-level word overlap.

\subsection{BERTScore Example}
\label{apx_bertscore}
Consider the same source sentence and two reconstructed sentences:

\textbf{Source sentence $S$:}
``The weather is cold today.''

\textbf{Meaning-preserving reconstruction $\hat{S}_1$:}
``It is freezing today.''

\textbf{Unrelated reconstruction $\hat{S}_2$:}
``A dog is sleeping on the sofa.''

BERTScore compares contextual token embeddings instead of exact word overlap. Let $x_i$ denote the contextual embedding of the $i$-th token in $S$, and let $\hat{x}_j$ denote the contextual embedding of the $j$-th token in $\hat{S}$. For each token, BERTScore finds the most similar token in the other sentence using cosine similarity.

For the meaning-preserving reconstruction $\hat{S}_1$, the token ``freezing'' can be matched with ``cold'' because their contextual embeddings are expected to be close in the semantic space. The words ``is'' and ``today'' can also be matched directly. Therefore, BERTScore can assign partial semantic credit even though BLEU or ROUGE may give a low score due to limited exact n-gram overlap.

For the unrelated reconstruction $\hat{S}_2$, most tokens such as ``dog,'' ``sleeping,'' and ``sofa'' are not semantically close to the weather-related tokens in $S$. Therefore, the maximum token-level similarities are expected to be lower, resulting in a lower BERTScore.

This example illustrates why BERTScore is more suitable than surface-overlap metrics for semantic communication: it can reward meaning-preserving substitutions such as ``cold'' and ``freezing,'' while still penalizing unrelated reconstructions.

\subsection{BLEURT Example}
\label{apx_bleurt_example}

Consider the same source sentence and two reconstructed sentences:

\textbf{Source/reference sentence $S$:}
``The weather is cold today.''

\textbf{Meaning-preserving reconstruction $\hat{S}_1$:}
``It is freezing today.''

\textbf{Unrelated reconstruction $\hat{S}_2$:}
``A dog is sleeping on the sofa.''

BLEURT is a learned evaluation metric that takes a reference sentence and a candidate sentence as input and outputs a scalar quality score. Unlike BLEU, ROUGE, or METEOR, BLEURT does not compute the score using a fixed overlap formula. Instead, it uses a BERT-based neural model trained to predict human quality judgments.

For the pair $(S, \hat{S}_1)$, the input to BLEURT can be represented as:

\begin{quote}
[CLS] The weather is cold today. [SEP] It is freezing today. [SEP]
\end{quote}

Because $\hat{S}_1$ preserves the weather-related meaning of $S$, a well-trained BLEURT model is expected to assign it a relatively high quality score.

For the pair $(S, \hat{S}_2)$, the input becomes:

\begin{quote}
[CLS] The weather is cold today. [SEP] A dog is sleeping on the sofa. [SEP]
\end{quote}

Since $\hat{S}_2$ changes the topic completely and does not preserve the meaning of the source sentence, BLEURT is expected to assign it a much lower score.

This example shows how BLEURT can act as a learned semantic reconstruction metric in semantic communication. However, the exact numerical score depends on the BLEURT checkpoint, training data, and domain, so the model version should always be reported.

\subsection{COMET Example}
\label{apx_comet_example}

Consider the same source sentence and two reconstructed sentences:

\textbf{Source sentence $S$:}
``The weather is cold today.''

\textbf{Meaning-preserving reconstruction $\hat{S}_1$:}
``It is freezing today.''

\textbf{Unrelated reconstruction $\hat{S}_2$:}
``A dog is sleeping on the sofa.''

COMET is a learned evaluator originally developed for machine translation evaluation. A reference-based COMET model usually receives three inputs: a source sentence, a candidate output, and a reference sentence. In a monolingual semantic communication setting, the original sentence can be used as the source/reference, and the reconstructed sentence can be treated as the candidate output.

For the meaning-preserving reconstruction, the COMET input can be represented as:

\begin{quote}
Source: ``The weather is cold today.'' \\
Candidate: ``It is freezing today.'' \\
Reference: ``The weather is cold today.''
\end{quote}

The model encodes the source, candidate, and reference using a pretrained Transformer encoder. It then builds comparison features between their embeddings and passes them through a regression head to produce a scalar quality score. Since $\hat{S}_1$ preserves the meaning of the source sentence, COMET is expected to assign a relatively high score.

For the unrelated reconstruction, the input becomes:

\begin{quote}
Source: ``The weather is cold today.'' \\
Candidate: ``A dog is sleeping on the sofa.'' \\
Reference: ``The weather is cold today.''
\end{quote}

Because $\hat{S}_2$ does not preserve the meaning of the original sentence, COMET is expected to assign a lower score.

This example illustrates how COMET can be used as a learned, human-aligned semantic reconstruction metric. However, the exact score depends on the COMET variant, checkpoint, and input format. Therefore, semantic communication papers should clearly specify whether they use reference-based COMET or a reference-free COMET-QE variant.

\section{Illustrative Examples of Goal-Oriented Text Semantic Communication KPIs}
\label{apx_goal_oriented}

% This appendix provides illustrative examples for the goal-oriented text semantic communication KPIs discussed in Section~3.2.
This appendix provides illustrative examples for the goal-oriented text semantic communication KPIs discussed in Section~\ref{sec:goal_oriented_text_kpis}. Unlike semantic reconstruction metrics, these KPIs do not mainly evaluate whether the reconstructed sentence matches the original sentence. Instead, they evaluate whether the receiver can correctly infer the intended task output.

Consider the following source sentence:

\begin{quote}
$S = ``\text{Book a flight to Montreal tomorrow.}''$
\end{quote}

The ground-truth task output is:

\begin{quote}
Intent: BookFlight \\
Slots: destination = Montreal, date = tomorrow
\end{quote}

Assume the receiver predicts:

\begin{quote}
Predicted intent: BookFlight \\
Predicted slots: destination = Montreal
\end{quote}

In this example, the receiver correctly identifies the intent and one slot, but misses the date slot. This allows us to illustrate how different goal-oriented KPIs behave.

\subsection{Intent Classification Accuracy Example}
\label{apx_intent_classification}
Intent classification evaluates whether the receiver correctly predicts the high-level task or action intended by the source sentence.

For the source sentence:

\begin{quote}
$S = ``\text{Book a flight to Montreal tomorrow.}''$
\end{quote}

the ground-truth intent is:

\begin{quote}
$Y = \text{BookFlight}$
\end{quote}

Assume the receiver predicts:

\begin{quote}
$\hat{Y} = \text{BookFlight}$
\end{quote}

Since the predicted intent matches the ground-truth intent, this sample is counted as correct.

\begin{equation}
\mathrm{Accuracy}
=
\frac{1}{1}
=
1.
\end{equation}

If the evaluation set contains $N$ samples, intent classification accuracy is computed as the number of correctly predicted intents divided by the total number of samples. This example shows that intent accuracy can be high even when some detailed semantic information is missing. In this case, the receiver recovered the general goal, BookFlight, but did not recover all required slot values.

\subsection{Slot Filling Precision, Recall, and F1 Example}
\label{apx_slot_filling_example}

Slot filling evaluates whether the receiver correctly extracts the task-relevant arguments from the source sentence.

For the source sentence:

\begin{quote}
$S = ``\text{Book a flight to Montreal tomorrow.}''$
\end{quote}

the ground-truth slots are:

\begin{quote}
destination = Montreal, date = tomorrow
\end{quote}

Assume the receiver predicts:

\begin{quote}
destination = Montreal
\end{quote}

The predicted destination slot is correct, but the date slot is missing. Therefore:

\begin{equation}
TP = 1,
\end{equation}

\begin{equation}
FP = 0,
\end{equation}

\begin{equation}
FN = 1.
\end{equation}

The precision is:

\begin{equation}
\mathrm{Precision}
=
\frac{TP}{TP + FP}
=
\frac{1}{1 + 0}
=
1.
\end{equation}

The recall is:

\begin{equation}
\mathrm{Recall}
=
\frac{TP}{TP + FN}
=
\frac{1}{1 + 1}
=
0.5.
\end{equation}

The F1 score is:

\begin{equation}
\mathrm{F1}
=
\frac{2 \times \mathrm{Precision} \times \mathrm{Recall}}
{\mathrm{Precision} + \mathrm{Recall}}.
\end{equation}

\begin{equation}
\mathrm{F1}
=
\frac{2 \times 1 \times 0.5}
{1 + 0.5}
\approx
0.67.
\end{equation}

This example shows that slot filling F1 gives partial credit when some semantic details are recovered correctly. The receiver preserved the destination information but lost the date information, so the task-relevant meaning was only partially recovered.

\subsection{Frame Accuracy / Exact Match Example}
\label{apx_frame_accuracy_example}

Frame accuracy evaluates whether the complete semantic frame is recovered correctly. A semantic frame usually includes both the intent and all required slots.

For the source sentence:

\begin{quote}
$S = ``\text{Book a flight to Montreal tomorrow.}''$
\end{quote}

the complete ground-truth frame is:

\begin{quote}
Intent: BookFlight \\
destination = Montreal \\
date = tomorrow
\end{quote}

Assume the receiver predicts:

\begin{quote}
Intent: BookFlight \\
destination = Montreal
\end{quote}

Although the intent is correct and the destination slot is correct, the date slot is missing. Therefore, the full predicted frame does not exactly match the ground-truth frame.

\begin{equation}
\mathrm{Frame\ Accuracy}
=
0.
\end{equation}

This example shows that frame accuracy is stricter than intent accuracy and slot F1. It is useful when the task can only succeed if all required semantic components are recovered correctly.

\subsection{Question Answering Exact Match and F1 Example}
\label{apx_qa_example}

Question answering evaluates whether the receiver produces the correct answer rather than reconstructing the original sentence.

Consider the following question:

\begin{quote}
Question: ``Where is the flight going?''
\end{quote}

Ground-truth answer:

\begin{quote}
$Y = ``\text{Montreal}''$
\end{quote}

Assume two possible receiver outputs:

\begin{quote}
$\hat{Y}_1 = ``\text{Montreal}''$ \\
$\hat{Y}_2 = ``\text{to Montreal city}''$
\end{quote}

For $\hat{Y}_1$, the predicted answer exactly matches the ground-truth answer. Therefore, Exact Match = 1.

For $\hat{Y}_2$, the predicted answer does not exactly match the ground-truth answer as a full string. Therefore, Exact Match = 0.

However, token-level F1 can still give partial credit. The ground-truth answer contains the token ``Montreal,'' and the predicted answer ``to Montreal city'' also contains ``Montreal.''

\begin{equation}
\mathrm{Number\ of\ overlapping\ tokens}
=
1.
\end{equation}
\begin{equation}
\mathrm{Number\ of\ predicted\ tokens}
=
3.
\end{equation}
\begin{equation}
\mathrm{Number\ of\ ground\text{-}truth\ tokens}
=
1.
\end{equation}
\begin{equation}
\mathrm{Precision}
=
\frac{1}{3}
=
0.33.
\end{equation}
\begin{equation}
\mathrm{Recall}
=
\frac{1}{1}
=
1.
\end{equation}
\begin{equation}
\mathrm{F1}
=
\frac{2 \times 0.33 \times 1}
{0.33 + 1}
\approx
0.50.
\end{equation}

This example shows that exact match is strict, while token-level F1 can give partial credit when the predicted answer contains the correct information but includes extra words.

\subsection{Retrieval Metrics Example: Recall@K and MRR}
\label{apx_retrieval_metrics}
Retrieval metrics evaluate whether the received semantic representation can retrieve the correct item from a candidate set.

Assume the source sentence is used as a query:

\begin{quote}
$S = ``\text{Book a flight to Montreal tomorrow.}''$
\end{quote}

Suppose the receiver retrieves a ranked list of documents or database entries:

\begin{quote}
Rank 1: Flight to Toronto tomorrow \\
Rank 2: Hotel booking in Montreal \\
Rank 3: Flight to Montreal tomorrow \\
Rank 4: Train ticket to Montreal \\
Rank 5: Flight to Vancouver tomorrow
\end{quote}

The correct item is: Flight to Montreal tomorrow

Its rank is:

\begin{equation}
\mathrm{rank}
=
3.
\end{equation}

Recall@K checks whether the correct item appears in the top $K$ results.

\begin{equation}
\mathrm{Recall@1}
=
0
\end{equation}

because the correct item is not ranked first.

\begin{equation}
\mathrm{Recall@3}
=
1
\end{equation}

because the correct item appears within the top three results.

Mean Reciprocal Rank, or MRR, uses the reciprocal of the rank of the first correct item. For this single query:

\begin{equation}
\mathrm{MRR}
=
\frac{1}{\mathrm{rank}}
=
\frac{1}{3}
\approx
0.33.
\end{equation}

This example shows that retrieval metrics evaluate whether the transmitted semantic information is sufficient to retrieve the correct item, even if the reconstructed text is not identical to the original sentence.

\subsection{Task Success Rate Example}
\label{apx_task_success_example}

Task success rate evaluates whether the final goal of the communication process is completed successfully.

Consider again the source sentence:

\begin{quote}
$S = ``\text{Book a flight to Montreal tomorrow.}''$
\end{quote}

The task is successful only if the system books a flight with the correct destination and date.

Ground-truth task requirement: destination = Montreal, date = tomorrow

Assume the receiver output is: destination = Montreal, date = missing

Because the date is missing, the system cannot complete the booking correctly. Therefore, this task is counted as unsuccessful.

\begin{equation}
\mathrm{Task\ Success\ Rate}
=
\frac{0}{1}
=
0.
\end{equation}

If an evaluation set contains 100 task requests and 82 are completed correctly, then:

\begin{equation}
\mathrm{Task\ Success\ Rate}
=
\frac{82}{100}
=
0.82.
\end{equation}

This example shows that task success rate is an end-to-end KPI. It does not directly measure text similarity but instead evaluates whether the receiver preserved enough semantic information to complete the intended task.

\section{Illustrative Examples of Reference-Free Text Semantic Communication KPIs}

\label{apx_reference_free}

% This appendix provides illustrative examples for the reference-free text semantic communication KPIs discussed in Section~3.3.
This appendix provides illustrative examples for the reference-free text semantic communication KPIs discussed in Section~\ref{sec:reference_free_text_kpis}. Unlike reference-based reconstruction metrics, these KPIs are designed for settings where the receiver does not have access to the original source sentence or a gold reference during deployment.

The examples use the same source and reconstructed sentences introduced in Appendix~A and summarized in Table~\ref{tab:reference_free_text_examples}, but the evaluation perspective is different: the receiver must estimate whether the received message is semantically acceptable using a learned quality estimator, a transmitted semantic anchor, a similarity threshold, or a fluency proxy.

\begin{table}[t]
\centering
\caption{Example sentences for reference-free text semantic communication KPIs}
\label{tab:reference_free_text_examples}
\begin{tabular}{p{0.35\linewidth} p{0.55\linewidth}}
\toprule
\textbf{Item} & \textbf{Value} \\
\midrule
Source sentence $S$ & ``The weather is cold today.'' \\
Meaning-preserving reconstruction $\hat{S}_1$ & ``It is freezing today.'' \\
Unrelated reconstruction $\hat{S}_2$ & ``A dog is sleeping on the sofa.'' \\
\bottomrule
\end{tabular}
\end{table}

\subsection{Quality Estimation Example}
\label{apx_quality_estimation}
Quality estimation evaluates the quality of a received or reconstructed sentence without comparing it to a gold reference sentence at the receiver. A learned quality estimation model may use only the reconstructed sentence, or it may use the source-side information available to the system depending on the variant.

For example, assume a reference-free quality estimation model receives the reconstructed sentence and outputs a scalar score in the range $[0, 1]$, where a larger value indicates higher predicted quality, as shown in Table~\ref{tab:quality_estimation_scores}.

\begin{table}[t]
\centering
\caption{Illustrative quality estimation scores}
\label{tab:quality_estimation_scores}
\begin{tabular}{p{0.45\linewidth} p{0.35\linewidth}}
\toprule
\textbf{Item} & \textbf{Value} \\
\midrule
Reconstruction $\hat{S}_1$ & ``It is freezing today.'' \\
Predicted QE score & 0.86 \\
Reconstruction $\hat{S}_2$ & ``A dog is sleeping on the sofa.'' \\
Predicted QE score & 0.22 \\
\bottomrule
\end{tabular}
\end{table}

In this illustrative example, the quality estimation model assigns a higher score to $\hat{S}_1$ because it is fluent and semantically compatible with the expected message context. It assigns a lower score to $\hat{S}_2$ because the output changes the topic completely. The numerical values are illustrative; in practice, the exact score depends on the specific QE model, checkpoint, training data, and input format.

This KPI is useful for deployment because it does not require a reference sentence at the receiver. However, it is model-dependent and may fail under domain shift.

\subsection{Anchor-Embedding Similarity Example}
\label{apx_anchor_embedding_example}

Anchor-embedding similarity evaluates semantic consistency using a compact semantic anchor transmitted by the sender. The sender computes an embedding of the source sentence and sends a compressed version, hash, or semantic tag as side information. The receiver computes an embedding of the reconstructed sentence and compares it with the transmitted anchor.

Let $e$ denote the semantic anchor produced at the transmitter and $\hat{e}$ denote the embedding computed at the receiver from the reconstructed sentence. The similarity can be computed as:

\begin{equation}
\mathrm{Sim}(e,\hat{e})
=
\cos(e,\hat{e}).
\end{equation}

For illustration, assume the transmitted semantic anchor is:

\begin{equation}
e
=
[0.80, 0.50, 0.30].
\end{equation}

For the meaning-preserving reconstruction $\hat{S}_1 = ``\text{It is freezing today}''$, suppose the receiver computes:

\begin{equation}
\hat{e}_1
=
[0.75, 0.55, 0.35].
\end{equation}

The cosine similarity is approximately:

\begin{equation}
\mathrm{Sim}(e,\hat{e}_1)
\approx
0.995.
\end{equation}

For the unrelated reconstruction $\hat{S}_2 = ``\text{A dog is sleeping on the sofa}''$, suppose the receiver computes:

\begin{equation}
\hat{e}_2
=
[0.10, 0.20, 0.90].
\end{equation}

The cosine similarity is approximately:

\begin{equation}
\mathrm{Sim}(e,\hat{e}_2)
\approx
0.49.
\end{equation}

Therefore, the receiver can treat $\hat{S}_1$ as semantically acceptable and $\hat{S}_2$ as semantically unreliable, without having access to the full original sentence $S$. This approach is reference-free at the receiver, but it introduces extra overhead because the anchor or tag must be transmitted.

\subsection{Semantic Success Rate and Semantic Outage Example}
\label{apx_semantic_success_outage}
Semantic success rate converts a semantic similarity score into a binary success or failure decision using a threshold. Semantic outage is the complementary probability that the received message does not satisfy the required semantic quality threshold.

For a single message, define success as:

\begin{equation}
\mathrm{Success}
=
\mathbf{1}
\left[
\mathrm{Sim} \geq \tau
\right],
\end{equation}

where $\tau$ is the semantic similarity threshold. The semantic success rate over $N$ messages is:

\begin{equation}
\mathrm{Success\ Rate}
=
\frac{1}{N}
\sum_i
\mathbf{1}
\left[
\mathrm{Sim}_i \geq \tau
\right].
\end{equation}

The semantic outage probability is:

\begin{equation}
\mathrm{Outage}
=
1 - \mathrm{Success\ Rate}.
\end{equation}

Suppose the semantic similarity threshold is $\tau = 0.80$. Table~\ref{tab:semantic_success_outage_example} reports the similarity values assumed for the two reconstructions in this appendix:

\begin{table}[t]
\centering
\caption{Illustrative semantic similarity values for semantic success and outage}
\label{tab:semantic_success_outage_example}
\begin{tabular}{p{0.40\linewidth} p{0.40\linewidth}}
\toprule
\textbf{Item} & \textbf{Value} \\
\midrule
$\mathrm{Sim}(S,\hat{S}_1)$ & 0.995 \\
$\mathrm{Sim}(S,\hat{S}_2)$ & 0.49 \\
Threshold $\tau$ & 0.80 \\
\bottomrule
\end{tabular}
\end{table}

For $\hat{S}_1$, the similarity is 0.995, which is greater than 0.80. Therefore, the semantic success indicator is 1. For $\hat{S}_2$, the similarity is 0.49, which is lower than 0.80. Therefore, the semantic success indicator is 0.

For these two examples:

\begin{equation}
\mathrm{Success\ Rate}
=
\frac{1 + 0}{2}
=
0.5.
\end{equation}

\begin{equation}
\mathrm{Outage}
=
1 - 0.5
=
0.5.
\end{equation}

This example shows how semantic similarity can be converted into a telecom-friendly reliability measure. The main challenge is choosing a meaningful threshold $\tau$ and specifying which semantic similarity model is used.

\subsection{Perplexity Example}
\label{apx_perplexity_example}

Perplexity is a reference-free fluency proxy. It measures how likely a language model considers a sentence. Lower perplexity usually indicates that the sentence is more fluent or natural according to the language model.

For a sentence with tokens $w_1, w_2, ..., w_T$, perplexity can be written as:

\begin{equation}
\mathrm{PPL}
=
\exp
\left(
-
\frac{1}{T}
\sum_t
\log p(w_t | w_{<t})
\right),
\end{equation}

where $p(w_t | w_{<t})$ is the probability assigned by the language model to token $w_t$ given the previous context. In practice, the reconstructed sentence is passed through a pretrained language model, the average negative log-likelihood is computed, and the exponential of this value gives the perplexity.

For illustration, Table~\ref{tab:perplexity_example} reports the perplexity values assumed for this illustration:

\begin{table}[t]
\centering
\caption{Illustrative perplexity values}
\label{tab:perplexity_example}
\begin{tabular}{p{0.35\linewidth} p{0.50\linewidth}}
\toprule
\textbf{Item} & \textbf{Value} \\
\midrule
$\hat{S}_1$ & ``It is freezing today.'' \\
$\mathrm{PPL}(\hat{S}_1)$ & 18 \\
$\hat{S}_2$ & ``A dog is sleeping on the sofa.'' \\
$\mathrm{PPL}(\hat{S}_2)$ & 22 \\
Garbled output & ``Weather sofa tomorrow cold dog is.'' \\
$\mathrm{PPL}(\mathrm{garbled\ output})$ & 120 \\
\bottomrule
\end{tabular}
\end{table}

The garbled output receives a much higher perplexity because it is less fluent and less predictable. However, perplexity alone does not guarantee semantic correctness. The unrelated sentence ``A dog is sleeping on the sofa'' can have low perplexity because it is fluent, even though it does not preserve the meaning of the original weather-related source sentence.

Therefore, perplexity can help detect corrupted or ungrammatical outputs, but it should not be used alone as a semantic communication KPI.

\section{Illustrative Examples of Telecom-Style Text Semantic State KPIs}
% appD.tex

\label{apx_telecom_state}

% This appendix provides illustrative examples for the telecom-style semantic state recovery KPIs discussed in Section~3.4. 
This appendix provides illustrative examples for the telecom-style semantic state recovery KPIs discussed in Section~\ref{sec:telecom_state_text_kpis}. Unlike sentence reconstruction metrics, these KPIs evaluate the recovery of a semantic state $W$ rather than the exact reconstruction of the original sentence $S$.

Consider the source sentence and semantic state in Table~\ref{tab:semantic_state_examples}:

\begin{table}[t]
\centering
\caption{Example source sentence and semantic states}
\label{tab:semantic_state_examples}
\begin{tabular}{p{0.30\linewidth} p{0.60\linewidth}}
\toprule
\textbf{Item} & \textbf{Value} \\
\midrule
Source sentence $S$ & ``Book a flight to Montreal tomorrow.'' \\
Semantic state $W$ & intent = BookFlight; destination = Montreal; date = tomorrow \\
Perfect recovery $\hat{W}_1$ & intent = BookFlight; destination = Montreal; date = tomorrow \\
Partial recovery $\hat{W}_2$ & intent = BookFlight; destination = Montreal; date = missing \\
Bad recovery $\hat{W}_3$ & intent = CancelBooking; destination = Toronto; date = tomorrow \\
\bottomrule
\end{tabular}
\end{table}

\subsection{Semantic Distortion Example}
\label{apx_semantic_distortion}
Semantic distortion measures how different the recovered semantic state $\hat{W}$ is from the original semantic state $W$. The distortion function $d(W,\hat{W})$ depends on how the semantic state is represented. For a structured semantic state with intent and slots, one simple distortion function can assign different weights to different semantic components.

For this example, assume the component weights in Table~\ref{tab:semantic_distortion_component_weights}:

\begin{table}[t]
\centering
\caption{Component weights for semantic distortion}
\label{tab:semantic_distortion_component_weights}
\begin{tabular}{p{0.25\linewidth} p{0.20\linewidth} p{0.45\linewidth}}
\toprule
\textbf{Component} & \textbf{Weight} & \textbf{Reason} \\
\midrule
intent & 0.5 & Wrong intent changes the main task. \\
destination & 0.25 & Destination is a required slot. \\
date & 0.25 & Date is a required slot. \\
\bottomrule
\end{tabular}
\end{table}

The distortion can be computed as the weighted sum of mismatched components:

\begin{equation}
d(W,\hat{W})
=
\sum_k
\alpha_k
\cdot
\mathbf{1}
\left[
W_k \neq \hat{W}_k
\right],
\end{equation}

where $\alpha_k$ is the weight of component $k$ and $\mathbf{1}[\cdot]$ is an indicator function that equals 1 when the component is incorrect or missing. Table~\ref{tab:semantic_distortion_examples} shows how this distortion score behaves across several recovered states.

\begin{table}[t]
\centering
\caption{Semantic distortion examples}
\label{tab:semantic_distortion_examples}
\begin{tabular}{p{0.25\linewidth} p{0.25\linewidth} p{0.15\linewidth} p{0.25\linewidth}}
\toprule
\textbf{Recovered state} & \textbf{Mismatched components} & \textbf{Distortion} & \textbf{Interpretation} \\
\midrule
$\hat{W}_1$ & none & 0 & Perfect semantic recovery \\
$\hat{W}_2$ & date & 0.25 & Partial semantic recovery \\
$\hat{W}_3$ & intent, destination & 0.75 & Severe semantic mismatch \\
\bottomrule
\end{tabular}
\end{table}

This example shows that semantic distortion can represent partial semantic failure. Unlike a binary success/failure metric, it can distinguish between a small missing detail and a major semantic error.

\subsection{Semantic Error Probability Example}
\label{apx_semantic_error_prob_example}

Semantic error probability measures how often the recovered semantic state differs from the original semantic state. For a discrete semantic state, the error event can be defined as $\hat{W} \neq W$.

\begin{equation}
P_{\mathrm{sem}}
=
P(\hat{W} \neq W).
\end{equation}

In an empirical test set with $N$ samples, it can be estimated as:

\begin{equation}
P_{\mathrm{sem}}
=
\frac{1}{N}
\sum_i
\mathbf{1}
\left[
\hat{W}_i \neq W_i
\right].
\end{equation}

Assume that five semantic messages are transmitted and the receiver outputs the states shown in Table~\ref{tab:semantic_error_probability_example}:

\begin{table}[t]
\centering
\caption{Semantic error probability example}
\label{tab:semantic_error_probability_example}
\begin{tabular}{p{0.10\linewidth} p{0.32\linewidth} p{0.32\linewidth} p{0.15\linewidth}}
\toprule
\textbf{Sample} & \textbf{Original $W$} & \textbf{Recovered $\hat{W}$} & \textbf{Error?} \\
\midrule
1 & BookFlight, Montreal, tomorrow & BookFlight, Montreal, tomorrow & No \\
2 & BookFlight, Montreal, tomorrow & BookFlight, Montreal, missing & Yes \\
3 & CancelBooking, Toronto, today & CancelBooking, Toronto, today & No \\
4 & PayBill, Hydro, Friday & PayBill, Hydro, Friday & No \\
5 & BookHotel, Montreal, tonight & BookFlight, Montreal, tonight & Yes \\
\bottomrule
\end{tabular}
\end{table}

Here, 2 out of 5 recovered semantic states are incorrect. Therefore:

\begin{equation}
P_{\mathrm{sem}}
=
\frac{2}{5}
=
0.40.
\end{equation}

This KPI is telecom-friendly because it resembles classical error probability, but the error is defined over semantic states rather than bits or symbols.

\subsection{Semantic Outage Probability Example}
\label{apx_semantic_outage_prob_example}

Semantic outage probability measures how often semantic quality falls below an acceptable threshold. Instead of requiring exact semantic-state equality, the system defines a semantic similarity or quality score $q(W,\hat{W})$ and declares outage when this score is below a threshold $\tau$.

\begin{equation}
P_{\mathrm{out}}
=
P(q(W,\hat{W}) < \tau).
\end{equation}

In an empirical test set, this can be estimated as:

\begin{equation}
P_{\mathrm{out}}
=
\frac{1}{N}
\sum_i
\mathbf{1}
\left[
q(W_i,\hat{W}_i) < \tau
\right].
\end{equation}

Assume the semantic similarity threshold is $\tau = 0.80$; Table~\ref{tab:semantic_outage_probability_example} reports the scores produced by five transmissions:

\begin{table}[t]
\centering
\caption{Semantic outage probability example}
\label{tab:semantic_outage_probability_example}
\begin{tabular}{p{0.15\linewidth} p{0.45\linewidth} p{0.20\linewidth}}
\toprule
\textbf{Sample} & \textbf{Semantic similarity $q(W,\hat{W})$} & \textbf{Outage?} \\
\midrule
1 & 0.95 & No \\
2 & 0.83 & No \\
3 & 0.76 & Yes \\
4 & 0.91 & No \\
5 & 0.60 & Yes \\
\bottomrule
\end{tabular}
\end{table}

Two out of five transmissions fall below the threshold. Therefore:

\begin{equation}
P_{\mathrm{out}}
=
\frac{2}{5}
=
0.40.
\end{equation}

This example shows how semantic outage converts a continuous semantic quality score into a reliability measure. It is useful for plotting semantic reliability versus SNR, bandwidth, rate, or latency.

\subsection{Rate-Semantic Tradeoff Example}
\label{apx_rate_tradeoff}
Rate-semantic tradeoff evaluates how semantic quality changes as the communication resource budget changes. In semantic communication, the goal is often to preserve enough meaning while transmitting fewer symbols, bits, or features.

Assume a text semantic communication system transmits different numbers of semantic symbols per message. The semantic quality is measured using a similarity score $q(W,\hat{W})$, as reported in Table~\ref{tab:rate_semantic_tradeoff_example}.

\begin{table}[t]
\centering
\caption{Rate-semantic tradeoff example}
\label{tab:rate_semantic_tradeoff_example}
\begin{tabular}{p{0.22\linewidth} p{0.25\linewidth} p{0.20\linewidth} p{0.23\linewidth}}
\toprule
\textbf{Configuration} & \textbf{Semantic symbols sent} & \textbf{Resource cost} & \textbf{Semantic similarity} \\
\midrule
A & 20 & low & 0.70 \\
B & 40 & medium & 0.86 \\
C & 80 & high & 0.93 \\
\bottomrule
\end{tabular}
\end{table}

If the acceptable semantic similarity threshold is $\tau = 0.80$, then configuration A fails while configurations B and C succeed. However, configuration C uses twice as many semantic symbols as configuration B while only improving the similarity from 0.86 to 0.93.

This illustrates the rate-semantic tradeoff: sending more semantic symbols can improve semantic quality, but it also increases communication cost. A practical semantic communication system should therefore report semantic quality together with resource usage, for example by plotting semantic similarity, semantic distortion, or semantic outage versus rate or SNR.

\section{Illustrative Examples of Image-Based Semantic Communication KPIs}
\label{apx_image_kpis}

This appendix provides a controlled experimental illustration of selected image-based KPI families discussed in Section~4. The goal is not to benchmark a complete semantic communication system, but to demonstrate how different sample-level KPIs respond to increasing visual degradation and semantic mismatch. The experiment uses one original street-scene image as the source image ($I$) and several receiver-side outputs ($\hat{I}_k$) generated at different degradation levels. The original image ($I$) is compared independently with each receiver-side output ($\hat{I}_k$). Pixel-fidelity metrics, structural/perceptual metrics, and CLIP-based visual semantic similarity are computed for each image pair. A simple channel-aware semantic reliability example is then provided by applying a threshold to CLIP image-image similarity.

The original image is COCO val2017 image 000000169996, available from the following URL: \url{http://images.cocodataset.org/val2017/000000169996.jpg}. The image contains an urban street scene with cyclists, bicycles, vehicles, a traffic light, road markings, buildings, and other background objects. This type of image is useful for illustrating image semantic communication because it contains both low-level visual details and task-relevant semantic objects.

In the experiment, the receiver-side outputs are generated using controlled degradations rather than a learned semantic communication model. This makes the experiment reproducible and easy to interpret. The mild output represents a high-quality channel or weak compression condition, the medium output represents moderate visual degradation, the severe output represents a low-quality channel condition, and the unrelated output represents semantic failure. If a very severe output is also generated, it can be used to show the transition between partial semantic preservation and semantic outage.

\subsection{Experimental Setup}

The experiment starts from one original image $I$ and creates several receiver-side outputs, summarized in Table~\ref{tab:image_kpi_experimental_setup}. Each output is treated as a possible received image $\hat{I}$ after transmission. The outputs are generated with increasing degradation severity so that the metric behavior can be compared in a controlled and reproducible way.

\begin{table*}[t]
\centering
\caption{Experimental setup for image-based semantic communication KPI examples}
\label{tab:image_kpi_experimental_setup}
\begin{tabular}{p{0.10\textwidth} p{0.35\textwidth} p{0.50\textwidth}}
\toprule
\textbf{Symbol} & \textbf{Meaning} & \textbf{Role in the experiment} \\
\midrule
$I$ & Original COCO validation image 000000169996 & Reference image used for pairwise KPI computation \\
$\hat{I}_1$ & Mild degradation & Simulates a high-quality receiver-side output \\
$\hat{I}_2$ & Medium degradation & Simulates moderate channel/compression degradation \\
$\hat{I}_3$ & Severe degradation & Simulates strong channel/compression degradation \\
$\hat{I}_4$ & Very severe degradation & Simulates near-failure visual reconstruction \\
$\hat{I}_5$ & Unrelated image & Represents a semantic failure case rather than a degraded version of $I$ \\
\bottomrule
\end{tabular}
\end{table*}

In short, the experiment asks: as the received image becomes more degraded, do pixel-level, perceptual, semantic, and reliability-based KPIs change in the expected direction?

\subsection{Receiver-Side Image Generation}

The degraded receiver-side outputs are generated using controlled image operations such as JPEG compression, Gaussian noise, Gaussian blur, and downsampling followed by upsampling, as detailed in Table~\ref{tab:receiver_side_image_generation}. These operations are used only to simulate different degradation levels. They should not be interpreted as the output of a trained semantic communication encoder-decoder model.

\begin{table*}[t]
\centering
\caption{Receiver-side image generation settings}
\label{tab:receiver_side_image_generation}
\begin{tabular}{p{0.055\textwidth} p{0.13\textwidth} p{0.36\textwidth} p{0.45\textwidth}}
\toprule
\textbf{Output} & \textbf{Degradation level} & \textbf{Typical operations} & \textbf{Expected visual effect} \\
\midrule
$\hat{I}_1$ & Mild & Light compression/noise/blur & Image remains visually close to the original \\
$\hat{I}_2$ & Medium & Moderate compression/noise/blur & Some details and local structures are degraded \\
$\hat{I}_3$ & Severe & Strong compression/noise/blur & Edges, textures, and small objects are strongly affected \\
$\hat{I}_4$ & Very severe & Heavy degradation and/or strong downsample-upsample & Image quality is poor and some visual meaning may be ambiguous \\
$\hat{I}_5$ & Unrelated & Different COCO image resized to the same size & Scene content is different from the original image \\
\bottomrule
\end{tabular}
\end{table*}

\begin{figure*}[t]
\centering
\includegraphics[width=1\textwidth]{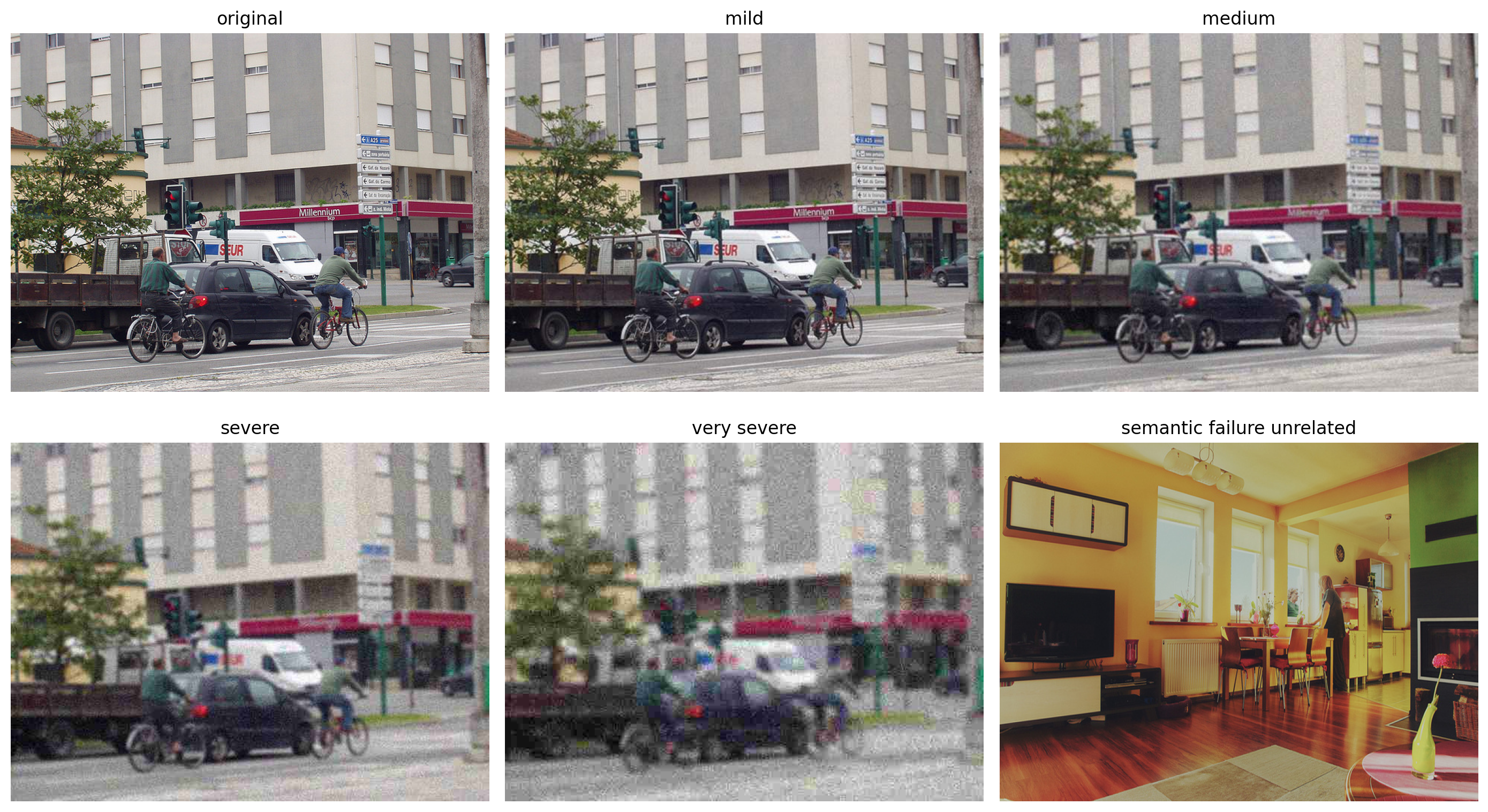}
\caption{The panel shows the original image, mild, medium, severe, and very severe receiver-side outputs, and an unrelated semantic-failure image.}
\label{fig:appE-degradation-panel}
\end{figure*}

Fig.~\ref{fig:appE-degradation-panel} shows the original image alongside the mild, medium, severe, and very severe receiver-side outputs, as well as the unrelated semantic-failure image.

\subsection{How the KPIs Were Computed}

For sample-level metrics, the original image $I$ is compared independently with each receiver-side output $\hat{I}_k$. Therefore, each row of the result table corresponds to one pair: $(I,\hat{I}_k)$. This pairwise comparison is used for MSE, PSNR, SSIM, MS-SSIM, LPIPS, and CLIP image-image similarity, as summarized in Table~\ref{tab:image_kpi_computation_summary}.

\begin{table*}[t]
\centering
\caption{Computation and interpretation of image-based semantic communication KPIs}
\label{tab:image_kpi_computation_summary}
\begin{tabular}{p{0.20\textwidth} p{0.16\textwidth} p{0.26\textwidth} p{0.33\textwidth}}
\toprule
\textbf{KPI family} & \textbf{Metrics} & \textbf{What is compared?} & \textbf{How to interpret the score} \\
\midrule
Pixel fidelity & MSE, PSNR & Pixel values of $I$ and $\hat{I}_k$ & Lower MSE and higher PSNR indicate closer numerical reconstruction. \\
Structural/perceptual quality & SSIM, MS-SSIM, LPIPS & Local structure or learned perceptual features of $I$ and $\hat{I}_k$ & Higher SSIM/MS-SSIM and lower LPIPS indicate better perceived visual quality. \\
Visual semantic similarity & CLIP image-image cosine similarity & CLIP image embeddings of $I$ and $\hat{I}_k$ & Higher cosine similarity indicates stronger representation-level semantic preservation. \\
Channel-aware semantic reliability & Semantic success/outage & A selected quality score compared with a threshold & An output is successful if the score satisfies the chosen threshold; otherwise it is in semantic outage. \\
Downstream task utility & Accuracy, mAP, IoU, Dice, Recall@K & Task prediction $\hat{Y}$ compared with task ground truth $Y$ & Requires task labels or a downstream model; not computed in the basic single-image experiment unless added. \\
\bottomrule
\end{tabular}
\end{table*}

This separation is important because not all image KPIs are computed in the same way. Some metrics compare image pairs, while task-level metrics require labels or a downstream model. Therefore, this controlled experiment focuses on sample-level image-pair KPIs and a threshold-based semantic outage example.

\subsection{Metric Implementation Notes}

The implementation should report the metric libraries and model variants used to compute the results, as outlined in Table~\ref{tab:image_kpi_implementation_notes}. This is especially important for learned metrics, because values can vary with preprocessing, image size, feature extractor, and model checkpoint.

\begin{table*}[t]
\centering
\caption{Implementation details to report for image-based KPI computation}
\label{tab:image_kpi_implementation_notes}
\begin{tabular}{p{0.15\textwidth} p{0.43\textwidth} p{0.36\textwidth}}
\toprule
\textbf{Metric} & \textbf{Implementation detail to report} & \textbf{Reason} \\
\midrule
MSE / PSNR & Pixel range and image normalization & PSNR depends on the maximum pixel value. \\
SSIM & Window size, channel handling, and implementation/library & SSIM can vary with local window settings. \\
MS-SSIM & Number of scales and implementation/library & Multi-scale weighting affects the score. \\
LPIPS & Backbone network, such as AlexNet, VGG, or SqueezeNet & LPIPS is model-dependent. \\
CLIP similarity & CLIP model name and preprocessing & CLIP embeddings depend on the selected checkpoint. \\
Semantic outage & Selected metric and threshold & Outage results depend directly on threshold choice. \\
\bottomrule
\end{tabular}
\end{table*}

The controlled image KPI experiment was implemented in Python. Pillow was used for image loading, RGB conversion, resizing, Gaussian blur, JPEG round-trip compression, and image saving. NumPy and math were used for MSE and PSNR computation, \texttt{skimage.metrics.structural\_similarity} was used for SSIM, \texttt{pytorch-msssim} was used for MS-SSIM, \texttt{lpips} with a PyTorch backend was used for LPIPS, and Hugging Face \texttt{transformers} was used to compute CLIP image-image cosine similarity. The LPIPS metric used the AlexNet backbone, and CLIP similarity was computed using the \texttt{openai/clip-vit-base-patch32} checkpoint.

The reference image was COCO val2017 image \texttt{000000169996.jpg}, loaded from the COCO validation image URL, converted to RGB, and kept at its original size of $640 \times 480$ pixels. The unrelated semantic-failure image was COCO val2017 image \texttt{000000000139.jpg}, converted to RGB, and resized to match the reference image size using bicubic interpolation. Four degraded receiver-side outputs were generated from the reference image using controlled combinations of downsampling/upsampling, Gaussian blur, additive Gaussian noise, and JPEG compression.

The degradation settings were: mild degradation with JPEG quality 65, Gaussian noise sigma = 4, blur radius 0.4, and downsample scale 0.85; medium degradation with JPEG quality 40, noise sigma = 10, blur radius 1.0, and scale 0.60; severe degradation with JPEG quality 22, noise sigma = 18, blur radius 1.8, and scale 0.40; and very severe degradation with JPEG quality 10, noise sigma = 30, blur radius 2.8, and scale 0.25.

For MSE, PSNR, SSIM, and MS-SSIM, images were converted to float32 arrays scaled to $[0,1]$. SSIM was computed with \texttt{channel\_axis = 2} and \texttt{data\_range = 1.0}, while MS-SSIM used PyTorch tensors in NCHW format with \texttt{data\_range = 1.0}. LPIPS used NCHW PyTorch tensors scaled from $[0,1]$ to $[-1,1]$. CLIP preprocessing was handled by \texttt{CLIPProcessor}, and image embeddings obtained from \texttt{CLIPModel.get\_image\_features} were L2-normalized before computing cosine similarity. The notebook exported the visual panel, metric table, Excel table, and KPI plot under the \texttt{outputs/controlled\_image\_kpi\_experiment/} directory.

Semantic outage was illustrated using CLIP image-image cosine similarity with threshold $\tau = 0.80$. Receiver-side outputs with CLIP similarity greater than or equal to the threshold were counted as semantic successes, while outputs below the threshold were counted as semantic outage cases. This threshold is used only for illustration and is not intended as a universal criterion.

\subsection{Sample-Level KPI Results}
\label{apx_sample_level_results}

Table~\ref{tab:sample_level_image_kpi_results} reports the sample-level KPI values computed between the original image $I$ and each receiver-side output $\hat{I}_k$. The expected trend is that mild degradation should remain close to the original image, while severe and very severe degradation should reduce pixel fidelity, structural quality, perceptual similarity, and semantic similarity. The unrelated image should produce the weakest semantic and structural correspondence.

\begin{table*}[t]
\centering
\caption{Sample-level KPI values for controlled image degradation. Each row compares the original image $I$ with one receiver-side output $\hat{I}_k$.}
\label{tab:sample_level_image_kpi_results}
\begin{tabular}{p{0.16\linewidth} p{0.08\linewidth} p{0.08\linewidth} p{0.08\linewidth} p{0.09\linewidth} p{0.08\linewidth} p{0.10\linewidth} p{0.24\linewidth}}
\toprule
\textbf{Output} & \textbf{MSE $\downarrow$} & \textbf{PSNR $\uparrow$} & \textbf{SSIM $\uparrow$} & \textbf{MS-SSIM $\uparrow$} & \textbf{LPIPS $\downarrow$} & \textbf{CLIP cosine $\uparrow$} & \textbf{Interpretation} \\
\midrule
$\hat{I}_1$ mild & 0.0034 & 24.6771 & 0.8138 & 0.9719 & 0.1701 & 0.9653 & Visually close to the original \\
$\hat{I}_2$ medium & 0.0072 & 21.3859 & 0.6182 & 0.9165 & 0.4732 & 0.8787 & Moderate degradation \\
$\hat{I}_3$ severe & 0.0105 & 19.7549 & 0.4394 & 0.8222 & 0.6322 & 0.8108 & Strong visual degradation \\
$\hat{I}_4$ very severe & 0.0140 & 18.5173 & 0.3109 & 0.6938 & 0.6854 & 0.7045 & Near-failure reconstruction \\
$\hat{I}_5$ unrelated & 0.0865 & 10.6273 & 0.2141 & 0.1346 & 0.8067 & 0.6691 & Different semantic content \\
\bottomrule
\end{tabular}
\end{table*}
\begin{figure}[t]
\centering
\includegraphics[
    width=0.80\linewidth,
    height=0.45\textheight,
    keepaspectratio
]{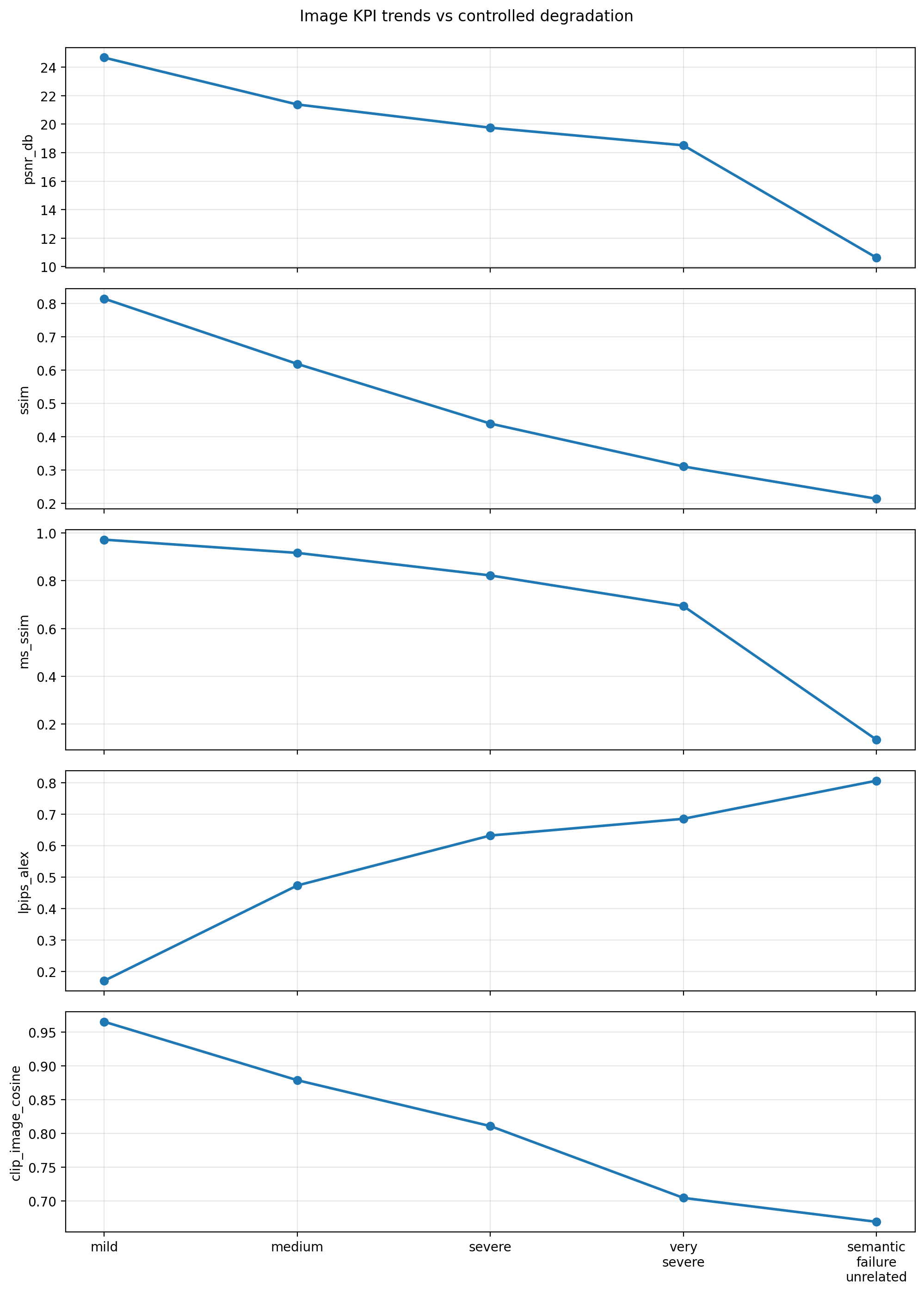}
\caption{KPI trends for the controlled image degradation experiment. Pixel, structural, perceptual, and semantic metrics are plotted across mild, medium, severe, very severe, and unrelated receiver-side outputs. Higher values indicate better quality for PSNR, SSIM, MS-SSIM, and CLIP cosine similarity, while lower values indicate better quality for LPIPS. The unrelated image acts as a semantic-failure reference.}
\label{fig:controlled-image-kpi-trends}
\end{figure}

\subsection{Interpreting the Sample-Level Results}

Fig.~\ref{fig:controlled-image-kpi-trends} plots these sample-level KPI trends across the five receiver-side outputs. The pixel-fidelity metrics evaluate numerical reconstruction quality. As degradation increases, MSE is expected to increase and PSNR is expected to decrease. However, these metrics do not identify whether the corrupted pixels belong to a task-critical object or an unimportant background region.

The structural and perceptual metrics evaluate visual quality beyond raw pixel error. SSIM and MS-SSIM are expected to decrease when edges, local contrast, and scene structure are damaged. LPIPS is expected to increase when the reconstructed image becomes perceptually different from the original image. These metrics are more aligned with visual perception than MSE and PSNR, but they still do not directly evaluate semantic task success.

The representation-level semantic metric evaluates whether the high-level visual content is preserved in a learned feature space. In this experiment, CLIP image-image cosine similarity compares the CLIP embedding of $I$ with the CLIP embedding of each $\hat{I}_k$. This score is expected to remain relatively high for mild degradation and drop for severe degradation or unrelated content. Nevertheless, CLIP similarity can miss small task-critical details, especially in domain-specific applications.

\subsection{Semantic Success and Outage Example}
\label{apx_image_success_outage}
To connect the experiment to telecom-style reliability, a semantic similarity score can be converted into a binary success or outage decision using a threshold. For example, if CLIP image-image similarity is used as the semantic quality score, semantic success can be defined as:

\begin{equation}
\mathrm{Success}_k
=
\mathbf{1}
\left[
\mathrm{CLIPSim}(I,\hat{I}_k) \geq \tau
\right],
\end{equation}

and semantic outage can be defined as:

\begin{equation}
\mathrm{Outage}_k
=
\mathbf{1}
\left[
\mathrm{CLIPSim}(I,\hat{I}_k) < \tau
\right].
\end{equation}

For all evaluated receiver-side outputs, the empirical outage rate is:

\begin{equation}
P_{\mathrm{out}}
=
\frac{1}{N}
\sum_k
\mathbf{1}
\left[
\mathrm{CLIPSim}(I,\hat{I}_k) < \tau
\right].
\end{equation}

The threshold $\tau$ should be chosen and reported explicitly. The purpose of this example is not to claim a universal semantic threshold, but to illustrate how a continuous semantic similarity score can be converted into a communication-style reliability measure, as shown in Table~\ref{tab:clip_semantic_success_outage}.

% \begin{table}[tbp]
\begin{table}[!htb]
\centering
\caption{Semantic success and outage example using CLIP image-image similarity with threshold $\tau = 0.80$}
\label{tab:clip_semantic_success_outage}
% \begin{tabular}{p{0.24\textwidth} p{0.18\textwidth} p{0.18\textwidth} p{0.20\textwidth} p{0.12\textwidth}}
\begin{tabular}{l c c c c}
\toprule
\textbf{Output} & \textbf{CLIP similarity} & \textbf{Threshold $\tau$} & \textbf{Semantic success?} & \textbf{Outage?} \\
\midrule
$\hat{I}_1$ mild & 0.965343 & 0.80 & Yes & No \\
$\hat{I}_2$ medium & 0.878795 & 0.80 & Yes & No \\
$\hat{I}_3$ severe & 0.810859 & 0.80 & Yes & No \\
$\hat{I}_4$ very severe & 0.704589 & 0.80 & No & Yes \\
$\hat{I}_5$ unrelated & 0.669182 & 0.80 & No & Yes \\
\bottomrule
\end{tabular}
\end{table}

\end{document}